\documentclass[11ptUSenglish,onecolumn]{article}
\usepackage[a4paper, total={6in, 8in}]{geometry}
\usepackage{natbib}
\usepackage{setspace}
\usepackage{amssymb}
\usepackage{amsmath}
\usepackage{amsthm}
\usepackage{xcolor}
\usepackage{hyperref}
\usepackage{authblk}

\usepackage{listings}
\newtheorem{Error}{Error}

\begin{document}




\title{ChatGPT and post-test probability
}


\author{Samuel J. Weisenthal\\\url{samuel_weisenthal@urmc.rochester.edu}}
\affil{University of Rochester School of Medicine and Dentistry, \\University of Rochester Medical Center,\\
    601 Elmwood Ave, 
            Rochester,
            14642, 
            NY,
            USA}

\maketitle
\section{Abstract}
Reinforcement learning-based large language models, such as ChatGPT, are  believed to have potential to aid human experts in many domains, including healthcare. There is, however, little work on ChatGPT's ability to perform a key task in healthcare: formal, probabilistic medical diagnostic reasoning. This type of reasoning is used,  for example, to update a pre-test probability to a post-test probability. In this work, we probe ChatGPT's ability to perform this task. In particular, we ask ChatGPT to give examples of how to use Bayes rule for medical diagnosis. Our prompts  range from queries that use terminology from pure probability (e.g., requests for a posterior of A given B and C) to queries that use terminology from medical diagnosis (e.g., requests for a posterior probability of Covid given a test result and cough).  We show how the introduction of medical variable names leads to an increase in the number of errors that ChatGPT makes. Given our results, we also show how one can use prompt engineering to facilitate ChatGPT's partial avoidance of these errors.  We discuss our results in light of recent commentaries on sensitivity and specificity. We also discuss how our results might inform new research directions for large language models.









\section{Introduction}
\label{sec:intro}
In concert with reinforcement learning techniques (e.g., \cite{schulman2017proximal}), large language models (e.g., \citep{radford2018improving,brown2020language}) have led to impressive, human-like chat capabilities.  Systems such as ChatGPT \citep{chatGPT} are  believed to have potential  to aid  human experts in many specialized domains, including medicine and public health  \citep{khan2023chatgpt,biswas2023role}. This enthusiasm is based partly on reports that ChatGPT has shown good performance on certain medically-relevant tasks, such as medical licensing exams \citep{kumah2023chatgpt,fijavcko2023can,kung2023performance,ayers2023evaluating,goodman2023accuracy}.
Indeed, large language models have excellent ability to access  information, which might help them answer medical questions and even generate  differential diagnoses \citep{10.1001/jama.2023.8288, 10.1093/jamia/ocad072}.   However, medical diagnosis also requires reasoning.  Although systems like ChatGPT can emulate reasoning,  authors  have criticized its true deductive reasoning capabilities \citep{shakarian2023independent, borji2023categorical,lievin2022can,wang2023can,bang2023multitask}.    To our knowledge, no study has investigated ChatGPT's ability to perform formal, probabilistic medical diagnostic reasoning. Probabilistic medical reasoning is used, for example, to update a pre-test probability to a post-test probability. 

In this work, we  probe ChatGPT's insight into probabilistic  medical diagnostic reasoning  by asking it to devise examples in which Bayes rule is used to obtain a post-test probability. In particular, we give it a challenging clinical scenario in which not only the test, but also the patient symptoms, must be taken into account. We discuss the errors that we observe in the responses of ChatGPT in the light of certain commentaries, such as those by \citet{moons1997limitations}, \cite{harrell2001regression}, \citet{moons2003sensitivity}, and \citet{guggenmoos2000validity}, on the assumption that sensitivity is a constant characteristic of the test, and therefore does not depend on patient covariates. 
We also explore prompt engineering approaches in order to improve ChatGPT's ability to perform probabilistic diagnostic reasoning, and we briefly discuss the implications of our results on new research directions for large language models. 

\section{Background}
\subsection{Diagnostic problem}
In the standard diagnostic problem, if $dz+$ is a binary random variable representing disease presence and $test+$ is a binary random variable representing a positive test result, we have that
 \begin{align}
 \label{eq:standard}
\notag p(dz+|test+) &= \frac{p(test+|dz+)p(dz+)}{p(test+)}\\
 &=\frac{(\text{sensitivity})(\text{pre-test probability})}{p(test+)},
 \end{align}
 where sensitivity is $p(test+|dz+)$ and pre-test probability is $p(dz+)$. Using the law of total probability, the denominator can be further written as
 $p(test+)=(\text{sensitivity})(\text{pre-test probability})  +(1-\text{specificity})(1-(\text{pre-test probability})).$

In this work, since we will be assessing the ability of ChatGPT to perform probabilistic diagnostic reasoning, we will also condition on the presence of a symptom, cough.  The symptom, cough, is also known as a patient covariate.  This is different from the standard medical diagnosis problems,  shown in (\ref{eq:standard}), which often only consider the case where when one conditions on a test. For example expositions of the prevailing framework, see, e.g., \citet{ledley1959reasoning},\citet{parikh2008understanding}, and \citet{trevethan2017sensitivity}. In particular, conditioning on the covariate cough makes this a more difficult diagnostic reasoning problem, as we will now describe.

If we let $cough$ be a binary random variable, we obtain the post-test probability
\begin{align}
\label{eq:ptpConditionOnCough}
p(dz+|test+,cough)&=\frac{p(test+|dz+,cough)p(dz+|cough)}{p(test+|cough)}\\
\notag &=
\frac{(\text{sensitivity})(\text{pre-test probability})}{p(test+|cough)}.
\end{align}  
Note that  the denominator can be rewritten
\begin{align*}
p(test+|cough)&=(\text{sensitivity})(\text{pre-test probability})\\&\ \ \  +(1-\text{specificity})(1-(\text{pre-test probability})).
\end{align*}

One might also  write 
\begin{align}
\label{eq:uncondpriorptp}
\notag p(dz+|test+,cough)&=\frac{p(test+,cough|dz+)p(dz+)}{p(test+,cough)}\\
\notag &=\frac{p(test+|cough,dz+)p(cough|dz+)p(dz+)}{p(test+,cough)}\\
&=\frac{(\text{sensitivity})p(cough|dz+)(\text{pre-test probability})}{p(test+,cough)}.
\end{align}  
Note that, as for (\ref{eq:ptpConditionOnCough}), the denominator in (\ref{eq:uncondpriorptp}) can be rewritten
\begin{align*}
p(test+,cough)&=p(test+,dz+,cough)+p(test+,dz-,cough)\\
&=p(test+,cough|dz+)p(dz+)\\
&\ \ \ +p(test+,cough|dz-)p(dz-)\\
&=p(test+|cough,dz+)p(cough|dz+)p(dz+)\\
&\ \ \ +p(test+|cough,dz-)p(cough|dz-)p(dz-)\\
&=p(test+|cough,dz+)p(cough|dz+)p(dz+)\\
&\ \ \ +(1-p(test-|cough,dz-))p(cough|dz-)(1-p(dz+))\\
&=(\text{sensitivity})p(cough|dz+)(\text{pre-test probability})\\
&\ \ \  +(1-\text{specificity})p(cough|dz-)(1-(\text{pre-test probability}))
\end{align*}


In both (\ref{eq:ptpConditionOnCough} and \ref{eq:uncondpriorptp}) the presence of the covariate, cough, leads to a different post-test probability expression than the one shown in (\ref{eq:standard}), which is most often seen in the standard medical diagnosis literature. The presence of the covariate cough therefore prohibits pure recall solutions.   In this work, we investigate whether this problem will be challenging for ChatGPT.

\subsection{Potential errors}
\label{sec:PotentialErrorsDefined}

We first list common errors that appear in probabilistic medical diagnostic reasoning, and we briefly discuss the connection of these errors  with the prevailing probabilistic medical diagnosis framework.  We will tally the frequency of these errors in the responses of ChatGPT. 

The first error occurs when one ignores information; namely, when one arrives at a post-test probability that ignores the fact that the patient has a cough.

\begin{Error}
\label{error:noConditionCough}
Even though one has information about the patient's cough, one still provides, as a post-test probability, $p(dz+|test+),$  instead of $p(dz+|test+,cough).$ I.e., one assumes either that $dz+\perp cough$ or  that the quantity of interest is  $p(dz+|test+)$ instead of $p(dz+|test+,cough)$.
\end{Error}


The assumption that the quantity of interest is $p(dz+|test+)$ instead of $p(dz+|test+,cough)$ might occur because the running example in the medical diagnosis literature focuses on $p(dz+|test+)$. In our prompts, we will request the post-test probability that conditions on cough. So, if ChatGPT answers with (\ref{eq:standard}), we will count it as Error \ref{error:noConditionCough}.



The next error occurs when one considers the test to be independent of patient covariates. 
\begin{Error}
\label{error:constantSens}
When computing the posterior $p(dz+|test+,dz+),$ one writes $p(test+|dz+)$  instead of $p(test+|dz+,cough)$ on the right hand side of (\ref{eq:ptpConditionOnCough}). I.e., one assumes that $test+\perp cough.$ 
\end{Error}

Error \ref{error:constantSens} assumes that sensitivity does not depend on patient covariates.  The assumption that sensitivity does not depend on patient covariates is criticized and thought to be generally false \citep{harrell2001regression,moons1997limitations,moons2003sensitivity,hlatky1984factors}.
Indeed, in Covid-19, which is our focus here, symptoms affect sensitivity, as shown in \citet{prince2021evaluation}, where the authors find rapid test sensitivities of 0.926 in symptomatic and 0.786 in asymptomatic patients. We would not expect ChatGPT to reference \citet{prince2021evaluation}, but if it assumes independence between the test and patient symptoms, which is a false assumption in general, we will consider it an error.
 Note that Error  \ref{error:constantSens} can only occur if one is writing the post-test probability $p(dz+|test+,cough),$ which is (\ref{eq:ptpConditionOnCough}). I.e., if one claims to write $p(dz+|test+),$ as in (\ref{eq:standard}), then writing sensitivity as $p(test+|dz+)$ is correct. Note that we ask ChatGPT to provide $p(dz+|test+,cough),$ so any time sensitivity does not condition on cough, it is Error \ref{error:constantSens}.



The next error occurs when one does not condition on cough in the pre-test probability.
Although we prefer the pre-test probability written in (\ref{eq:ptpConditionOnCough}), one might also write the pre-test probability as $p(dz+),$  as in (\ref{eq:uncondpriorptp}).

\begin{Error}
\label{error:uncondPrior}
One writes (\ref{eq:uncondpriorptp}) instead of (\ref{eq:ptpConditionOnCough}); hence, one writes $p(dz+)$ instead of $p(dz+|cough)$ as the pre-test probability.  
\end{Error}

Error \ref{eq:uncondpriorptp}, using pre-test $p(dz+)$ instead of $p(dz+|cough)$ when $p(dz+)$  appears in (\ref{eq:uncondpriorptp}), is not a probability error; it is a clinical  error. It is a clinical error because usually one would condition a pre-test probability on all available information (e.g., as with  \citet{wells1998use}, which gives us $p(dz+|covariates)$ instead of $p(dz+)$). In our case, one would ideally define the pre-test probability as $p(dz+|cough)$, which is more patient-specific (the disease is more likely when there is a cough), and arguably easier to obtain than both $p(dz+)$ and $p(cough|dz+),$ which must be obtained in concert to evaluate (\ref{eq:uncondpriorptp}). 
Note that  $p(covariates|dz+)$ is not typically reported in studies.
 It has also been noted that it is difficult to actually estimate $p(dz+),$ since prevalence (or pre-test probability) is almost always conditional on something \citep{harrell2001regression}. For these reasons, providing (\ref{eq:uncondpriorptp}) can be considered an error.
 
 Error \ref{error:uncondPrior} might be induced by the focus in the prevailing literature on the concept of the unconditional ``prevalence.''  As with the claim that sensitivity is defined as $p(test+|dz+)$, one could also claim that prevalence is defined as  $p(dz+),$ in which case prevalence could only be used to compute  $p(test+|dz+)$. This, however, would make it inapplicable in most clinical settings \citep{moons2003sensitivity}.


Errors \ref{error:noConditionCough}-\ref{error:uncondPrior}, mentioned above, might be expected based on the medical diagnosis literature. We designate the following error, Error \ref{error:miscProb}, to apply to any other probability error besides the ones listed above.  
\begin{Error}
\label{error:miscProb}
    Miscellaneous probability error.
\end{Error}
In this work,  we are most concerned with Errors \ref{error:noConditionCough}-\ref{error:uncondPrior}, but we do tally the occurrences of Error \ref{error:miscProb}. We check for Error \ref{error:miscProb}  in the first symbolic expression that ChatGPT gives for $p(dz|test,cough).$ Sometimes, to complement this first equation, the system also provides follow up equations or lengthy numerical examples, which we do not check. 

\section{Methods}

Before challenging ChatGPT's to perform probabilistic medical diagnostic reasoning, we first assess whether it can correctly perform probabilistic reasoning at all.
Namely, we ask ChatGPT to give an example of using Bayes rule to compute $P(C|A,B).$ Note that, as shown in the following prompt, we do so without medical terminology.  
\begin{lstlisting}[ escapeinside={(*}{*)},backgroundcolor = \color{yellow},caption={Pure probability},label={lst:pureProb}]
User
Can you give an example of using Bayes rule to compute posterior  probability of A given B and C?
\end{lstlisting}

Following this investigation, we then ask ChatGPT to perform medical diagnostic reasoning, based on a prompt that simply renames the variables $A,B,C$ as ``Covid,'' ``positive Covid test'' and ``cough,'' respectfully. 
This puts the problem in a medical context, unlike the prompt in Listing \ref{lst:pureProb}.  This more medically relevant prompt is shown below.
\begin{lstlisting}[ escapeinside={(*}{*)},backgroundcolor = \color{yellow},caption={Medical variables},label={lst:medicalNames}]
User
Can you give an example of using Bayes rule to compute posterior probability of Covid given a positive Covid test and a cough?
\end{lstlisting}
We repeat the prompt in Listing \ref{lst:medicalNames} 20 times and tally any errors that occur in ChatGPT's expressions for $P(\text{Covid}|\text{test,cough})$.


In the next prompt, we slightly amend the prompt in Listing \ref{lst:medicalNames} by adding a request to use the ``sensitivity'' and ``specificity'' of the test, which are medical diagnosis terms.  This might further encourage ChatGPT to access the medical diagnosis literature rather than the literature on pure probability.  In alignment with this, rather than asking ChatGPT to give a ``posterior'' probability, which is a probabilistic term, we ask it for a ``post-test probability, which is a medical term.
 

\begin{lstlisting}[ escapeinside={(*}{*)},backgroundcolor = \color{yellow},caption={Medical variables and diagnostic terminology},label={lst:medicalTerms}]
User
Can you give an example of using Bayes rule to compute post test probability in a patient with a positive Covid test and a cough using sensitivity and specificity of the test.
\end{lstlisting}

If ChatGPT neglects to take cough into account at all, i.e., if it only gives $P(Covid\mbox{-}19|test+)$, we record this as  Error \ref{error:noConditionCough}, and then ask ChatGPT to condition on cough, and to give $p(Covid\mbox{-}19|cough,test+),$ so that we can still evaluate the expression it gives for $p(Covid\mbox{-}19|cough,test+)$.

\begin{lstlisting}[ escapeinside={(*}{*)},backgroundcolor = \color{yellow}]
User
This is P(COVID-19|Positive test), not P(COVID-19|Positive test, cough) though isn't it?
\end{lstlisting}

We repeat the prompt in Listing \ref{lst:medicalTerms} 20 times and tally the errors in ChatGPT's expression for $p(Covid-19|test,cough).$  


We next develop a prompt that breaks the probabilistic medical reasoning procedure into steps and asks ChatGPT to avoid certain probabilistic assumptions that are associated with the errors in Section \ref{sec:PotentialErrorsDefined}.  In particular, we first prompt ChatGPT as follows.

\begin{lstlisting}[escapeinside={(*}{*)},backgroundcolor = \color{yellow},caption={Prompt engineering},label={lst:promptEngineer}]
User
In the following conversation, I will give you prompts. Please use the following steps to respond.
Step 1.  Determine what is the disease, what is the test result, and what is the covariate. 
Step 2. Write P(disease | test result, covariates) using Bayes rule. Condition on covariates throughout. Do not make any independence assumptions. Write the denominator using the law of total probability, conditional on covariates.
Step 3. Write the last formula in terms of the disease, test result, and covariate given in the prompt.
Step 4. State which parts of the last formula correspond to the sensitivity and pre-test probability. 
If you understand these steps, please say only "understood".
\end{lstlisting}

After ChatGPT has confirmed that it has understood the steps in Listing \ref{lst:promptEngineer}, we then repeat the prompt in Listing \ref{lst:medicalTerms}, which is shown below for convenience.

\begin{lstlisting}[ escapeinside={(*}{*)},backgroundcolor = \color{yellow}]
User
Can you give an example of using Bayes rule to compute post test probability in a patient with a positive Covid test and a cough using sensitivity and specificity of the test.
\end{lstlisting}

For the prompt in Listing \ref{lst:promptEngineer}, we show an example conversation, and then we perform 20 replicates and tally the expressions that are observed.

For each conversation, we opened a new, separate session with ChatGPT.  Conversations were performed through the browser interface to allow for interaction. All conversations were executed from the same user's account, and the complete conversations are recorded in the Appendix.  All conversations took place between May and July of 2023 with the standard, free, publicly available ChatGPT available at that time (version 3.5) at \url{https://chat.openai.com} (note that the timestamps on the conversations with links in the footnotes in Section \ref{sec:results} below are from November, because that is when the sharing links were created). Confidence intervals were computed using Wilson's method. 
 
\section{Results}
\label{sec:results}
\subsection{Pure probability}
\label{sec:ABC}
We first ask ChatGPT to compute the probability of $A$ given $B$ and $C$ as in the prompt in Listing \ref{lst:pureProb}.
We show an example conversation (for compactness here, we have used an ellipsis (...) to truncate some of ChatGPT's response, but we include the full conversation in \ref{app:fullconvos:ExampleProb}).\footnote{Alternatively, see \url{https://chat.openai.com/share/f1ade3a4-dab0-4baa-87e4-a8305a055888}.}
\begin{lstlisting}[ escapeinside={(*}{*)},backgroundcolor = \color{yellow}] 
User
Can you give an example of using Bayes rule to compute posterior probability of A given B and C?
\end{lstlisting}
\begin{lstlisting}[ escapeinside={(*}{*)},backgroundcolor = \color{lightgray}]
ChatGPT
Certainly! Bayes' rule allows us to calculate the posterior probability of an event A given evidence B and background information C. The formula for Bayes' rule is as follows:
P(A|B, C) = (P(B|A, C) * P(A|C)) / P(B|C)
...

\end{lstlisting}
We see that ChatGPT computes the posterior, $P(A|B, C) = \frac{P(B|A, C) P(A|C)}{P(B|C)},$ correctly.   Interestingly, even though ChatGPT is completely unprompted to give an example in a medical setting, it gives, in addition to  the symbolic expression for $P(A|B,C),$ a real-world medical example, which is reported in \ref{app:fullconvos:ExampleProb}). 

Results for the 20 repetitions of this conversation are shown in Table \ref{tab:ABCtally}.  The full (annotated) conversations are recorded in \ref{app:fullconvos:ProbReplicates}. 
\begin{table}[]
    \centering
    \begin{tabular}{l|l|l}
         Expression &  Tally & Proportion (95\% CI)\\
         \hline
         $P(A|B, C) = (P(B|A, C) P(A|C)) / P(B|C)$ & 19/20 & 0.95 (0.76, 1.00)\\
               \hline
          $P(A|B, C) = (P(B, C|A) P(A)) / P(B, C)$ & 1/20 & 0.05 (0.00, 0.24)
    \end{tabular}
    \caption{Post-test probability expressions given in the responses of ChatGPT when prompted to give an example of using Bayes rule to compute $P(A|B,C).$}
    \label{tab:ABCtally}
\end{table} 
Note that, in the conversations tallied in Table \ref{tab:ABCtally}, ChatGPT does not make any of the Errors \ref{error:noConditionCough}-\ref{error:uncondPrior}. 
Note that once, ChatGPT computes the pure-probability analog  to (\ref{eq:uncondpriorptp}), in which the prior/pretest is the ``unconditional'' $P(A)$ (in this case, for the prompt in Listing \ref{lst:pureProb}, since we do not specify that this is a medical diagnosis problem, this is not technically Error \ref{error:uncondPrior}) (note that we say analog, because the variable names are   $A,B,$ and $C$  instead of ``Covid'', ``test'' and ``cough'').
However, as in the example conversation above, it generally computes the pure-probability analog to (\ref{eq:ptpConditionOnCough}), because it defines the prior/pretest to be $p(A|C)$.  It makes no probability errors in the main expression for $P(A|B,C).$ 

\subsection{Medical variables}
\label{sec:covidconvoPosterResults}
We next use the prompt in Listing \ref{lst:medicalNames}, replacing the abstract variable names, $A,B,C,$ with medical names. We show an example conversation; note that the full conversation is reported in \ref{app:fullCovidExampleConvoPosterior}.\footnote{Alternatively, see \url{https://chat.openai.com/share/3e7a0ccb-9fe5-4ac7-9b9e-f1719750db55}.} 
\begin{lstlisting}[ escapeinside={(*}{*)},backgroundcolor = \color{yellow}]
User
Can you give an example of using Bayes rule to compute posterior probability of Covid given a positive Covid test and a cough?
\end{lstlisting}

\begin{lstlisting}[ escapeinside={(*}{*)},backgroundcolor = \color{lightgray}]
ChatGPT
...
P(Covid | Positive Test, Cough) = (P(Positive Test | Covid) * P(Cough | Covid) * P(Covid)) / P(Positive Test, Cough)
...
\end{lstlisting}
Note that ChatGPT writes (we use its names for variables here)
\begin{align*}
&P(\text{Covid} | \text{Positive Test, Cough})\\ 
&\ \ = \frac{P(\text{Positive Test} | \text{Covid})  P(\text{Cough} | \text{Covid})  P(\text{Covid})}{P(\text{Positive Test, Cough})}.
\end{align*} It should be  $P(\text{Positive Test} | \text{Covid, Cough}),$ not $P(\text{Positive Test}|\text{Covid})$, in the numerator. This is Error \ref{error:constantSens}.  It also makes Error \ref{error:uncondPrior}.  

Full, annotated results for the 20 repeated conversations, based on Listing \ref{lst:medicalNames}, are provided in \ref{app:fullCovidExampleConvoPostTestReplicatesPosteriors}. We tally the errors in Table \ref{tab:errorsPosterior}.

\begin{table}[htp!]
    \centering
    \begin{tabular}{l|l|l}
     Error & Tally & Proportion (95\% CI)\\
     \hline
        Error \ref{error:noConditionCough} (post-test does not condition on cough) & 0/20 & 0.00 (0.00, 0.16) \\
              \hline
        Error \ref{error:constantSens} (sensitivity does not condition on cough) & 10/20 & 0.50 (0.30, 0.70)\\
              \hline
        Error \ref{error:uncondPrior} (pre-test does not condition on cough)  & 12/20 & 0.60 (0.39, 0.78)\\
              \hline
        Error \ref{error:miscProb}  Miscellaneous probability error &  0/20 & 0.00 (0.00, 0.16)
    \end{tabular}
    \caption{Errors made in ChatGPT responses to 20 requests to provide an example of the posterior probability of Covid-19 in a patient with a positive test and a cough, as in Listing \ref{lst:medicalNames}}
    \label{tab:errorsPosterior}
\end{table}
We see that ChatGPT makes errors about half the time. This is in contrast to the results in Table \ref{tab:ABCtally}, in which only Error \ref{error:uncondPrior} was made (and, in the context in Section \ref{sec:ABC}, because no medical context was given in the prompt, Error \ref{error:uncondPrior} was not technically an error).  
The simple change in the names of the variables from $A,B,C$ to ``Covid'', ``test'', and ``cough'' has  changed the behavior of ChatGPT, even though the problem it is solving is essentially identical to the one it solved correctly in Section \ref{sec:ABC}. 

\subsection{Medical variables and diagnostic terminology}
\label{sec:covidconvoPostTestResults}
We will now show the results for the prompt in Listing \ref{lst:medicalTerms}, in which we also use the word ``post-test probability'' instead of ``posterior probability'' and included the words ``sensitivity'' and ``specificity''. We show an example conversation here, sometimes made more compact by ellipses. The full conversation is reported in \ref{app:fullCovidExampleConvoPostTest}.\footnote{Alternatively, see \url{https://chat.openai.com/share/e45c0c4e-c643-419c-88d2-01318d6b237e}.}

\begin{lstlisting}[ escapeinside={(*}{*)},backgroundcolor = \color{yellow}] 
User
Can you give an example of using Bayes rule to compute post test probability in a patient with a positive Covid test and a cough using sensitivity and specificity of the test.
\end{lstlisting}

\begin{lstlisting}[ escapeinside={(*}{*)},backgroundcolor = \color{lightgray}]
ChatGPT
...
P(COVID-19|Positive test) = P(Positive test|COVID-19) * P(COVID-19) / P(Positive test)
...
To calculate P(Positive test), we need to use the Law of Total Probability:

P(Positive test) = P(Positive test|COVID-19) * P(COVID-19) + P(Positive test|Not COVID-19) * P(Not COVID-19)
...
Therefore, the probability that the patient actually has COVID-19 given the positive test result and cough is ...
\end{lstlisting}

Even though ChatGPT says that it takes cough into account in its post-test probability, it does not condition on cough in the probabilistic expression, $P(\text{COVID-19}|\text{Positive Test})$. 
This is Error \ref{error:noConditionCough}.  
We respond that ChatGPT forgot to condition on cough, and it states that it misinterpreted the question and tries to correct itself.
\begin{lstlisting}[ escapeinside={(*}{*)},backgroundcolor = \color{yellow}]
User
This is P(COVID-19|Positive test), not P(COVID-19|Positive test, cough) though isn't it?
\end{lstlisting}

\begin{lstlisting}[ escapeinside={(*}{*)},backgroundcolor = \color{lightgray}]
ChatGPT
You are correct, my apologies for the confusion. I made an error in my interpretation of the question. To calculate the probability of COVID-19 given a positive test and cough, we need to use the following formula:
P(COVID-19|Positive test, cough) = P(Positive test, cough|COVID-19) * P(COVID-19) / P(Positive test, cough)
...
\end{lstlisting}
Note, however, that in correcting itself to condition on cough, it writes 
\begin{align*}
&\text{P(COVID-19}|\text{Positive test, cough})\\ 
&= \frac{\text{P(Positive test, cough}|\text{COVID-19) * P(COVID-19)}}{\text{P(Positive test, cough)}}.
\end{align*}
ChatGPT has written the post-test probability to depend on $p(dz+),$ not $p(dz+|cough),$ as in (\ref{eq:uncondpriorptp}), which is Error \ref{error:uncondPrior}.  

In the same response, ChatGPT then goes on to define \[P(\text{Positive test, cough}|\text{COVID-19})\] as follows.
\begin{lstlisting}[ escapeinside={(*}{*)},backgroundcolor = \color{lightgray}]
ChatGPT
...
We can calculate the probability of a positive test and cough given COVID-19 using the sensitivity and specificity of the test as follows:

P(Positive test, cough|COVID-19) = P(Positive test|COVID-19) * P(cough|COVID-19)
...
\end{lstlisting}
Note that writing $\text{p(Positive test}|\text{Covid)}$ instead of  $\text{p(Positive test}|\text{Covid, cough)}$ is Error \ref{error:constantSens}.  

We repeat this conversation 20 times and provide full, annotated conversations in \ref{app:fullCovidExampleConvoPostTestReplicatesPostTest}. Errors are tallied in Table \ref{tab:errors}.

\begin{table}[htp!]
    \centering
    \begin{tabular}{l|l|l}
     Error & Tally & Proportion (95\% CI)\\
     \hline
        Error \ref{error:noConditionCough} (post-test does not condition on cough) & 3/20 & 0.15 (0.05, 0.36) \\
              \hline
        Error \ref{error:constantSens} (sensitivity does not condition on cough) & 16/20 & 0.80 (0.58, 0.92)\\
              \hline
        Error \ref{error:uncondPrior} (pre-test does not condition on cough)  & 17/20 & 0.85 (0.64, 0.95) \\
              \hline
        Error \ref{error:miscProb}  Miscellaneous probability error &  4/20 & 0.20 (0.08, 0.42)
    \end{tabular}
    \caption{Errors made in ChatGPT responses to requests to provide an example of the post-test probability of Covid-19 in a patient with a positive test and a cough using sensitivity and specificity of the test, as in the prompt in Listing \ref{lst:medicalTerms}.}
    \label{tab:errors}
\end{table}

We see that ChatGPT consistently makes the same errors as before. Qualitatively,  the results in Table \ref{tab:errors} appear to be slightly worse than those in Table \ref{tab:errorsPosterior}. 
Table \ref{tab:errors} also shows some probability errors. For example, there is one instance in which ChatGPT writes $p(test+,cough)$ in the denominator when it conditioned on cough in the numerator, which was written as $p(test+|dz+,cough)p(dz+|cough).$ It should have instead written $p(test+|cough)$ in the denominator. There is another instance in which it writes $p(dz+|test,cough)$ in the numerator but only writes $p(test)$ in the denominator (both conversations are fully shown, with comments, in \ref{app:fullCovidExampleConvoPostTestReplicatesPostTest}). 

\subsection{Prompt engineering}
\label{sec:engineer}

We show an example conversation that was observed in response to prompt engineering. The full conversation is given in \ref{app:fullconvos:promptEngineerExample}.\footnote{Alternatively, see \url{https://chat.openai.com/share/bbacdb98-d476-4fe6-8548-2453392e9b74 }.}

\begin{lstlisting}[ escapeinside={(*}{*)},backgroundcolor = \color{yellow}]
User
In the following conversation, I will give you prompts. Please use the following steps to respond.
Step 1.  Determine what is the disease, what is the test result, and what is the covariate. 
Step 2. Write P(disease | test result, covariates) using Bayes rule. Condition on covariates throughout. Do not make any independence assumptions. Write the denominator using the law of total probability, conditional on covariates.
Step 3. Write the last formula in terms of the disease, test result, and covariate given in the prompt.
Step 4. State which parts of the last formula correspond to the sensitivity and pre-test probability. 
If you understand these steps, please say only "understood".
\end{lstlisting}
\begin{lstlisting}[ escapeinside={(*}{*)},backgroundcolor = \color{lightgray}]
ChatGPT
Understood.
\end{lstlisting}

\begin{lstlisting}[ escapeinside={(*}{*)},backgroundcolor = \color{yellow}]
User
Can you give an example of using Bayes rule to compute post test probability in a patient with a positive Covid test and a cough using sensitivity and specificity of the test.
\end{lstlisting}

\begin{lstlisting}[ escapeinside={(*}{*)},backgroundcolor = \color{lightgray}]
ChatGPT
Certainly! Let's walk through the steps using Bayes' rule to compute the post-test probability in a patient with a positive COVID test and a cough, given the sensitivity and specificity of the test.

Step 1:
Disease: COVID-19
Test result: Positive
Covariate: Cough

Step 2:
We want to compute P(COVID-19 | Positive test, Cough) using Bayes' rule:

P(COVID-19 | Positive test, Cough) = (P(Positive test | COVID-19, Cough) * P(COVID-19 | Cough)) / P(Positive test | Cough)

To compute the denominator, we can use the law of total probability:

P(Positive test | Cough) = P(Positive test | COVID-19, Cough) * P(COVID-19 | Cough) + P(Positive test | Not COVID-19, Cough) * P(Not COVID-19 | Cough)

Step 3:
Now let's rewrite the formula in terms of the given information:

P(COVID-19 | Positive test, Cough) = (P(Positive test | COVID-19, Cough) * P(COVID-19 | Cough)) / (P(Positive test | COVID-19, Cough) * P(COVID-19 | Cough) + P(Positive test | Not COVID-19, Cough) * P(Not COVID-19 | Cough))

Step 4:
In the last formula:

The sensitivity (true positive rate) corresponds to P(Positive test | COVID-19, Cough).
The pre-test probability corresponds to P(COVID-19 | Cough).
...
\end{lstlisting}
We see that ChatGPT appears to benefit from the prompt engineering.  Full results (with comments) for the 20 repeated conversations based on this engineered prompt are given in \ref{app:fullconvos:promptEngineerReplicates}, and the post-test probability expressions are given in Table \ref{tab:promptEngineerRep}.
\begin{table}[htp!]
    \centering
    \begin{tabular}{l|l|l}
        Expression & Tally & Proportion (95\% CI)  \\
        \hline
         $P(C19|test+,cough) = \frac{p(test+|C19,cough)p(C19|cough)}{p(test+|cough)}$ & $10/20$ & 0.50 (0.30, 0.70) \\
                   \ \ \ \ \ \ \ \ \ \ \ \ \ \ \ \ \ \  \  \ \ \ \ \ \ \ did expand $p(test+|cough)^{\dagger}$ &  \\
                                  \ \ \ \ \ \ \ \ \ \ \ \ \ \ \ \ \ \  \  \ \ \ \ \ \ \  did not name dz in 2/10$^*$ &  \\
               \hline
        
                        $P(C19|test+,cough) = \frac{p(test+|C19,cough)p(C19|cough)}{p(test+|cough)},$ & $9/20$ & 0.45 (0.26, 0.66) \\
                                                    \ \ \ \ \ \ \ \ \ \ \ \ \ \ \ \ \ \  \  \ \ \ \ \ \ \ did not expand $p(test+|cough)$& \\
                                                                             \ \ \ \ \ \ \ \ \ \ \ \ \ \ \ \ \ \  \  \ \ \ \ \ \ \ did not name dz, covar  in 1/9& \\
                                                                             \hline
             $P(C19 | test+, Cough) = \frac{P(test+ | C19, Cough) * P(C19 | Cough)}{P(test+, Cough)},$ & 
             1/20 & 0.05 (0.00, 0.24) \\
                  \ \ \ \ \ \ \ \ \ \ \ \ \ \ \ \ \ \  \  \ \ \ \ \ \ \ with error in denominator& \\
                  \hline
    \end{tabular}
    \caption{Expressions from the engineered prompt in Listing \ref{lst:promptEngineer}. For compactness, in this table, we write $C19,test+,$ and $cough$ in place of the different names ChatGPT gives to these items; full conversations and expressions are in \ref{app:fullconvos:promptEngineerReplicates}. 
    $^\dagger$Expanded refers to expanded according to the law of total probability, which was requested in the prompt.
    $^*$In some of the conversations, ChatGPT did not rename variables as C19, cough, or test, but instead left them in terms of disease, covariates, or test.  It almost always did so, except for 2 cases in the top row and one case in the second row.  The renaming was requested in the prompt. 
    }
    \label{tab:promptEngineerRep}
\end{table}
As shown in Table \ref{tab:promptEngineerRep}, 
given the prompt in Listing \ref{lst:medicalTerms}, ChatGPT does not appear to make Errors \ref{error:noConditionCough}-\ref{error:uncondPrior} in any conversation, which is a large contrast from the results in Tables \ref{tab:errorsPosterior} and \ref{tab:errors}. Usually, ChatGPT writes the post-test probability correctly, although there is some variation in its output; in particular, sometimes it expands the denominator according to the law of total probability, and sometimes it does not (both are reasonable responses, but the former was requested in the prompt).  Sometimes, as shown in Table \ref{tab:promptEngineerRep}, ChatGPT does not substitute the medical terms given in the prompt for the more generic terms ``disease, test, covariate.'' This was requested in the prompt, but this is not truly an error. ChatGPT does still make a probability error, as tallied in Table \ref{tab:promptEngineerRep}.

\section{Discussion}


In this work, we explored ChatGPT's ability to perform probabilistic medical diagnostic reasoning.
We gave ChatGPT a problem in which there was not just a test result, but also the presence of a symptom, which was an unusual,  difficult diagnostic problem. This can be challenging for language models, since they perform best when they have seen many examples of the problem they are solving in their training data.  This said, ChatGPT also has access to large amounts of text on pure probability calculations.  

We prompted ChatGPT to give us examples of post-test probability using terms that ranged from those in pure probability texts (Section \ref{sec:ABC}) to those that appear  in the medical diagnosis literature (Section \ref{sec:covidconvoPostTestResults}).   We found that the presence of medical terminology caused ChatGPT to perform poorly.
It is interesting to note that ChatGPT often considers sensitivity to be unrelated to patient covariates, which is Error \ref{error:constantSens}. 
This is an error that is commonly made by humans who are using the medical diagnosis framework, which is discussed in \citet{moons1997limitations,harrell2001regression}.  Large language models, because they learn from data that is generated by humans, can therefore learn not just harmful biases, as demonstrated by \citet{omiye2023large}, but also, as we show, tendencies to make statistical errors.  Possibly, our investigation might show us how to better develop educational resources on medical probabilistic reasoning. Concretely, it may be important to better integrate educational material about probability into the medical diagnosis literature.  This would allow ChatGPT---and human users---to simultaneously access text on both probability and medical reasoning in the same place, which might ultimately enhance their ability to perform  probabilistic medical diagnostic reasoning for nonstandard problems, such as the one we show here.

We also suspect that the terminology in the standard medical diagnostic reasoning framework takes ChatGPT to a (latent) space that is too far from where the probability question-answer examples themselves live. This causes probability errors.
The engineered prompt in Listing \ref{lst:promptEngineer} breaks the dependence between interpreting the medical terminology and the probabilistic manipulations themselves, and we find that this improves ChatGPT's responses, as shown in  Section \ref{sec:engineer}. The engineered prompt also guards against erroneous independence assumptions. The procedure in Listing \ref{lst:promptEngineer} is quite general; it could apply to a variety of disease, test, covariate scenarios; 
note, however, that ChatGPT still makes a probability error when using Listing \ref{lst:promptEngineer}, precluding the use of such a procedure for real diagnostic reasoning.  

It is remarkable that ChatGPT can, with or without prompt engineering, somewhat mimic probabilistic medical diagnostic reasoning.  
However, we show here that medical diagnostic reasoning, which requires that one must map medical terms to probability objects, perform reasoning on the probability objects, and then map back to medical terms, might not be reliably solved by systems such as ChatGPT. Although ChatGPT is considered a marvel of modern computing, it does not appear to be able to reliably perform rule-based logical reasoning. Ironically, computer programs usually excel at rule-based logical reasoning. We suspect that this might be a structural issue with ChatGPT, in which case more ``training data'' might not help. It might be necessary to combine language models, such as ChatGPT, with formal, probabilistic solvers, so that the language model can interpret the user input, a rule-based solver can be called to perform symbolic math, and then the language model can translate the results back to human-readable form.

\section{Acknowledgments}

The author gratefully acknowledges funding from the  National Institute of General Medical Sciences (NIGMS) under T32GM007356 (views expressed in this manuscript are solely the responsibility of the author and not the National Institutes of Health).

\bibliographystyle{elsarticle-harv} 
\bibliography{elsarticle-template-harv}

\newpage

\appendix
\section{Full conversations}
\label{app:fullconvos}
\subsection{Probability statements}
\subsubsection{Example conversation}
\label{app:fullconvos:ExampleProb}

\url{https://chat.openai.com/share/f1ade3a4-dab0-4baa-87e4-a8305a055888}

\begin{lstlisting}[ escapeinside={(*}{*)},backgroundcolor = \color{yellow}] 
User
Can you give an example of using Bayes rule to compute posterior  probability of A given B and C?
\end{lstlisting}

\begin{lstlisting}[ escapeinside={(*}{*)},backgroundcolor = \color{lightgray}]
ChatGPT
Certainly! Bayes' rule allows us to calculate the posterior probability of an event A given evidence B and background information C. The formula for Bayes' rule is as follows:

P(A|B, C) = (P(B|A, C) * P(A|C)) / P(B|C)

Here's an example to illustrate how Bayes' rule can be used:

Let's say we want to calculate the posterior probability of a person having a certain disease A, given that they have tested positive for a specific symptom B and have a family history of the disease C.

P(A|C): The prior probability of having the disease given the family history. This can be estimated based on statistical data or medical knowledge.
P(B|A, C): The probability of testing positive for the symptom given that the person has the disease and a family history. This can be determined through medical studies or testing accuracy rates.
P(B|C): The probability of testing positive for the symptom given the family history, regardless of whether the person has the disease. This can be estimated from testing accuracy rates and the prevalence of the symptom in the general population.
Once we have these probabilities, we can use Bayes' rule to calculate the posterior probability:

P(A|B, C) = (P(B|A, C) * P(A|C)) / P(B|C)

For example, let's say:

P(A|C) = 0.1 (10% chance of having the disease given the family history)
P(B|A, C) = 0.8 (80% chance of testing positive for the symptom given the disease and family history)
P(B|C) = 0.2 (20% chance of testing positive for the symptom given the family history)
Using these values, we can calculate the posterior probability:

P(A|B, C) = (0.8 * 0.1) / 0.2 = 0.4

Therefore, the posterior probability of the person having the disease given the positive symptom and family history is 0.4 or 40%.

Please note that the values used in this example are arbitrary and for illustrative purposes. In real-world scenarios, these probabilities would need to be estimated based on relevant data or expert knowledge.

\end{lstlisting}

\subsubsection{Replicated conversations}
\label{app:fullconvos:ProbReplicates}

{1}
\url{https://chat.openai.com/share/109c9e57-d04f-4e2e-a085-563359577820}
\begin{lstlisting}[ escapeinside={(*}{*)},backgroundcolor = \color{yellow}]
User
Can you give an example of using Bayes rule to compute posterior  probability of A given B and C?
\end{lstlisting}

\begin{lstlisting}[ escapeinside={(*}{*)},backgroundcolor = \color{lightgray}]

ChatGPT
Certainly! Bayes' rule is a fundamental theorem in probability theory that allows us to update our beliefs about an event based on new evidence. The formula for Bayes' rule is as follows:
P(A|B, C) = (P(B|A, C) * P(A|C)) / P(B|C)
Let's consider a specific example to illustrate how Bayes' rule can be used to compute the posterior probability of event A given events B and C.
Suppose we are interested in the probability of a patient having a certain disease (event A) given that they have tested positive for the disease (event B) and they have a family history of the disease (event C). We have the following information:
P(A) = 0.05 (prior probability of having the disease)
P(B|A, C) = 0.9 (probability of testing positive given the disease and family history)
P(A|C) = 0.08 (probability of having the disease given family history)
P(B|C) = 0.15 (probability of testing positive given family history)
Using Bayes' rule, we can compute the posterior probability P(A|B, C) as follows:
P(A|B, C) = (P(B|A, C) * P(A|C)) / P(B|C)
= (0.9 * 0.08) / 0.15
= 0.048 / 0.15
= 0.32
Therefore, the posterior probability of the patient having the disease given a positive test result and a family history of the disease is 0.32.
Please note that this is just a hypothetical example, and the actual probabilities and values would depend on the specific context or scenario.
\end{lstlisting}
$P(A|B, C) = (P(B|A, C) * P(A|C)) / P(B|C)$ is correct.

{2}
\url{https://chat.openai.com/share/f3f4ebfe-4fc9-44fb-ace5-1c6f505a20ab}
\begin{lstlisting}[ escapeinside={(*}{*)},backgroundcolor = \color{yellow}]
User
Can you give an example of using Bayes rule to compute posterior  probability of A given B and C?
\end{lstlisting}

\begin{lstlisting}[ escapeinside={(*}{*)},backgroundcolor = \color{lightgray}]
ChatGPT
Certainly! Let's say we have two events, A and B, and we want to compute the posterior probability of event A given that events B and C have occurred. We can use Bayes' rule to do this.
Bayes' rule states:
P(A|B, C) = (P(B|A, C) * P(A|C)) / P(B|C)
Here's an example to illustrate how to use Bayes' rule:
Suppose we have the following probabilities:
P(A) = 0.3 (prior probability of A)
P(B|A, C) = 0.8 (probability of B given A and C)
P(A|C) = 0.5 (probability of A given C)
P(B|C) = 0.6 (probability of B given C)
We want to find P(A|B, C), the posterior probability of A given B and C.
Using Bayes' rule, we can substitute the given probabilities:
P(A|B, C) = (P(B|A, C) * P(A|C)) / P(B|C)
= (0.8 * 0.5) / 0.6
= 0.4 / 0.6
= 0.67
Therefore, the posterior probability of A given that events B and C have occurred is 0.67 (or 67%).
\end{lstlisting}
$P(A|B, C) = (P(B|A, C) * P(A|C)) / P(B|C)$ is correct

{3}
\url{https://chat.openai.com/share/4df1be4d-57cb-41bb-9b9b-f5e410d92410}
\begin{lstlisting}[ escapeinside={(*}{*)},backgroundcolor = \color{yellow}]
User
Can you give an example of using Bayes rule to compute posterior  probability of A given B and C?
\end{lstlisting}


\begin{lstlisting}[ escapeinside={(*}{*)},backgroundcolor = \color{lightgray}]
ChatGPT
Certainly! Let's say we have two events, A and B, and we want to compute the posterior probability of A given B and C using Bayes' rule. Bayes' rule states:
P(A|B, C) = (P(B|A, C) * P(A|C)) / P(B|C)
Here's an example scenario:
Event A: The car's engine is faulty.
Event B: The car is producing strange noises.
Event C: The car is low on oil.
We want to find the probability that the car's engine is faulty (A) given that it is producing strange noises (B) and is low on oil (C).
Let's assume we have the following probabilities:
P(A|C) = 0.2 (the probability that the engine is faulty given it's low on oil)
P(B|A, C) = 0.8 (the probability of hearing strange noises when the engine is faulty and it's low on oil)
P(B|C) = 0.4 (the probability of hearing strange noises when the car is low on oil)
Now we can plug these values into Bayes' rule:
P(A|B, C) = (P(B|A, C) * P(A|C)) / P(B|C)
= (0.8 * 0.2) / 0.4
= 0.16 / 0.4
= 0.4
Therefore, the posterior probability of the car's engine being faulty (A) given that it is producing strange noises (B) and is low on oil (C) is 0.4, or 40%.
\end{lstlisting}
$P(A|B, C) = (P(B|A, C) * P(A|C)) / P(B|C)$ is correct.

{4}
\url{https://chat.openai.com/share/2d094158-f0bf-409e-a77f-5e97693cbbad}
\begin{lstlisting}[ escapeinside={(*}{*)},backgroundcolor = \color{yellow}]
User
Can you give an example of using Bayes rule to compute posterior  probability of A given B and C?
\end{lstlisting}

\begin{lstlisting}[ escapeinside={(*}{*)},backgroundcolor = \color{lightgray}]
ChatGPT
Certainly! Bayes' rule is a fundamental theorem in probability theory that allows us to update our beliefs or assess the posterior probability of an event A given the occurrence of events B and C. The formula for Bayes' rule is as follows:
P(A|B, C) = (P(B|A, C) * P(A|C)) / P(B|C)
Let's consider a simple example to illustrate its usage:
Suppose we have a medical test for a certain disease, and we want to calculate the probability that a patient has the disease (event A) given that the test is positive (event B) and the patient is a smoker (event C). We also know the following probabilities:
P(A) = 0.02 (prior probability of having the disease)
P(B|A, C) = 0.95 (probability of a positive test result given the disease and smoking)
P(A|C) = 0.10 (probability of having the disease given smoking)
P(B|C) = 0.20 (probability of a positive test result given smoking)
Now we can plug these values into Bayes' rule to compute the posterior probability:
P(A|B, C) = (P(B|A, C) * P(A|C)) / P(B|C)
P(A|B, C) = (0.95 * 0.10) / 0.20
P(A|B, C) = 0.095 / 0.20
P(A|B, C) (*$\approx$*) 0.475
Therefore, given that the test is positive and the patient is a smoker, the posterior probability of having the disease is approximately 0.475 or 47.5%.
\end{lstlisting}
$P(A|B, C) = (P(B|A, C) * P(A|C)) / P(B|C)$ is correct.

{5}
\url{https://chat.openai.com/share/b8f5823b-3faa-4769-a653-2cee82ab53bc}
\begin{lstlisting}[ escapeinside={(*}{*)},backgroundcolor = \color{yellow}]
User
Can you give an example of using Bayes rule to compute posterior  probability of A given B and C?
\end{lstlisting}

\begin{lstlisting}[ escapeinside={(*}{*)},backgroundcolor = \color{lightgray}]
ChatGPT
Certainly! Bayes' rule is a fundamental concept in probability theory that allows us to update our beliefs about an event based on new evidence. The formula for Bayes' rule is as follows:
P(A | B, C) = (P(B | A, C) * P(A | C)) / P(B | C)
Here's an example to illustrate how to use Bayes' rule to compute the posterior probability of event A given events B and C:
Let's say we have two events:
	Event A: A patient has a certain disease.
	Event B: A diagnostic test for the disease is positive.
	Event C: The patient belongs to a specific demographic group known to have a higher risk of the disease.
We have the following probabilities:
	P(A | C): The probability of a patient having the disease given their demographic group (prior probability).
	P(B | A, C): The probability of the test being positive given the patient has the disease and belongs to the specific demographic group (likelihood).
	P(B | C): The probability of the test being positive given the patient belongs to the specific demographic group (marginal likelihood).
Let's assume we know the following probabilities:
	P(A | C) = 0.2 (20% of patients from the demographic group have the disease).
	P(B | A, C) = 0.9 (the test has a 90% true positive rate).
	P(B | C) = 0.5 (the test has a 50% positive rate in the demographic group).
Now, we want to compute the posterior probability of a patient having the disease given that the test is positive and the patient belongs to the specific demographic group: P(A | B, C).
Using Bayes' rule:
P(A | B, C) = (P(B | A, C) * P(A | C)) / P(B | C)
Plugging in the values:
P(A | B, C) = (0.9 * 0.2) / 0.5
P(A | B, C) = 0.36
Therefore, the posterior probability of a patient having the disease given a positive test result and belonging to the specific demographic group is 0.36, or 36%.
\end{lstlisting}
$P(A | B, C) = (P(B | A, C) * P(A | C)) / P(B | C)$ is correct.

{6}
\url{https://chat.openai.com/share/b5ffc67b-4715-41d3-92ca-4899ba9efd2a}
\begin{lstlisting}[ escapeinside={(*}{*)},backgroundcolor = \color{yellow}]
User
Can you give an example of using Bayes rule to compute posterior  probability of A given B and C?
\end{lstlisting}

\begin{lstlisting}[ escapeinside={(*}{*)},backgroundcolor = \color{lightgray}]
ChatGPT
Certainly! Let's say we have two events A and B, and we want to compute the posterior probability of A given B and another event C. Bayes' rule allows us to calculate this probability using the following formula:
P(A | B, C) = (P(B | A, C) * P(A | C)) / P(B | C)
Here's an example to illustrate the calculation:
Suppose we are interested in determining the probability of a person having a particular disease A, given that they test positive for a diagnostic test B and have a family history of the disease C. Let's assume the following probabilities:
P(A) = 0.02 (prior probability of having the disease)
P(B | A, C) = 0.9 (probability of testing positive given the person has the disease and family history)
P(A | C) = 0.1 (probability of having the disease given the family history)
P(B | C) = 0.3 (probability of testing positive given the family history)
We can now substitute these values into Bayes' rule:
P(A | B, C) = (P(B | A, C) * P(A | C)) / P(B | C)
= (0.9 * 0.1) / 0.3
= 0.09 / 0.3
= 0.3
Therefore, the posterior probability of having the disease A, given that a person tests positive for the diagnostic test B and has a family history of the disease C, is 0.3 or 30%.
Please note that this example is for illustrative purposes, and the actual values of probabilities would vary depending on the specific context and data.
\end{lstlisting}
$P(A | B, C) = (P(B | A, C) * P(A | C)) / P(B | C)$ is correct.

{7}
\url{https://chat.openai.com/share/70ddee8f-a405-4cf9-9b59-02bf36a918f5}
\begin{lstlisting}[ escapeinside={(*}{*)},backgroundcolor = \color{yellow}]
User
Can you give an example of using Bayes rule to compute posterior  probability of A given B and C?
\end{lstlisting}

\begin{lstlisting}[ escapeinside={(*}{*)},backgroundcolor = \color{lightgray}]
ChatGPT
Certainly! Let's say we have three events: A, B, and C. We want to compute the posterior probability of event A given that events B and C have occurred. Bayes' theorem can help us with this calculation.
Bayes' theorem states:
P(A|B, C) = (P(B|A, C) * P(A|C)) / P(B|C)
In this equation:
	P(A|B, C) is the posterior probability of A given B and C (the probability we want to compute).
	P(B|A, C) is the probability of B given A and C.
	P(A|C) is the prior probability of A given C.
	P(B|C) is the probability of B given C.
To demonstrate, let's consider a specific example:
Suppose we are interested in determining the probability of a person being a smoker (A) given that they have a persistent cough (B) and a family history of lung cancer (C). We have the following information:
	P(B|A, C) = 0.8 (the probability of having a persistent cough given that the person is a smoker and has a family history of lung cancer).
	P(A|C) = 0.3 (the prior probability of being a smoker given a family history of lung cancer).
	P(B|C) = 0.2 (the probability of having a persistent cough given a family history of lung cancer).
Using these values, we can calculate the posterior probability:
P(A|B, C) = (P(B|A, C) * P(A|C)) / P(B|C)
= (0.8 * 0.3) / 0.2
= 0.24 / 0.2
= 1.2
Therefore, the posterior probability of someone being a smoker given that they have a persistent cough and a family history of lung cancer is 1.2 (or 120%).
Note that probabilities should always be between 0 and 1. In this case, the computed value exceeds 1, which indicates that there might be an error or inconsistency in the provided probabilities or the assumptions made.
\end{lstlisting}
$P(A|B, C) = (P(B|A, C) * P(A|C)) / P(B|C)$ is correct. Note that the same question was accidentally repeated in this conversation; only the first response was taken.

{8}
\url{https://chat.openai.com/share/063861fe-f510-4d13-ae04-6c58958d78aa}
\begin{lstlisting}[ escapeinside={(*}{*)},backgroundcolor = \color{yellow}]
User
Can you give an example of using Bayes rule to compute posterior  probability of A given B and C?
\end{lstlisting}

\begin{lstlisting}[ escapeinside={(*}{*)},backgroundcolor = \color{lightgray}]
ChatGPT
Certainly! Let's say we have three events: A, B, and C. We want to compute the posterior probability of event A given that events B and C have occurred. We can use Bayes' rule for this purpose.
Bayes' rule states:
P(A|B, C) = (P(B|A, C) * P(A|C)) / P(B|C)
Here's an example to illustrate the computation of the posterior probability:
Suppose we are interested in the probability of a person having a specific medical condition (event A) given that they tested positive for a related genetic marker (event B) and have a family history of the condition (event C). We have the following information:
P(A) = 0.02 (prior probability of having the medical condition)
P(B|A, C) = 0.8 (probability of testing positive for the genetic marker given the medical condition and family history)
P(A|C) = 0.05 (probability of having the medical condition given the family history)
P(B|C) = 0.15 (probability of testing positive for the genetic marker given the family history)
Using Bayes' rule, we can calculate the posterior probability P(A|B, C):
P(A|B, C) = (P(B|A, C) * P(A|C)) / P(B|C)
P(A|B, C) = (0.8 * 0.05) / 0.15
P(A|B, C) = 0.04 / 0.15
P(A|B, C) (*$\approx$*) 0.267
So, the posterior probability of having the medical condition given a positive test for the genetic marker and a family history is approximately 0.267, or 26.7%.
Please note that the values used in this example are arbitrary and for illustrative purposes only. In real-world scenarios, these probabilities would be specific to the context and obtained from relevant data or domain knowledge.
\end{lstlisting}
$P(A|B, C) = (P(B|A, C) * P(A|C)) / P(B|C)$ is correct

{9}
\url{https://chat.openai.com/share/9cb941a3-5af9-4771-86d5-474190c5f0c2}
\begin{lstlisting}[ escapeinside={(*}{*)},backgroundcolor = \color{yellow}]
User
Can you give an example of using Bayes rule to compute posterior  probability of A given B and C?
\end{lstlisting}

\begin{lstlisting}[ escapeinside={(*}{*)},backgroundcolor = \color{lightgray}]
ChatGPT
Certainly! Bayes' rule, also known as Bayes' theorem, is a fundamental concept in probability theory used to calculate conditional probabilities. The formula for Bayes' rule is as follows:
P(A|B, C) = (P(B|A, C) * P(A|C)) / P(B|C)
Here's an example to demonstrate the application of Bayes' rule:
Let's say we have a medical test for a disease called XYZ. We know that the probability of having the disease, P(Disease), is 0.01 (1%). We also know that the probability of the test correctly identifying the disease, given that the person has it, P(Positive|Disease), is 0.95 (95%). Additionally, the probability of a positive test result when the person does not have the disease, P(Positive|No Disease), is 0.02 (2%).
Now, let's suppose we want to calculate the probability that a person has the disease given that they tested positive, P(Disease|Positive), taking into account the overall prevalence of the disease.
Using Bayes' rule, we can calculate it as follows:
P(Disease|Positive) = (P(Positive|Disease) * P(Disease)) / P(Positive)
To compute P(Positive), we need to consider both scenarios where the person has the disease and where they don't:
P(Positive) = P(Positive|Disease) * P(Disease) + P(Positive|No Disease) * P(No Disease)
= 0.95 * 0.01 + 0.02 * 0.99
= 0.0095 + 0.0198
= 0.0293 (2.93%)
Now, substituting the values into the formula:
P(Disease|Positive) = (0.95 * 0.01) / 0.0293
= 0.0095 / 0.0293
(*$\approx$*) 0.324 (32.4%)
So, the probability that a person has the disease given that they tested positive is approximately 32.4%.
Remember, this is just a simplified example to illustrate how Bayes' rule is used to compute posterior probabilities. In practice, more factors and information would be considered to make accurate calculations.
\end{lstlisting}
$P(A|B, C) = (P(B|A, C) * P(A|C)) / P(B|C)$ is correct.

{10}
\url{https://chat.openai.com/share/3d4e938d-e3bf-4651-86ee-4a624423802f}
\begin{lstlisting}[ escapeinside={(*}{*)},backgroundcolor = \color{yellow}]
User
Can you give an example of using Bayes rule to compute posterior  probability of A given B and C?
\end{lstlisting}

\begin{lstlisting}[ escapeinside={(*}{*)},backgroundcolor = \color{lightgray}]
ChatGPT
Certainly! Let's say we have three events: A, B, and C. We want to compute the posterior probability of event A given that events B and C have occurred. We can use Bayes' rule to calculate this probability.
Bayes' rule states:
P(A|B, C) = (P(B, C|A) * P(A)) / P(B, C)
where:
P(A|B, C) is the posterior probability of A given B and C.
P(B, C|A) is the probability of B and C given A.
P(A) is the prior probability of A.
P(B, C) is the probability of B and C.
Let's assume we have the following probabilities:
P(A) = 0.3 (prior probability of A)
P(B, C|A) = 0.6 (probability of B and C given A)
P(B, C) = 0.4 (probability of B and C)
We can substitute these values into the formula to calculate the posterior probability of A given B and C:
P(A|B, C) = (P(B, C|A) * P(A)) / P(B, C)
= (0.6 * 0.3) / 0.4
= 0.18 / 0.4
= 0.45
Therefore, the posterior probability of A given that B and C have occurred is 0.45.
\end{lstlisting}
$P(A|B, C) = (P(B, C|A) * P(A)) / P(B, C)$ is correct, however the prior is not conditional (although we did not specify that it should be for this problem).

{11}
\url{https://chat.openai.com/share/d19d3265-a700-431d-b2db-c19ca61c6f95}
\begin{lstlisting}[ escapeinside={(*}{*)},backgroundcolor = \color{yellow}]
User
Can you give an example of using Bayes rule to compute posterior  probability of A given B and C?
\end{lstlisting}

\begin{lstlisting}[ escapeinside={(*}{*)},backgroundcolor = \color{lightgray}]
ChatGPT
Certainly! Let's say we have two events A and B, and we want to compute the posterior probability of A given that events B and C have occurred. We can use Bayes' rule to calculate it.
Bayes' rule states:
P(A|B, C) = (P(B|A, C) * P(A|C)) / P(B|C)
In this equation, P(A|B, C) represents the posterior probability of event A given events B and C, P(B|A, C) is the probability of event B given events A and C, P(A|C) is the probability of event A given event C, and P(B|C) is the probability of event B given event C.
Here's an example to illustrate the calculation:
Let's assume we are considering the probability of a person having a cold (A), given that they have a cough (B) and a fever (C). We have the following probabilities:
P(A) = 0.1 (prior probability of having a cold)
P(B|A, C) = 0.8 (probability of having a cough given a cold and fever)
P(A|C) = 0.3 (probability of having a cold given a fever)
P(B|C) = 0.5 (probability of having a cough given a fever)
Using Bayes' rule, we can calculate:
P(A|B, C) = (P(B|A, C) * P(A|C)) / P(B|C)
= (0.8 * 0.3) / 0.5
= 0.24 / 0.5
= 0.48
Therefore, the posterior probability of a person having a cold given that they have a cough and a fever is 0.48 or 48%.
Please note that the values I used in this example are arbitrary and for illustrative purposes. In practice, you would need specific probabilities based on relevant data or information.
\end{lstlisting}
$P(A|B, C) = (P(B|A, C) * P(A|C)) / P(B|C)$  is correct.

{12}
\url{https://chat.openai.com/share/d94d9ddc-0fa7-46a6-84f5-61e629461293}
\begin{lstlisting}[ escapeinside={(*}{*)},backgroundcolor = \color{yellow}]
User
Can you give an example of using Bayes rule to compute posterior  probability of A given B and C?
\end{lstlisting}

\begin{lstlisting}[ escapeinside={(*}{*)},backgroundcolor = \color{lightgray}]
ChatGPT
Certainly! Let's say we have three events: A, B, and C. We want to calculate the posterior probability of event A occurring, given that events B and C have already occurred. Bayes' theorem allows us to do that.
Bayes' theorem states:
P(A|B, C) = (P(B|A, C) * P(A|C)) / P(B|C)
Here's an example of how we can use Bayes' theorem to calculate the posterior probability:
Assume we have the following prior probabilities:
P(A) = 0.4 (prior probability of event A occurring)
P(B|A, C) = 0.7 (probability of event B occurring given A and C)
P(A|C) = 0.6 (probability of event A occurring given C)
P
\end{lstlisting}
$P(A|B, C) = (P(B|A, C) * P(A|C)) / P(B|C)$ is correct.

{13}
\url{https://chat.openai.com/share/cd519508-e71e-478a-8032-6b2065396928}
\begin{lstlisting}[ escapeinside={(*}{*)},backgroundcolor = \color{yellow}]
User
Can you give an example of using Bayes rule to compute posterior  probability of A given B and C?
\end{lstlisting}

\begin{lstlisting}[ escapeinside={(*}{*)},backgroundcolor = \color{lightgray}]
ChatGPT
Certainly! Bayes' rule allows us to calculate the posterior probability of an event given some observed evidence. The formula for Bayes' rule is as follows:
P(A|B, C) = (P(B|A, C) * P(A|C)) / P(B|C)
Let's say we have three events: A, B, and C. We want to calculate the posterior probability of event A occurring given that events B and C have both occurred.
Here's an example scenario:
	Event A: A person has a fever.
	Event B: The person has a sore throat.
	Event C: The person tested positive for a particular virus.
We want to calculate P(A|B, C), which is the probability that the person has a fever given they have a sore throat and tested positive for the virus.
To apply Bayes' rule, we need to know the following probabilities:
	P(B|A, C): The probability of having a sore throat given the person has a fever and tested positive for the virus.
	P(A|C): The probability of having a fever given the person tested positive for the virus.
	P(B|C): The probability of having a sore throat given the person tested positive for the virus.
Once we have these probabilities, we can plug them into Bayes' rule to calculate P(A|B, C).
Please note that in this response, I'm providing an example scenario to demonstrate the application of Bayes' rule. The specific values for the probabilities mentioned above would depend on the context and available data.
\end{lstlisting}
$P(A|B, C) = (P(B|A, C) * P(A|C)) / P(B|C)$ is correct.

{14}
\url{https://chat.openai.com/share/bc644e71-88f4-463d-adf4-be2eff6c1740}
\begin{lstlisting}[ escapeinside={(*}{*)},backgroundcolor = \color{yellow}]
User
Can you give an example of using Bayes rule to compute posterior  probability of A given B and C?
\end{lstlisting}

\begin{lstlisting}[ escapeinside={(*}{*)},backgroundcolor = \color{lightgray}]
ChatGPT
Certainly! Bayes' rule allows us to compute the posterior probability of event A given events B and C. Here's an example:
Let's say we have two events, A and B, and we want to compute the posterior probability of A given that B and C have occurred.
	P(A): Prior probability of event A
	P(B): Prior probability of event B
	P(C): Prior probability of event C
	P(A|B): Probability of event A given B
	P(A|C): Probability of event A given C
Using Bayes' rule, the posterior probability of A given B and C can be calculated as follows:
P(A|B,C) = (P(B|A,C) * P(A|C)) / P(B|C)
Here's a step-by-step breakdown:
	1	Compute the joint probability P(B, C) using the product rule:
	2	P(B, C) = P(B|C) * P(C)
	3	Compute the numerator, which is the joint probability of A and B given C:
	4	P(B|A, C) = (P(A, B|C) / P(B, C))
	5	This can be rewritten using the product rule:
	6	P(B|A, C) = (P(A|B, C) * P(B|C)) / P(C)
	7	Multiply the numerator by the prior probability P(A|C):
	8	P(B|A, C) * P(A|C) = [(P(A|B, C) * P(B|C)) / P(C)] * P(A|C)
	9	Simplify the expression:
	10	P(B|A, C) * P(A|C) = (P(A|B, C) * P(B|C) * P(A|C)) / P(C)
	11	Finally, divide by the marginal probability P(B|C) to normalize the posterior probability:
	12	P(A|B, C) = [(P(A|B, C) * P(B|C) * P(A|C)) / P(C)] / P(B|C)
This is the formula for calculating the posterior probability of event A given events B and C using Bayes' rule.
\end{lstlisting}
$P(A|B,C) = (P(B|A,C) * P(A|C)) / P(B|C)$ is correct.

{15}
\url{https://chat.openai.com/share/1ddec501-dab9-4ebf-8e89-455ad4d11a1a}
\begin{lstlisting}[ escapeinside={(*}{*)},backgroundcolor = \color{yellow}]
User
Can you give an example of using Bayes rule to compute posterior  probability of A given B and C?
\end{lstlisting}

\begin{lstlisting}[ escapeinside={(*}{*)},backgroundcolor = \color{lightgray}]
ChatGPT
Certainly! Let's say we have two events, A and B, and we want to compute the posterior probability of A given that events B and C have occurred. We can use Bayes' rule to calculate this probability. Bayes' rule states:
P(A | B, C) = (P(B | A, C) * P(A | C)) / P(B | C)
Here's an example to illustrate the calculation:
Let's say we have the following probabilities:
P(A) = 0.4 (prior probability of A)
P(B | A, C) = 0.8 (probability of B given A and C)
P(A | C) = 0.6 (probability of A given C)
P(B | C) = 0.7 (probability of B given C)
To find P(A | B, C), we substitute the given values into Bayes' rule:
P(A | B, C) = (P(B | A, C) * P(A | C)) / P(B | C)
= (0.8 * 0.6) / 0.7
= 0.48 / 0.7
(*$\approx$*) 0.6857
Therefore, the posterior probability of event A given that events B and C have occurred is approximately 0.6857.
\end{lstlisting}
$P(A | B, C) = (P(B | A, C) * P(A | C)) / P(B | C)$ is correct.

{16}
\url{https://chat.openai.com/share/ed878b75-36ae-42a9-939b-aeaa14165b79}
\begin{lstlisting}[ escapeinside={(*}{*)},backgroundcolor = \color{yellow}]
User
Can you give an example of using Bayes rule to compute posterior  probability of A given B and C?
\end{lstlisting}

\begin{lstlisting}[ escapeinside={(*}{*)},backgroundcolor = \color{lightgray}]
ChatGPT
Certainly! Bayes' rule allows us to update our prior beliefs about an event based on new evidence. To compute the posterior probability of event A given events B and C, we can use the following formula:
P(A | B, C) = (P(B | A, C) * P(A | C)) / P(B | C)
Let's consider an example to illustrate this. Suppose we are interested in determining the probability of a patient having a certain disease A, given that they have tested positive for a diagnostic test B and have a family history of the disease C.
Let's assume the following probabilities:
P(A) = 0.01 (prior probability of having the disease)
P(B | A, C) = 0.95 (probability of testing positive given the disease and family history)
P(A | C) = 0.02 (probability of having the disease given family history)
P(B | C) = 0.07 (probability of testing positive given family history)
Using Bayes' rule, we can calculate the posterior probability:
P(A | B, C) = (P(B | A, C) * P(A | C)) / P(B | C)
= (0.95 * 0.02) / 0.07
(*$\approx$*) 0.2714
Therefore, the posterior probability of the patient having the disease A, given that they tested positive for B and have a family history of C, is approximately 0.2714.
\end{lstlisting}
$P(A | B, C) = (P(B | A, C) * P(A | C)) / P(B | C)$ is correct.

{17}
\url{https://chat.openai.com/share/869c7ebf-afe2-4f16-9973-a4437607f86d}
\begin{lstlisting}[ escapeinside={(*}{*)},backgroundcolor = \color{yellow}]
User
Can you give an example of using Bayes rule to compute posterior  probability of A given B and C?
\end{lstlisting}

\begin{lstlisting}[ escapeinside={(*}{*)},backgroundcolor = \color{lightgray}]
ChatGPT
Certainly! Let's say we have two events, A and B, and we want to compute the posterior probability of A given B and C using Bayes' rule. Bayes' rule is defined as:
P(A|B, C) = (P(B|A, C) * P(A|C)) / P(B|C)
where:
P(A|B, C) is the posterior probability of A given B and C,
P(B|A, C) is the probability of B given A and C,
P(A|C) is the prior probability of A given C,
P(B|C) is the probability of B given C.
Let's assume we have the following probabilities:
P(A|C) = 0.3 (prior probability of A given C)
P(B|A, C) = 0.7 (probability of B given A and C)
P(B|C) = 0.5 (probability of B given C)
Using these values, we can calculate the posterior probability P(A|B, C) as follows:
P(A|B, C) = (P(B|A, C) * P(A|C)) / P(B|C)
= (0.7 * 0.3) / 0.5
= 0.21 / 0.5
= 0.42
Therefore, the posterior probability of A given B and C is 0.42.
\end{lstlisting}
$P(A|B, C) = (P(B|A, C) * P(A|C)) / P(B|C)$ is correct.

{18}
\url{https://chat.openai.com/share/dcc9a74a-aad9-4f53-b86a-aa52a398bb9c}
\begin{lstlisting}[ escapeinside={(*}{*)},backgroundcolor = \color{yellow}]
User
Can you give an example of using Bayes rule to compute posterior  probability of A given B and C?
\end{lstlisting}

\begin{lstlisting}[ escapeinside={(*}{*)},backgroundcolor = \color{lightgray}]
ChatGPT
Certainly! Let's say we have two events, A and B, and we want to compute the posterior probability of event A given that events B and C have occurred. We can use Bayes' rule to calculate it.
Bayes' rule states:
P(A|B, C) = (P(B|A, C) * P(A|C)) / P(B|C)
Here's an example scenario:
Suppose we are testing for a certain disease, and we have the following information:
P(A) = 0.01: Prior probability of having the disease (event A).
P(B|A, C) = 0.99: Probability of a positive test result (event B) given that the person has the disease (event A) and another relevant factor C.
P(A|C) = 0.02: Probability of having the disease (event A) given the relevant factor C.
P(B|C) = 0.03: Probability of a positive test result (event B) given the relevant factor C (regardless of having the disease).
Using Bayes' rule, we can calculate the posterior probability:
P(A|B, C) = (P(B|A, C) * P(A|C)) / P(B|C)
= (0.99 * 0.02) / 0.03
= 0.066 / 0.03
(*$\approx$*) 0.22
Therefore, the posterior probability of having the disease (A) given a positive test result (B) and the relevant factor (C) is approximately 0.22, or 22%.
\end{lstlisting}
$P(A|B, C) = (P(B|A, C) * P(A|C)) / P(B|C)$ is correct.

{19}
\url{https://chat.openai.com/share/871cef43-e209-457a-8a6d-ed96ecf923fc}
\begin{lstlisting}[ escapeinside={(*}{*)},backgroundcolor = \color{yellow}]
User
Can you give an example of using Bayes rule to compute posterior  probability of A given B and C?
\end{lstlisting}

\begin{lstlisting}[ escapeinside={(*}{*)},backgroundcolor = \color{lightgray}]
ChatGPT
Certainly! Let's say we have three events: A, B, and C. We want to compute the posterior probability of event A given that events B and C have occurred, using Bayes' rule.
Bayes' rule states:
P(A|B, C) = (P(B|A, C) * P(A|C)) / P(B|C)
Where:
P(A|B, C) is the posterior probability of A given B and C.
P(B|A, C) is the probability of B given A and C.
P(A|C) is the prior probability of A given C.
P(B|C) is the probability of B given C.
To compute the posterior probability of A given B and C, we need to know the values of the prior probabilities and conditional probabilities involved. Let's assume we have the following information:
P(A|C) = 0.6 (prior probability of A given C)
P(B|A, C) = 0.8 (probability of B given A and C)
P(B|C) = 0.4 (probability of B given C)
Using these values, we can substitute them into Bayes' rule to calculate the posterior probability of A given B and C:
P(A|B, C) = (P(B|A, C) * P(A|C)) / P(B|C)
= (0.8 * 0.6) / 0.4
= 0.48 / 0.4
= 1.2
Therefore, the posterior probability of event A given that events B and C have occurred is 1.2 or 120% (which can be interpreted as a normalization factor for the probability distribution).
\end{lstlisting}
$P(A|B, C) = (P(B|A, C) * P(A|C)) / P(B|C)$ is correct.

{20}
\url{https://chat.openai.com/share/99247049-979c-492c-a1cb-9e87e3686539}
\begin{lstlisting}[ escapeinside={(*}{*)},backgroundcolor = \color{yellow}]
User
Can you give an example of using Bayes rule to compute posterior  probability of A given B and C?
\end{lstlisting}

\begin{lstlisting}[ escapeinside={(*}{*)},backgroundcolor = \color{lightgray}]
ChatGPT
Certainly! Let's say we have two events, A and B, and we want to compute the posterior probability of event A given that events B and C have occurred. We can use Bayes' rule to calculate this probability.
Bayes' rule states:
P(A | B, C) = (P(B | A, C) * P(A | C)) / P(B | C)
Here's an example:
Suppose we are interested in the probability of a patient having a certain disease (event A) given that they have a positive test result (event B) and they are a smoker (event C). Let's assume we have the following information:
P(A) = 0.05 (prior probability of the disease)
P(B | A, C) = 0.95 (probability of a positive test result given the patient has the disease and is a smoker)
P(B | (*$\sim$*)A, C) = 0.10 (probability of a positive test result given the patient does not have the disease but is a smoker)
P(A | C) = 0.20 (probability of having the disease given the patient is a smoker)
P((*$\sim$*)A | C) = 0.80 (probability of not having the disease given the patient is a smoker)
Using these probabilities, we can calculate the posterior probability of the patient having the disease given a positive test result and being a smoker:
P(A | B, C) = (P(B | A, C) * P(A | C)) / P(B | C)
= (0.95 * 0.20) / ((0.95 * 0.20) + (0.10 * 0.80))
= 0.19 / (0.19 + 0.08)
= 0.19 / 0.27
(*$\approx$*) 0.7037
Therefore, the posterior probability of the patient having the disease given a positive test result and being a smoker is approximately 0.7037, or 70.37%.
\end{lstlisting}
$P(A | B, C) = (P(B | A, C) * P(A | C)) / P(B | C)$ is correct.

\subsection{Medical variables}

\subsubsection{Example conversation}
\label{app:fullCovidExampleConvoPosterior}

\url{https://chat.openai.com/share/3e7a0ccb-9fe5-4ac7-9b9e-f1719750db55}

\begin{lstlisting}[ escapeinside={(*}{*)},backgroundcolor = \color{yellow}]
User
Can you give an example of using Bayes rule to compute posterior probability of Covid given a positive Covid test and a cough?
\end{lstlisting}

\begin{lstlisting}[ escapeinside={(*}{*)},backgroundcolor = \color{lightgray}]
ChatGPT
Certainly! To compute the posterior probability of having Covid given a positive Covid test and a cough, we can apply Bayes' rule. Let's assume the following:
P(Covid) = 0.05 (Prior probability of having Covid, based on general prevalence)
P(Positive Test | Covid) = 0.95 (Probability of testing positive given that you have Covid)
P(Cough | Covid) = 0.80 (Probability of having a cough given that you have Covid)
P(Positive Test | No Covid) = 0.02 (Probability of testing positive given that you don't have Covid)
P(Cough | No Covid) = 0.10 (Probability of having a cough given that you don't have Covid)
We want to find P(Covid | Positive Test, Cough) (Posterior probability of having Covid given a positive test and a cough).
By applying Bayes' rule, we have:
P(Covid | Positive Test, Cough) = (P(Positive Test | Covid) * P(Cough | Covid) * P(Covid)) / P(Positive Test, Cough)
To calculate the denominator, we use the law of total probability:
P(Positive Test, Cough) = P(Positive Test, Cough | Covid) * P(Covid) + P(Positive Test, Cough | No Covid) * P(No Covid)
P(Positive Test, Cough | Covid) = P(Positive Test | Covid) * P(Cough | Covid)
P(Positive Test, Cough | No Covid) = P(Positive Test | No Covid) * P(Cough | No Covid)
P(No Covid) = 1 - P(Covid)
Substituting these values into the denominator:
P(Positive Test, Cough) = (P(Positive Test | Covid) * P(Cough | Covid) * P(Covid)) + (P(Positive Test | No Covid) * P(Cough | No Covid) * P(No Covid))
Finally, we can substitute the given probabilities into the equation and calculate the posterior probability.

\end{lstlisting}

\subsubsection{Replicated Conversations}
\label{app:fullCovidExampleConvoPostTestReplicatesPosteriors}

1
\url{https://chat.openai.com/share/323095b8-b6e2-4fd8-936c-a270914026fa}
\begin{lstlisting}[ escapeinside={(*}{*)},backgroundcolor = \color{yellow}]
User
Can you give an example of using Bayes rule to compute posterior probability of Covid given a positive Covid test and a cough?
\end{lstlisting}

\begin{lstlisting}[ escapeinside={(*}{*)},backgroundcolor = \color{lightgray}]
ChatGPT
Certainly! Let's assume we have the following information:
P(Covid) = 0.05 (prior probability of having Covid, based on general prevalence)
P(Positive Test | Covid) = 0.95 (probability of testing positive given that you have Covid)
P(Positive Test | No Covid) = 0.10 (probability of testing positive given that you do not have Covid)
P(Cough | Covid) = 0.70 (probability of having a cough given that you have Covid)
P(Cough | No Covid) = 0.20 (probability of having a cough given that you do not have Covid)
Now, we want to calculate the posterior probability of having Covid given a positive Covid test and a cough, using Bayes' rule.
Let's define the events:
A = Having Covid
B = Positive Test
C = Cough
We want to find P(A | B, C), the probability of having Covid given a positive test and a cough.
Using Bayes' rule, we have:
P(A | B, C) = (P(B | A, C) * P(A | C)) / P(B | C)
P(B | A, C) = P(Positive Test | Covid) * P(Cough | Covid) = 0.95 * 0.70 = 0.665
P(A | C) = P(Covid) = 0.05
P(B | C) = (P(Positive Test | Covid) * P(Cough | Covid) * P(Covid)) + (P(Positive Test | No Covid) * P(Cough | No Covid) * P(No Covid))
= (0.95 * 0.70 * 0.05) + (0.10 * 0.20 * (1 - 0.05))
= 0.03325 + 0.019
= 0.05225
Now we can calculate the posterior probability:
P(A | B, C) = (0.665 * 0.05) / 0.05225
= 0.03325 / 0.05225
(*$\approx$*) 0.636
Therefore, the posterior probability of having Covid given a positive Covid test and a cough is approximately 0.636, or 63.6%.
\end{lstlisting}
It writes  $P(B | A, C)= P(Positive Test | Covid) * P(Cough | Covid)$, which is Error \ref{error:constantSens}. 

2
\url{https://chat.openai.com/share/a448b919-9f40-49d5-95d6-628f8c1ade8e}

\begin{lstlisting}[ escapeinside={(*}{*)},backgroundcolor = \color{yellow}]
User
Can you give an example of using Bayes rule to compute posterior probability of Covid given a positive Covid test and a cough?
\end{lstlisting}

\begin{lstlisting}[ escapeinside={(*}{*)},backgroundcolor = \color{lightgray}]
ChatGPT
Certainly! To compute the posterior probability of having Covid given a positive Covid test result and a cough, we can use Bayes' theorem.
Let's define the following events:
A = Having Covid
B = Testing positive for Covid
C = Having a cough
We want to find P(A|B,C), the probability of having Covid given a positive Covid test and a cough.
According to Bayes' theorem, we have:
P(A|B,C) = (P(B|A,C) * P(A|C)) / P(B|C)
To calculate this, we need the following probabilities:
		P(B|A,C): The probability of testing positive for Covid given that you have Covid and have a cough. This is the sensitivity of the test, denoted as Sensitivity.
		P(A|C): The probability of having Covid given that you have a cough. This is the prevalence of Covid in the population.
		P(B|C): The probability of testing positive for Covid given that you have a cough. This is the specificity of the test, denoted as Specificity.
Assuming we have these probabilities, we can plug them into the formula to calculate the posterior probability:
P(A|B,C) = (Sensitivity * Prevalence) / ((Sensitivity * Prevalence) + ((1 - Specificity) * (1 - Prevalence)))
It's important to note that the values for Sensitivity, Specificity, and Prevalence may vary depending on the specific Covid test and the population being considered. These values are typically determined through empirical studies.
Please keep in mind that the above formula assumes conditional independence between the events A and B, given C, which may not always hold true in practice. Additionally, the accuracy of the computed posterior probability relies heavily on the accuracy of the input probabilities.
\end{lstlisting}
No errors; $P(A|B,C) = (P(B|A,C) * P(A|C)) / P(B|C)$ is correct.

3
\url{https://chat.openai.com/share/8ccbbdf0-56b0-4159-be12-f874b8c48cbb}
\begin{lstlisting}[ escapeinside={(*}{*)},backgroundcolor = \color{yellow}]
User
Can you give an example of using Bayes rule to compute posterior probability of Covid given a positive Covid test and a cough?
\end{lstlisting}

\begin{lstlisting}[ escapeinside={(*}{*)},backgroundcolor = \color{lightgray}]
ChatGPT
Certainly! Let's use Bayes' rule to compute the posterior probability of having Covid-19 given a positive Covid test and the presence of a cough.
Let's define the following terms:
		A: Event of having Covid-19
		B: Event of testing positive for Covid-19
		C: Event of having a cough
Bayes' rule states:
P(A | B, C) = (P(B | A, C) * P(A | C)) / P(B | C)
To compute the posterior probability of having Covid-19 given a positive Covid test and a cough, we need the following probabilities:
		Prior probability: P(A)
		This is the initial probability of having Covid-19 before considering any additional information. Let's assume the prior probability is 0.05 (or 5% based on available data).
		Sensitivity: P(B | A, C)
		This is the probability of testing positive for Covid-19 given that you have Covid-19 and a cough. Let's assume the sensitivity is 0.95 (or 95% based on the test accuracy).
		Specificity: P(B' | A', C)
		This is the probability of testing negative for Covid-19 given that you don't have Covid-19 but have a cough. Let's assume the specificity is 0.90 (or 90% based on the test accuracy).
		Prevalence: P(A | C)
		This is the probability of having Covid-19 given that you have a cough. Let's assume the prevalence based on local data is 0.10 (or 10% of people with a cough have Covid-19).
Now, we can calculate the posterior probability of having Covid-19 given a positive Covid test and a cough:
P(A | B, C) = (P(B | A, C) * P(A | C)) / P(B | C)
= (0.95 * 0.10) / ((0.95 * 0.10) + (0.10 * 0.90))
Calculating the values, we get:
P(A | B, C) = 0.095 / (0.095 + 0.09)
= 0.095 / 0.185
(*$\approx$*) 0.5135
Therefore, the posterior probability of having Covid-19 given a positive Covid test and a cough is approximately 0.5135 or 51.35%.
\end{lstlisting}
No errors; $P(A | B, C) = (P(B | A, C) * P(A | C)) / P(B | C)$ is correct.

4
\url{https://chat.openai.com/share/144d5bfa-b36d-4f35-ab4a-28c5ea985a0e}
\begin{lstlisting}[ escapeinside={(*}{*)},backgroundcolor = \color{yellow}]
User
Can you give an example of using Bayes rule to compute posterior probability of Covid given a positive Covid test and a cough?
\end{lstlisting}

\begin{lstlisting}[ escapeinside={(*}{*)},backgroundcolor = \color{lightgray}]
ChatGPT
Certainly! To compute the posterior probability of having Covid-19 given a positive Covid test and a cough, we can apply Bayes' rule. Let's denote the events as follows:
A: Having Covid-19
B: Testing positive for Covid
C: Having a cough
We want to find P(A|B, C), which is the posterior probability of having Covid-19 given a positive Covid test and a cough. Bayes' rule states:
P(A|B, C) = (P(B|A, C) * P(A|C)) / P(B|C)
To calculate this probability, we need to know the following values:
P(B|A, C): The probability of testing positive for Covid given that you have Covid-19 and a cough.
P(A|C): The prior probability of having Covid-19 given a cough.
P(B|C): The probability of testing positive for Covid given a cough.
Assuming we have these probabilities, we can substitute the values into Bayes' rule to calculate the posterior probability P(A|B, C). Please note that the actual values for these probabilities may vary depending on factors such as the accuracy of the test and the prevalence of Covid-19 in the population.
Here's an example to illustrate the calculation:
Let's say:
P(B|A, C) = 0.95 (the test has a 95% accuracy rate for detecting Covid-19 in individuals with a cough)
P(A|C) = 0.02 (the prior probability of having Covid-19 given a cough is 2%)
P(B|C) = 0.12 (the probability of testing positive for Covid given a cough is 12%)
Using Bayes' rule:
P(A|B, C) = (P(B|A, C) * P(A|C)) / P(B|C)
P(A|B, C) = (0.95 * 0.02) / 0.12
P(A|B, C) (*$\approx$*) 0.1583
Therefore, in this example, the posterior probability of having Covid-19 given a positive Covid test and a cough is approximately 0.1583 or 15.83%.
Please note that these values are used for illustration purposes only, and the actual probabilities may vary based on specific circumstances and data.
\end{lstlisting}
No errors, $P(A|B, C) = (P(B|A, C) * P(A|C)) / P(B|C)$ is correct.

5
\url{https://chat.openai.com/share/3ff6405d-9e96-45b4-bccb-9a1f45afb8ad}

\begin{lstlisting}[ escapeinside={(*}{*)},backgroundcolor = \color{yellow}]
User
Can you give an example of using Bayes rule to compute posterior probability of Covid given a positive Covid test and a cough?
\end{lstlisting}

\begin{lstlisting}[ escapeinside={(*}{*)},backgroundcolor = \color{lightgray}]
ChatGPT
Certainly! Let's use Bayes' rule to compute the posterior probability of having Covid given a positive Covid test and a cough.
Let's define the following events:
A = Having Covid
B = Positive Covid test
C = Having a cough
We want to find P(A|B,C), the probability of having Covid given a positive Covid test and a cough. According to Bayes' rule, we have:
P(A|B,C) = (P(B,C|A) * P(A)) / P(B,C)
To calculate this, we need the following probabilities:
		P(B,C|A): The probability of getting a positive Covid test and a cough given that a person has Covid. This information is usually provided by the sensitivity and specificity of the test. Let's assume the sensitivity of the test is 0.95 and the specificity is 0.98. This means that if a person has Covid, there is a 95% chance of testing positive (true positive rate), and if a person does not have Covid, there is a 98% chance of testing negative (true negative rate). Therefore, P(B,C|A) = 0.95 * P(C|A).
		P(A): The prior probability of having Covid. This information is not provided in the question, so let's assume it to be the prevalence rate of Covid in the population. Let's assume the prevalence rate is 0.01, which means that 1% of the population has Covid. Therefore, P(A) = 0.01.
		P(B,C): The probability of getting a positive Covid test and a cough. This can be calculated using the law of total probability:
P(B,C) = P(B,C|A) * P(A) + P(B,C|(*$\sim$*)A) * P((*$\sim$*)A)
P(B,C|(*$\sim$*)A) is the probability of getting a positive Covid test and a cough given that a person does not have Covid. Let's assume this probability to be 0.05, which means a 5% false positive rate. P((*$\sim$*)A) is the probability of not having Covid, which is 1 - P(A).
Now, let's substitute the values into Bayes' rule:
P(A|B,C) = (P(B,C|A) * P(A)) / P(B,C)
= (0.95 * P(C|A) * 0.01) / (P(B,C|A) * P(A) + P(B,C|(*$\sim$*)A) * P((*$\sim$*)A))
This calculation requires additional information about P(C|A), the probability of having a cough given that a person has Covid. Without this information, we cannot provide a specific numerical answer. The values of P(C|A) and P(C|(*$\sim$*)A) can vary depending on several factors.
Note: This is a simplified example, and in real-world scenarios, the calculation would involve more complex considerations, such as incorporating additional symptoms, adjusting for test accuracy, and updating probabilities as more information becomes available.
\end{lstlisting}
It writes $P(A|B,C) = (P(B,C|A) * P(A)) / P(B,C)$ which is Error \ref{error:uncondPrior}.
It defines $P(B,C|A)$ to depend on sensitivity, and states ``[l]et's assume the sensitivity of the test is 0.95 and the specificity is 0.98. This means that if a person has Covid, there is a 95\% chance of testing positive (true positive rate),'' which does not condition on cough, which is Error \ref{error:constantSens}. It then substitutes 0.95, which it called sensitivity, into the following posterior expression,
\begin{lstlisting}
P(A|B,C) = (P(B,C|A) * P(A)) / P(B,C)
= (0.95 * P(C|A) * 0.01) / (P(B,C|A) * P(A) + P(B,C|\sim A) * P((\sim A)).
\end{lstlisting}

6
\url{https://chat.openai.com/share/1e8aa1b9-afc9-4a64-a97b-06623f433a27}
\begin{lstlisting}[ escapeinside={(*}{*)},backgroundcolor = \color{yellow}]
User
Can you give an example of using Bayes rule to compute posterior probability of Covid given a positive Covid test and a cough?
\end{lstlisting}

\begin{lstlisting}[ escapeinside={(*}{*)},backgroundcolor = \color{lightgray}]
ChatGPT
Certainly! To compute the posterior probability of having COVID-19 given a positive COVID test and a cough, we can use Bayes' rule. Let's define the following probabilities:

P(C) - Prior probability of having COVID-19
P((*$\sim$*)C) - Prior probability of not having COVID-19
P(Pos|C) - Probability of testing positive given COVID-19
P(Pos|(*$\sim$*)C) - Probability of testing positive given not having COVID-19
P(C|Pos, Cough) - Posterior probability of having COVID-19 given a positive COVID test and a cough

Now, let's assume we have the following information:

P(C) = 0.05 (prior probability of having COVID-19, you can choose any value)
P((*$\sim$*)C) = 1 - P(C) = 0.95 (prior probability of not having COVID-19)
P(Pos|C) = 0.9 (probability of testing positive given COVID-19)
P(Pos|(*$\sim$*)C) = 0.2 (probability of testing positive given not having COVID-19)

Now, we can use Bayes' rule to compute the posterior probability:

P(C|Pos, Cough) = (P(Pos|C) * P(C) * P(Cough|C)) / P(Pos)

Here, P(Cough|C) represents the probability of having a cough given COVID-19. Since we don't have the exact value, let's assume it to be 0.8.

P(Cough|C) = 0.8

To calculate P(Pos), we need to consider the probabilities of testing positive in both cases (having COVID-19 and not having COVID-19):

P(Pos) = (P(Pos|C) * P(C) * P(Cough|C)) + (P(Pos|(*$\sim$*)C) * P((*$\sim$*)C) * P(Cough|(*$\sim$*)C))

Now, we can substitute the given values into the equation to find the posterior probability:

P(Pos) = (0.9 * 0.05 * 0.8) + (0.2 * 0.95 * P(Cough|(*$\sim$*)C))
P(Pos) = 0.036 + (0.19 * P(Cough|(*$\sim$*)C))

Finally, we can calculate the posterior probability:

P(C|Pos, Cough) = (P(Pos|C) * P(C) * P(Cough|C)) / P(Pos)
P(C|Pos, Cough) = (0.9 * 0.05 * 0.8) / (0.036 + (0.19 * P(Cough|(*$\sim$*)C)))

Please note that we assume independence between the presence of a cough and having COVID-19, which may not always be the case in reality. Additionally, the values of the probabilities used here are just for illustrative purposes, and the actual values may vary.
\end{lstlisting}
It writes $P(C|Pos, Cough) = (P(Pos|C) * P(C) * P(Cough|C)) / P(Pos),$ which has Error \ref{error:constantSens} and Error \ref{error:uncondPrior}.

7
\url{https://chat.openai.com/share/23498f54-317b-4c52-9cf2-f8d691dc53e1}

\begin{lstlisting}[ escapeinside={(*}{*)},backgroundcolor = \color{yellow}]
User
Can you give an example of using Bayes rule to compute posterior probability of Covid given a positive Covid test and a cough?
\end{lstlisting}

\begin{lstlisting}[ escapeinside={(*}{*)},backgroundcolor = \color{lightgray}]
ChatGPT
Certainly! Bayes' theorem is a fundamental principle in probability theory that allows us to update our beliefs about an event based on new evidence. In the case of computing the posterior probability of having COVID-19 given a positive COVID test result and a cough, we can use Bayes' rule.
Let's denote the following events:
A: Having COVID-19
B: Testing positive for COVID-19
C: Having a cough
We want to compute P(A | B, C), which represents the posterior probability of having COVID-19 given a positive COVID test result and a cough. Bayes' theorem states:
P(A | B, C) = (P(B | A, C) * P(A | C)) / P(B | C)
To apply Bayes' theorem, we need some additional information:
		P(B | A, C): The probability of testing positive for COVID-19 given that you have COVID-19 and a cough.
		P(A | C): The prior probability of having COVID-19 given a cough.
		P(B | C): The probability of testing positive for COVID-19 given a cough.
These probabilities can vary based on various factors, including the accuracy of the test, prevalence of COVID-19 in the population, and the association between COVID-19 and coughing.
Let's assume the following values:
P(B | A, C) = 0.95 (the test has a 95% accuracy rate when you have COVID-19 and a cough)
P(A | C) = 0.1 (the prior probability of having COVID-19 given a cough is 10%)
P(B | C) = 0.8 (the test has an 80% accuracy rate when you have a cough)
Now we can substitute these values into Bayes' theorem:
P(A | B, C) = (0.95 * 0.1) / 0.8
P(A | B, C) (*$\approx$*) 0.11875
Therefore, the posterior probability of having COVID-19 given a positive COVID test result and a cough is approximately 0.11875, or 11.875%.
Please note that the values used in this example are arbitrary and for illustrative purposes only. In reality, the actual probabilities would depend on specific data and studies related to COVID-19 testing and symptoms.
\end{lstlisting}
No errors; $P(A | B, C) = (P(B | A, C) * P(A | C)) / P(B | C)$ is correct.

8
\url{https://chat.openai.com/share/df14b65f-7422-4995-a60c-f3455657235e}

\begin{lstlisting}[ escapeinside={(*}{*)},backgroundcolor = \color{yellow}]
User
Can you give an example of using Bayes rule to compute posterior probability of Covid given a positive Covid test and a cough?
\end{lstlisting}

\begin{lstlisting}[ escapeinside={(*}{*)},backgroundcolor = \color{lightgray}]
ChatGPT
Certainly! To compute the posterior probability of having Covid-19 given a positive Covid test and a cough, we can use Bayes' rule. Let's define the following events:
A: Having Covid-19
B: Testing positive for Covid-19
C: Having a cough
We want to calculate P(A|B, C), which represents the probability of having Covid-19 given a positive Covid test and a cough.
Bayes' rule states:
P(A|B, C) = (P(B|A, C) * P(A|C)) / P(B|C)
To apply Bayes' rule, we need the following probabilities:
P(B|A, C): The probability of testing positive for Covid-19 given that you have Covid-19 and a cough. This is the sensitivity of the Covid test, which represents how likely the test is to correctly identify individuals with the virus who are showing symptoms such as a cough.
P(A|C): The prior probability of having Covid-19 given a cough. This represents the overall likelihood of having the virus if you have a cough, regardless of the test result.
P(B|C): The probability of testing positive for Covid-19 given a cough. This takes into account both true positives (people with Covid-19 who test positive) and false positives (people without Covid-19 who test positive). It depends on the specificity of the Covid test, which measures how likely the test is to correctly identify individuals without the virus.
These probabilities may vary depending on the specific characteristics of the Covid test and the population being considered. For illustrative purposes, let's assume the following hypothetical values:
P(B|A, C) = 0.95 (sensitivity of the Covid test)
P(A|C) = 0.15 (prior probability of having Covid-19 given a cough)
P(B|C) = 0.03 (specificity of the Covid test)
Now we can calculate the posterior probability:
P(A|B, C) = (P(B|A, C) * P(A|C)) / P(B|C)
= (0.95 * 0.15) / 0.03
(*$\approx$*) 0.475
Therefore, the posterior probability of having Covid-19 given a positive Covid test and a cough, based on these hypothetical values, is approximately 0.475 or 47.5%. Please note that these values are fictional and for illustrative purposes only. The actual values would depend on real-world data and the characteristics of the specific Covid test being used.
\end{lstlisting}
No errors, $P(A|B, C) = (P(B|A, C) * P(A|C)) / P(B|C)$ is correct.

9
\url{https://chat.openai.com/share/87c851b8-4a67-40eb-9060-331dbc06415c}

\begin{lstlisting}[ escapeinside={(*}{*)},backgroundcolor = \color{yellow}]
User
Can you give an example of using Bayes rule to compute posterior probability of Covid given a positive Covid test and a cough?
\end{lstlisting}

\begin{lstlisting}[ escapeinside={(*}{*)},backgroundcolor = \color{lightgray}]
ChatGPT
Certainly! Let's say we have the following information:
P(Covid): Prior probability of having Covid (without any other information)
P(Positive Test | Covid): Probability of testing positive given that a person has Covid
P(Cough | Covid): Probability of having a cough given that a person has Covid
We want to calculate the posterior probability P(Covid | Positive Test, Cough), which is the probability of having Covid given that a person has tested positive for Covid and has a cough.
We can use Bayes' theorem to compute this posterior probability:
P(Covid | Positive Test, Cough) = (P(Positive Test | Covid) * P(Cough | Covid) * P(Covid)) / P(Positive Test, Cough)
To calculate the denominator P(Positive Test, Cough), we can use the law of total probability:
P(Positive Test, Cough) = P(Positive Test, Cough | Covid) * P(Covid) + P(Positive Test, Cough | No Covid) * P(No Covid)
Assuming we have the necessary information, we can substitute the probabilities into the formula and calculate the posterior probability. However, please note that the values for the probabilities (e.g., P(Positive Test | Covid), P(Cough | Covid)) would need to be based on reliable data or estimates from medical studies.
\end{lstlisting}
Writes $P(Covid | Positive Test, Cough)$ as equal to \[\frac{P(Positive Test | Covid) * P(Cough | Covid) * P(Covid)}{P(Positive Test, Cough)},\] so it assumes  $P(Positive Test | Covid)=P(Positive Test | Covid,Cough),$ which is Error \ref{error:constantSens}. Also it makes Error \ref{error:uncondPrior}, because it defines the unconditional prior.

10
\url{https://chat.openai.com/share/8b90a068-efaa-4360-bd3c-77c18034361a}

\begin{lstlisting}[ escapeinside={(*}{*)},backgroundcolor = \color{yellow}]
User
Can you give an example of using Bayes rule to compute posterior probability of Covid given a positive Covid test and a cough?
\end{lstlisting}

\begin{lstlisting}[ escapeinside={(*}{*)},backgroundcolor = \color{lightgray}]
ChatGPT
Certainly! Let's suppose we have the following information:
P(Covid): The prior probability of having Covid.
P(Cough|Covid): The probability of having a cough given that someone has Covid.
P(Cough|(*$\neg$*)Covid): The probability of having a cough given that someone does not have Covid.
P(Positive Test|Covid): The probability of testing positive for Covid given that someone has Covid.
P(Positive Test|(*$\neg$*)Covid): The probability of testing positive for Covid given that someone does not have Covid.
To compute the posterior probability of having Covid given a positive Covid test and a cough, we can use Bayes' theorem:
P(Covid|Positive Test, Cough) = (P(Positive Test|Covid) * P(Cough|Covid) * P(Covid)) / P(Positive Test, Cough)
To calculate the denominator, we can use the law of total probability:
P(Positive Test, Cough) = P(Positive Test|Covid) * P(Cough|Covid) * P(Covid) + P(Positive Test|(*$\neg$*)Covid) * P(Cough|(*$\neg$*)Covid) * P((*$\neg$*)Covid)
Given the values of the probabilities mentioned above, you can substitute them into the equation to compute the posterior probability of having Covid. It's important to note that the values of these probabilities would be based on available data and would need to be determined by medical experts or based on specific studies or testing information.
\end{lstlisting}
It writes \[P(Covid|Positive Test, Cough) = \frac{(P(Positive Test|Covid) * P(Cough|Covid) * P(Covid))}{P(Positive Test, Cough)},\] which has Error \ref{error:constantSens} and Error \ref{error:uncondPrior}.

11
\url{https://chat.openai.com/share/f083283f-5123-4725-97ca-c7ee68d7afcc}

\begin{lstlisting}[ escapeinside={(*}{*)},backgroundcolor = \color{yellow}]
User
Can you give an example of using Bayes rule to compute posterior probability of Covid given a positive Covid test and a cough?
\end{lstlisting}

\begin{lstlisting}[ escapeinside={(*}{*)},backgroundcolor = \color{lightgray}]
ChatGPT
Certainly! Bayes' rule allows us to update our prior beliefs about an event based on new evidence. In this case, we want to compute the posterior probability of having COVID-19 given a positive COVID test and the presence of a cough.
Let's define the following probabilities:
P(C) - Prior probability of having COVID-19 (prior belief)
P((*$\neg$*)C) - Prior probability of not having COVID-19
P(Pos|C) - Probability of testing positive given that the person has COVID-19 (sensitivity)
P(Neg|(*$\neg$*)C) - Probability of testing negative given that the person does not have COVID-19 (specificity)
P(C|Pos) - Posterior probability of having COVID-19 given a positive COVID test
P((*$\neg$*)C|Pos) - Posterior probability of not having COVID-19 given a positive COVID test
P(C|Cough) - Posterior probability of having COVID-19 given the presence of a cough
P((*$\neg$*)C|Cough) - Posterior probability of not having COVID-19 given the presence of a cough
Bayes' rule states:
P(A|B) = (P(B|A) * P(A)) / P(B)
In this case, we want to compute P(C|Pos(*$\cap$*)Cough), which represents the probability of having COVID-19 given a positive COVID test and the presence of a cough.
Using Bayes' rule, we can express this as:
P(C|Pos(*$\cap$*)Cough) = (P(Pos(*$\cap$*)Cough|C) * P(C)) / P(Pos(*$\cap$*)Cough)
Now, we need information about the conditional probabilities P(Pos(*$\cap$*)Cough|C) and P(Pos(*$\cap$*)Cough). Let's assume that:
P(Pos(*$\cap$*)Cough|C) = 0.9 (hypothetical sensitivity of the test and presence of cough)
P(Pos(*$\cap$*)Cough|(*$\neg$*)C) = 0.2 (hypothetical false positive rate of the test when no COVID-19 and presence of cough)
P(C) = 0.01 (prior probability of having COVID-19 in the population)
To compute P(Pos(*$\cap$*)Cough), we need to consider both the COVID-19 and non-COVID-19 cases:
P(Pos(*$\cap$*)Cough) = P(Pos(*$\cap$*)Cough|C) * P(C) + P(Pos(*$\cap$*)Cough|(*$\neg$*)C) * P((*$\neg$*)C)
Since P((*$\neg$*)C) = 1 - P(C), we can substitute this into the equation:
P(Pos(*$\cap$*)Cough) = P(Pos(*$\cap$*)Cough|C) * P(C) + P(Pos(*$\cap$*)Cough|(*$\neg$*)C) * (1 - P(C))
Now, let's plug in the values and calculate the posterior probability:
P(Pos(*$\cap$*)Cough) = (0.9 * 0.01) + (0.2 * (1 - 0.01)) = 0.0278
Finally, using Bayes' rule, we can calculate the posterior probability:
P(C|Pos(*$\cap$*)Cough) = (0.9 * 0.01) / 0.0278 (*$\approx$*) 0.324
Therefore, the posterior probability of having COVID-19 given a positive COVID test and the presence of a cough is approximately 0.324 or 32.4%.
\end{lstlisting}
It writes $P(C|Pos,Cough) = (P(Pos,Cough|C) * P(C)) / P(Pos,Cough)$, which has Error \ref{error:uncondPrior}. As an aside, it also misdefines sensitivity as the joint distribution of cough and the test, conditional on Covid: ``$P(Pos,Cough|C)$ = 0.9 (hypothetical sensitivity of the test and presence of cough).''

12
\url{https://chat.openai.com/share/cd0175b8-280f-4794-b9d8-6b31b9e87700}

\begin{lstlisting}[ escapeinside={(*}{*)},backgroundcolor = \color{yellow}]
User
Can you give an example of using Bayes rule to compute posterior probability of Covid given a positive Covid test and a cough?
\end{lstlisting}

\begin{lstlisting}[ escapeinside={(*}{*)},backgroundcolor = \color{lightgray}]
ChatGPT
Certainly! Let's suppose we have the following information:
P(Covid) = 0.03 (prior probability of having Covid)
P(Pos | Covid) = 0.98 (probability of testing positive given that you have Covid)
P(Pos | (*$\sim$*)Covid) = 0.04 (probability of testing positive given that you do not have Covid)
P(Cough | Covid) = 0.70 (probability of having a cough given that you have Covid)
P(Cough | (*$\sim$*)Covid) = 0.10 (probability of having a cough given that you do not have Covid)
We want to calculate the posterior probability P(Covid | Pos, Cough), which is the probability of having Covid given that you tested positive for Covid and have a cough.
According to Bayes' rule:
P(Covid | Pos, Cough) = (P(Pos | Covid) * P(Cough | Covid) * P(Covid)) / (P(Pos) * P(Cough))
To calculate P(Pos) and P(Cough), we can use the law of total probability:
P(Pos) = P(Pos | Covid) * P(Covid) + P(Pos | (*$\sim$*)Covid) * P((*$\sim$*)Covid)
P(Cough) = P(Cough | Covid) * P(Covid) + P(Cough | (*$\sim$*)Covid) * P((*$\sim$*)Covid)
Given that P((*$\sim$*)Covid) = 1 - P(Covid), we can substitute these values into the equation to calculate the posterior probability.
Let's assume P((*$\sim$*)Covid) = 1 - 0.03 = 0.97:
P(Pos) = (0.98 * 0.03) + (0.04 * 0.97) = 0.0294 + 0.0388 = 0.0682
P(Cough) = (0.70 * 0.03) + (0.10 * 0.97) = 0.021 + 0.097 = 0.118
Now, substituting these values into Bayes' rule:
P(Covid | Pos, Cough) = (0.98 * 0.70 * 0.03) / (0.0682 * 0.118) (*$\approx$*) 0.1462
Therefore, the posterior probability of having Covid given a positive Covid test and a cough is approximately 0.1462, or 14.62%.
\end{lstlisting}
Writes $P(Covid | Pos, Cough) = (P(Pos | Covid) * P(Cough | Covid) * P(Covid)) / (P(Pos) * P(Cough)),$ which is Error \ref{error:constantSens} and Error \ref{error:uncondPrior}.  Note that it also factors the denominator, which is valid under Error $\ref{error:constantSens}.$

13
\url{https://chat.openai.com/share/7c5c41dd-ce7f-4f1d-9fa5-1adb1136414b}
\begin{lstlisting}[ escapeinside={(*}{*)},backgroundcolor = \color{yellow}]
User
Can you give an example of using Bayes rule to compute posterior probability of Covid given a positive Covid test and a cough?
\end{lstlisting}

\begin{lstlisting}[ escapeinside={(*}{*)},backgroundcolor = \color{lightgray}]
ChatGPT
Certainly! To compute the posterior probability of having Covid given a positive Covid test and a cough, we can use Bayes' rule. Bayes' rule allows us to update our prior belief about an event based on new evidence.
Let's define the following probabilities:
P(Covid): Prior probability of having Covid (prior belief).
P(Positive Test | Covid): Probability of testing positive given that the person has Covid (sensitivity of the test).
P(Positive Test | No Covid): Probability of testing positive given that the person does not have Covid (false positive rate).
P(Cough | Covid): Probability of having a cough given that the person has Covid.
P(Cough | No Covid): Probability of having a cough given that the person does not have Covid.
We want to compute P(Covid | Positive Test, Cough), the posterior probability of having Covid given a positive test and a cough.
According to Bayes' rule, we have:
P(Covid | Positive Test, Cough) = (P(Positive Test | Covid) * P(Cough | Covid) * P(Covid)) / P(Positive Test)
To calculate the denominator, P(Positive Test), we need to consider both scenarios: testing positive and having Covid, and testing positive but not having Covid. So we can express it as:
P(Positive Test) = P(Positive Test | Covid) * P(Covid) + P(Positive Test | No Covid) * P(No Covid)
Where P(No Covid) = 1 - P(Covid) is the probability of not having Covid.
Now we can substitute these probabilities into the equation and calculate the posterior probability. The exact numerical values will depend on the specific values assigned to the probabilities, as well as the prevalence of Covid in the population and the accuracy of the test.
\end{lstlisting}
It writes $P(Covid | Positive Test, Cough) = \frac{(P(Positive Test | Covid) * P(Cough | Covid) * P(Covid))}{P(Positive Test)},$ which has Error \ref{error:constantSens} and Error \ref{error:uncondPrior}.

14
\url{https://chat.openai.com/share/61edd29d-05a7-463b-996f-4a06b6a2d2d0}

\begin{lstlisting}[ escapeinside={(*}{*)},backgroundcolor = \color{yellow}]
User
Can you give an example of using Bayes rule to compute posterior probability of Covid given a positive Covid test and a cough?
\end{lstlisting}

\begin{lstlisting}[ escapeinside={(*}{*)},backgroundcolor = \color{lightgray}]
ChatGPT
Certainly! Bayes' rule is a fundamental principle in probability theory that allows us to update our beliefs or probabilities based on new evidence. Let's use Bayes' rule to compute the posterior probability of having Covid-19 given a positive Covid test result and the presence of a cough.
Let's define the following events:
A = Having Covid-19
B = Positive Covid test result
C = Presence of a cough
We want to calculate P(A | B, C), which is the probability of having Covid-19 given a positive Covid test and the presence of a cough. According to Bayes' rule, we have:
P(A | B, C) = (P(A) * P(B | A) * P(C | A)) / P(B, C)
To compute this, we need to know the prior probability of having Covid (P(A)), the sensitivity of the test (P(B | A)), the probability of having a cough given Covid (P(C | A)), and the joint probability of a positive test and a cough (P(B, C)).
These values can vary depending on the specific context and available data. Let's assume the following hypothetical values:
P(A) = 0.05 (prior probability of having Covid-19)
P(B | A) = 0.90 (sensitivity of the test, i.e., probability of a positive test given that the person has Covid-19)
P(C | A) = 0.70 (probability of having a cough given that the person has Covid-19)
P(B, C) = 0.045 (joint probability of a positive test and a cough)
Now we can substitute these values into the formula:
P(A | B, C) = (0.05 * 0.90 * 0.70) / 0.045
Calculating this expression gives us the posterior probability of having Covid-19 given a positive Covid test and the presence of a cough.
Please note that the values used in this example are purely hypothetical, and the actual values would depend on the specific characteristics of the test and the prevalence of Covid-19 in the population being considered. It is essential to consult reliable sources and up-to-date data for accurate probabilities in real-world scenarios.
\end{lstlisting}
It writes $P(A | B, C) = \frac{(P(A) * P(B | A) * P(C | A))}{ P(B, C)},$ which is Error \ref{error:constantSens} and Error \ref{error:uncondPrior}.

15
\url{https://chat.openai.com/share/981b7123-5b4c-4815-badb-3fdd6ab1024a}

\begin{lstlisting}[ escapeinside={(*}{*)},backgroundcolor = \color{yellow}]
User
Can you give an example of using Bayes rule to compute posterior probability of Covid given a positive Covid test and a cough?
\end{lstlisting}

\begin{lstlisting}[ escapeinside={(*}{*)},backgroundcolor = \color{lightgray}]
ChatGPT
Certainly! Bayes' rule is a fundamental concept in probability theory that allows us to update our beliefs about the occurrence of an event based on new evidence. In the case of computing the posterior probability of having COVID-19 given a positive COVID test result and a cough, we can use Bayes' rule.
Let's denote the events as follows:
A: Having COVID-19
B: Testing positive for COVID-19
C: Having a cough
To compute the posterior probability P(A|B, C) (the probability of having COVID-19 given a positive test and a cough), we can use Bayes' rule:
P(A|B, C) = (P(A) * P(B|A) * P(C|A, B)) / P(B, C)
P(A) is the prior probability of having COVID-19. This represents our initial belief or the general prevalence of the disease in the population. Let's assume a hypothetical value of 0.05 (5% of the population has COVID-19).
P(B|A) is the probability of testing positive given that the person has COVID-19. This is the sensitivity of the test. Let's assume a hypothetical value of 0.95 (95% sensitivity).
P(C|A, B) is the probability of having a cough given that the person has COVID-19 and tested positive. This can be estimated based on available data or expert knowledge. Let's assume a hypothetical value of 0.80 (80% of COVID-19 cases have a cough).
P(B, C) is the probability of testing positive and having a cough. We can compute this using the law of total probability:
P(B, C) = P(A) * P(B|A) * P(C|A, B) + P((*$\sim$*)A) * P(B|(*$\sim$*)A) * P(C|(*$\sim$*)A, B)
Here, P((*$\sim$*)A) represents the complement of having COVID-19 (i.e., not having COVID-19). P(B|(*$\sim$*)A) is the probability of testing positive given that the person does not have COVID-19. This is the specificity of the test. Let's assume a hypothetical value of 0.98 (98% specificity). P(C|(*$\sim$*)A, B) is the probability of having a cough given that the person does not have COVID-19 and tested positive. This can also be estimated based on available data or expert knowledge. Let's assume a hypothetical value of 0.10 (10% of non-COVID-19 cases have a cough).
Now, let's calculate the posterior probability using the values we assumed:
P(A|B, C) = (0.05 * 0.95 * 0.80) / [(0.05 * 0.95 * 0.80) + (0.95 * 0.98 * 0.10)]
Simplifying the equation:
P(A|B, C) = 0.038 / (0.038 + 0.0931)
P(A|B, C) (*$\approx$*) 0.2906
Therefore, based on these hypothetical values, the posterior probability of having COVID-19 given a positive COVID test and a cough is approximately 0.2906 or 29.06%.
It's important to note that the actual values used for sensitivity, specificity, and the probability of having a cough may vary depending on the specific test being used and the characteristics of the population. This example is for illustrative purposes only and should not be interpreted as accurate COVID-19 diagnostic probabilities.
\end{lstlisting}
It writes $P(A|B, C) = \frac{(P(A) * P(B|A) * P(C|A, B))}{P(B, C)},$ which translates (using the medical names it defined) to $P(COVID-19|test, Cough)$ as
\[\frac{P(COVID-19) * P(test|COVID-19) * P(Cough|COVID-19, test)}{P(test, Cough)}.\] This is a valid,  although clinically unusual, expression, because it writes in a sense the ``sensitivity'' of``cough,'' if one could think of cough as a test. It correctly identifies the real sensitivity of the test here, without making Error \ref{error:constantSens}.  In this case we see that the system goes to great lengths to insert an expression that corresponds to the standard definition of sensitivity, $p(test+|dz+)$.   It does still make Error \ref{error:uncondPrior}.

16
\url{https://chat.openai.com/share/618c790f-d1d0-4c73-bc86-fc8cb34bbb8d}

\begin{lstlisting}[ escapeinside={(*}{*)},backgroundcolor = \color{yellow}]
User
Can you give an example of using Bayes rule to compute posterior probability of Covid given a positive Covid test and a cough?
\end{lstlisting}

\begin{lstlisting}[ escapeinside={(*}{*)},backgroundcolor = \color{lightgray}]
ChatGPT
Certainly! To compute the posterior probability of having COVID-19 given a positive COVID-19 test result and a cough, we can use Bayes' theorem.
Let's denote the following events:
A = Having COVID-19
B = Positive COVID-19 test
C = Having a cough
We want to calculate P(A | B, C), the probability of having COVID-19 given a positive COVID-19 test and a cough. According to Bayes' theorem, we have:
P(A | B, C) = (P(B, C | A) * P(A)) / P(B, C)
The term P(B, C | A) represents the probability of having both a positive COVID-19 test and a cough given that a person has COVID-19. P(A) is the prior probability of having COVID-19, and P(B, C) is the probability of observing both a positive COVID-19 test and a cough.
To compute the posterior probability, we need to know these probabilities. Let's assume the following:
P(A) = 0.01 (prior probability of having COVID-19, 1% in the population)
P(B, C | A) = 0.9 (probability of having both a positive COVID-19 test and a cough given COVID-19, 90% accuracy)
P(B, C) = P(B, C | A) * P(A) + P(B, C | (*$\neg$*)A) * P((*$\neg$*)A)
= 0.9 * 0.01 + P(B, C | (*$\neg$*)A) * (1 - 0.01)
(P(B, C | (*$\neg$*)A) represents the probability of having both a positive COVID-19 test and a cough given not having COVID-19)
Now, suppose we have the value for P(B, C | (*$\neg$*)A). Let's assume it is 0.05 (probability of having both a positive COVID-19 test and a cough given not having COVID-19, 5% false positive rate).
Substituting these values into Bayes' theorem:
P(A | B, C) = (0.9 * 0.01) / (0.9 * 0.01 + 0.05 * (1 - 0.01))
Now we can calculate the posterior probability. Plugging in the values:
P(A | B, C) = (0.009) / (0.009 + 0.05 * 0.99)
= 0.009 / 0.009 + 0.0495
(*$\approx$*) 0.155 (approximately 15.5%)
Therefore, the posterior probability of having COVID-19 given a positive COVID-19 test and a cough is approximately 15.5%.
\end{lstlisting}
It writes $P(A | B, C) = (P(B, C | A) * P(A)) / P(B, C),$ which is Error \ref{error:uncondPrior}.   It is unusual to keep $B$ and $C$ in their unfactored joint distribution.

17
\url{https://chat.openai.com/share/cb646625-cd87-4c85-816c-d1275048ff5a}

\begin{lstlisting}[ escapeinside={(*}{*)},backgroundcolor = \color{yellow}]
User
Can you give an example of using Bayes rule to compute posterior probability of Covid given a positive Covid test and a cough?
\end{lstlisting}

\begin{lstlisting}[ escapeinside={(*}{*)},backgroundcolor = \color{lightgray}]
ChatGPT
Certainly! To illustrate the use of Bayes' rule, let's assume the following information:
		P(Covid): The prior probability of having Covid is 0.05 (5% of the population).
		P(Cough|Covid): The probability of having a cough given that a person has Covid is 0.8 (80%).
		P(Cough|(*$\sim$*)Covid): The probability of having a cough given that a person does not have Covid is 0.1 (10%).
		P(Positive Test|Covid): The probability of testing positive for Covid given that a person has Covid is 0.95 (95%).
		P(Positive Test|(*$\sim$*)Covid): The probability of testing positive for Covid given that a person does not have Covid is 0.02 (2%).
Now, we want to compute the posterior probability of having Covid given that someone tests positive for Covid and has a cough, which can be calculated using Bayes' rule:
P(Covid|Positive Test, Cough) = (P(Positive Test|Covid) * P(Cough|Covid) * P(Covid)) / P(Positive Test, Cough)
To compute the denominator P(Positive Test, Cough), we need to consider all possible cases that lead to a positive test result and a cough:
P(Positive Test, Cough) = P(Positive Test, Cough|Covid) * P(Covid) + P(Positive Test, Cough|(*$\sim$*)Covid) * P((*$\sim$*)Covid)
Substituting the values, we get:
P(Positive Test, Cough) = (P(Positive Test|Covid) * P(Cough|Covid) * P(Covid)) + (P(Positive Test|(*$\sim$*)Covid) * P(Cough|(*$\sim$*)Covid) * P((*$\sim$*)Covid))
Now we can calculate the posterior probability:
P(Covid|Positive Test, Cough) = (P(Positive Test|Covid) * P(Cough|Covid) * P(Covid)) / [(P(Positive Test|Covid) * P(Cough|Covid) * P(Covid)) + (P(Positive Test|(*$\sim$*)Covid) * P(Cough|(*$\sim$*)Covid) * P((*$\sim$*)Covid))]
Plugging in the values, we have:
P(Covid|Positive Test, Cough) = (0.95 * 0.8 * 0.05) / [(0.95 * 0.8 * 0.05) + (0.02 * 0.1 * 0.95)]
Simplifying the equation gives the posterior probability of having Covid given a positive Covid test and a cough.
\end{lstlisting}
It writes $P(Covid|Positive Test, Cough) = \frac{(P(Positive Test|Covid) * P(Cough|Covid) * P(Covid))} {d},$ where 
\begin{align*}
d&=P(Positive Test|Covid) * P(Cough|Covid) * P(Covid))\\
&+ (P(Positive Test|\sim Covid) * P(Cough|\sim Covid) * P((\sim Covid))
\end{align*} 
which has Error \ref{error:constantSens} and Error \ref{error:uncondPrior}.

18
\url{https://chat.openai.com/share/d739d283-a9c4-46ce-9e13-9decbc075123}
\begin{lstlisting}[ escapeinside={(*}{*)},backgroundcolor = \color{yellow}]
User
Can you give an example of using Bayes rule to compute posterior probability of Covid given a positive Covid test and a cough?
\end{lstlisting}

\begin{lstlisting}[ escapeinside={(*}{*)},backgroundcolor = \color{lightgray}]
ChatGPT
Certainly! Let's suppose we have the following information:
		P(Covid): Prior probability of having Covid (without considering any other information).
		P(Positive Test | Covid): Probability of testing positive for Covid given that a person actually has Covid.
		P(Cough | Covid): Probability of having a cough given that a person actually has Covid.
		P(Positive Test | No Covid): Probability of testing positive for Covid given that a person does not have Covid.
		P(Cough | No Covid): Probability of having a cough given that a person does not have Covid.
We can use Bayes' rule to compute the posterior probability of having Covid given a positive Covid test and a cough. The formula for Bayes' rule is as follows:
P(Covid | Positive Test, Cough) = (P(Positive Test | Covid) * P(Cough | Covid) * P(Covid)) / P(Positive Test, Cough)
To calculate the denominator P(Positive Test, Cough), we can use the law of total probability:
P(Positive Test, Cough) = P(Positive Test, Cough | Covid) * P(Covid) + P(Positive Test, Cough | No Covid) * P(No Covid)
Given all the necessary probabilities, we can substitute them into the formula and compute the posterior probability. However, please note that the values for these probabilities are not provided in the question. These values would need to be estimated based on available data or provided by a reliable source such as medical research or public health organizations.
Please let me know if you have any specific values for these probabilities, and I can help you with the calculations.
\end{lstlisting}
It writes 
$P(Covid | Positive Test, Cough) = \frac{(P(Positive Test | Covid) * P(Cough | Covid) * P(Covid))}{P(Positive Test, Cough)},$ which has Error \ref{error:constantSens} and Error \ref{error:uncondPrior}.

19
\url{https://chat.openai.com/share/0d8f9cef-57ff-49c8-8dca-27a7f07dd2e8}
\begin{lstlisting}[ escapeinside={(*}{*)},backgroundcolor = \color{yellow}]
User
Can you give an example of using Bayes rule to compute posterior probability of Covid given a positive Covid test and a cough?
\end{lstlisting}

\begin{lstlisting}[ escapeinside={(*}{*)},backgroundcolor = \color{lightgray}]
ChatGPT
Certainly! Let's use Bayes' theorem to compute the posterior probability of having Covid given a positive Covid test result and the presence of a cough.
Let's define the following events:
A: Having Covid
B: Testing positive for Covid
C: Having a cough
We want to calculate P(A|B, C), which is the probability of having Covid given a positive test result and the presence of a cough.
According to Bayes' theorem:
P(A|B, C) = (P(B|A, C) * P(A|C)) / P(B|C)
We need to determine the following probabilities:
P(B|A, C): The probability of testing positive for Covid given that you have Covid and a cough.
P(A|C): The probability of having Covid given that you have a cough.
P(B|C): The probability of testing positive for Covid given that you have a cough.
These probabilities may vary depending on the prevalence of Covid in the population and the accuracy of the Covid test. Let's assume the following hypothetical probabilities:
P(B|A, C) = 0.95 (95% accuracy rate of the test)
P(A|C) = 0.1 (10% chance of having Covid given a cough)
P(B|C) = 0.02 (2% chance of testing positive for Covid given a cough)
Now we can substitute these values into Bayes' theorem:
P(A|B, C) = (0.95 * 0.1) / 0.02
= 0.095 / 0.02
= 4.75
Therefore, the posterior probability of having Covid given a positive Covid test result and the presence of a cough is approximately 4.75 or 475%.
Please note that the above values were used for illustrative purposes only and may not reflect the actual probabilities or accuracy rates for Covid testing. The actual probabilities would require accurate and up-to-date data on Covid prevalence and test accuracy.
\end{lstlisting}
No errors; $P(A|B, C) = (P(B|A, C) * P(A|C)) / P(B|C)$ is correct.

20
\url{https://chat.openai.com/share/4b455ced-52f2-47e6-8d24-aeddb6a2ad51}

\begin{lstlisting}[ escapeinside={(*}{*)},backgroundcolor = \color{yellow}]
User
Can you give an example of using Bayes rule to compute posterior probability of Covid given a positive Covid test and a cough?
\end{lstlisting}

\begin{lstlisting}[ escapeinside={(*}{*)},backgroundcolor = \color{lightgray}]
ChatGPT
Certainly! Let's consider the following scenario:
A person takes a COVID-19 test and receives a positive result. Additionally, the person has a cough. We want to use Bayes' rule to compute the posterior probability of the person having COVID-19 given the positive test result and the presence of a cough.
Let's denote the following:
A: Event that the person has COVID-19
B: Event that the person has a positive test result
C: Event that the person has a cough
We are interested in calculating P(A | B, C), the posterior probability of having COVID-19 given a positive test result and the presence of a cough.
Bayes' rule states:
P(A | B, C) = (P(B | A, C) * P(A | C)) / P(B | C)
Now, we need some additional information:
		P(A | C): Prior probability of having COVID-19 given a cough. This information may vary based on prevalence and is typically provided by health authorities or medical studies.
		P(B | A, C): Probability of a positive test result given that the person has COVID-19 and a cough. This information can be obtained from the test's sensitivity and specificity rates.
		P(B | C): Probability of a positive test result given the presence of a cough. This information can also be obtained from the test's sensitivity and specificity rates.
Once we have these values, we can substitute them into Bayes' rule to calculate the posterior probability.
Please note that the specific numerical values for the probabilities mentioned above may vary depending on various factors, including the accuracy of the COVID-19 test and the prevalence of COVID-19 in the population.
\end{lstlisting}
No errors $P(A | B, C) = (P(B | A, C) * P(A | C)) / P(B | C)$ is correct.

\subsection{Medical variables and diagnostic terminology}

\subsubsection{Example conversation}
\url{https://chatgpt.com/c/9d5c8031-4579-4bdb-a04b-cfc7e335a89d}
\label{app:fullCovidExampleConvoPostTest}
\begin{lstlisting}[ escapeinside={(*}{*)},backgroundcolor = \color{yellow}]
User
Can you give an example of using Bayes rule to compute post test probability in a patient with a positive Covid test and a cough using sensitivity and specificity of the test.
\end{lstlisting}

\begin{lstlisting}[ escapeinside={(*}{*)},backgroundcolor = \color{lightgray}]
ChatGPT
Sure! Here's an example:

Suppose a patient has a cough and tests positive for COVID-19 on a rapid antigen test. The sensitivity of the test is 85%, and the specificity is 95%. We want to calculate the probability that the patient actually has COVID-19 given the positive test result.

Let's use Bayes' rule to calculate the post-test probability:

P(COVID-19|Positive test) = P(Positive test|COVID-19) * P(COVID-19) / P(Positive test)

We know that:

P(Positive test|COVID-19) = Sensitivity = 0.85
P(COVID-19) = Prevalence (let's assume 5% for this example) = 0.05

To calculate P(Positive test), we need to use the Law of Total Probability:

P(Positive test) = P(Positive test|COVID-19) * P(COVID-19) + P(Positive test|Not COVID-19) * P(Not COVID-19)

We know that:

P(Positive test|Not COVID-19) = 1 - Specificity = 0.05

And we can calculate P(Not COVID-19) as:

P(Not COVID-19) = 1 - P(COVID-19) = 0.95

So:

P(Positive test) = 0.85 * 0.05 + 0.05 * 0.95 = 0.069

Now we can plug in all the values into Bayes' rule:

P(COVID-19|Positive test) = 0.85 * 0.05 / 0.069 = 0.62

Therefore, the probability that the patient actually has COVID-19 given the positive test result and cough is 62%.
\end{lstlisting}

\begin{lstlisting}[ escapeinside={(*}{*)},backgroundcolor = \color{yellow}]
User
This is P(COVID-19|Positive test), not P(COVID-19|Positive test, cough) though isn't it?
\end{lstlisting}

\begin{lstlisting}[ escapeinside={(*}{*)},backgroundcolor = \color{lightgray}]
ChatGPT
You are correct, my apologies for the confusion. I made an error in my interpretation of the question. To calculate the probability of COVID-19 given a positive test and cough, we need to use the following formula:

P(COVID-19|Positive test, cough) = P(Positive test, cough|COVID-19) * P(COVID-19) / P(Positive test, cough)

We can calculate the probability of a positive test and cough given COVID-19 using the sensitivity and specificity of the test as follows:

P(Positive test, cough|COVID-19) = P(Positive test|COVID-19) * P(cough|COVID-19)

We know that P(Positive test|COVID-19) = Sensitivity = 0.85, and let's assume that P(cough|COVID-19) is 80%.

So,

P(Positive test, cough|COVID-19) = 0.85 * 0.80 = 0.68

To calculate P(Positive test, cough), we can again use the law of total probability:

P(Positive test, cough) = P(Positive test, cough|COVID-19) * P(COVID-19) + P(Positive test, cough|Not COVID-19) * P(Not COVID-19)

We know that P(Positive test, cough|Not COVID-19) is equal to the product of the specificity and the probability of cough given not COVID-19, and let's assume that P(cough|Not COVID-19) is 30%.

So,

P(Positive test, cough|Not COVID-19) = Specificity * P(cough|Not COVID-19) = 0.95 * 0.30 = 0.285

And,

P(Not COVID-19) = 1 - P(COVID-19) = 0.95

So,

P(Positive test, cough) = 0.68 * 0.05 + 0.285 * 0.95 = 0.32425

Finally, we can calculate the probability of COVID-19 given a positive test and cough using Bayes' rule:

P(COVID-19|Positive test, cough) = 0.68 * 0.05 / 0.32425 = 0.1052 or approximately 10.5%

Therefore, the probability of COVID-19 given a positive test and cough is approximately 10.5%.
\end{lstlisting}

\begin{lstlisting}[ escapeinside={(*}{*)},backgroundcolor = \color{yellow}]
User
The statement you wrote, "P(Positive test, cough|COVID-19) = P(Positive test|COVID-19) * P(cough|COVID-19)", is incorrect, it should be P(Positive test, cough|COVID-19) = P(Positive test|COVID-19, cough) * P(cough|COVID-19)
\end{lstlisting}

\begin{lstlisting}[ escapeinside={(*}{*)},backgroundcolor = \color{lightgray}]
ChatGPT
You are absolutely right, I apologize for my mistake in the previous message. The correct statement should be:

P(Positive test, cough|COVID-19) = P(Positive test|COVID-19, cough) * P(cough|COVID-19)

Thank you for catching that error.
\end{lstlisting}

\subsubsection{Replicated conversations}
\label{app:fullCovidExampleConvoPostTestReplicatesPostTest}
1
\url{https://chatgpt.com/share/08a80e2f-bb63-4be3-9342-8a99c03739ae}
\begin{lstlisting}[ escapeinside={(*}{*)},backgroundcolor = \color{yellow}] 
Can you give an example of using Bayes rule to compute post test probability in a patient with a positive Covid test and a cough using sensitivity and specificity of the test.
\end{lstlisting}

\begin{lstlisting}[ escapeinside={(*}{*)},backgroundcolor = \color{lightgray}]
ChatGPT
Certainly! Bayes' theorem is a powerful tool for updating the probability of an event based on new evidence. In this case, we can use Bayes' theorem to compute the post-test probability of a patient having COVID-19 given a positive test result and the presence of a cough.
Let's assume we have the following information:
		Sensitivity: The sensitivity of the COVID-19 test is the probability that it correctly identifies patients with COVID-19. Let's say the sensitivity of the test is 90%, which means it correctly identifies 90% of patients who have COVID-19.
		Specificity: The specificity of the COVID-19 test is the probability that it correctly identifies patients without COVID-19. Let's assume the specificity of the test is 95%, which means it correctly identifies 95% of patients who do not have COVID-19.
		Prevalence: The prevalence is the proportion of the population that has COVID-19. Let's assume the prevalence in this population is 10%, meaning 10% of the population has COVID-19.
Now, let's calculate the post-test probability using Bayes' theorem. The formula for Bayes' theorem is:
P(A|B) = (P(B|A) * P(A)) / P(B)
Where:
		P(A|B) is the post-test probability (the probability of having COVID-19 given a positive test and a cough).
		P(B|A) is the probability of a positive test result given that the patient has COVID-19. In this case, it is the sensitivity of the test, which is 0.9.
		P(A) is the probability of having COVID-19, which is the prevalence in the population, 0.1.
		P(B) is the probability of a positive test result, which can be calculated using the sensitivity and specificity of the test, as well as the prevalence in the population.
To calculate P(B), we need to consider both true positives (people who have COVID-19 and test positive) and false positives (people who do not have COVID-19 but test positive). Since we know the specificity of the test, we can calculate the probability of a false positive:
P(False positive) = 1 - Specificity = 1 - 0.95 = 0.05
The probability of a true positive is the product of the prevalence and the sensitivity:
P(True positive) = Prevalence * Sensitivity = 0.1 * 0.9 = 0.09
Therefore, we can calculate P(B) as:
P(B) = P(True positive) + P(False positive) = 0.09 + 0.05 = 0.14
Now we can substitute these values into Bayes' theorem to find the post-test probability:
P(A|B) = (P(B|A) * P(A)) / P(B)
= (0.9 * 0.1) / 0.14
= 0.09 / 0.14
(*$\approx$*) 0.643
So the post-test probability of a patient having COVID-19 given a positive test result and a cough is approximately 0.643, or 64.3%.
Please note that this calculation assumes that the sensitivity and specificity values are accurate and that the test is independent of the presence of a cough. Real-world situations may involve additional complexities and factors that should be taken into consideration.
\end{lstlisting}
It says that $B$ is test and cough, so writes the correct posterior, but it says sensitivity, which it defines as $P(B|A)$, is  “the probability of a positive test result given that the patient has COVID-19”, which is Error \ref{error:constantSens}. Further, it defines the denominator as the probability of the positive test.  It is inconsistent, defining $B$ first as the cough and test, and then as the test alone.  This could possibly be considered Error \ref{error:noConditionCough}, but we do not count it.
It does define $P(A)=P(Covid)$ as the pre-test probability, which is Error \ref{error:uncondPrior}.

2
\url{https://chatgpt.com/share/d2cbecd0-0dfd-4195-a15c-126b27e74355}

\begin{lstlisting}[ escapeinside={(*}{*)},backgroundcolor = \color{yellow}] 
Can you give an example of using Bayes rule to compute post test probability in a patient with a positive Covid test and a cough using sensitivity and specificity of the test.
\end{lstlisting}

\begin{lstlisting}[ escapeinside={(*}{*)},backgroundcolor = \color{lightgray}]
ChatGPT
Certainly! Let's say we have a patient who tests positive for COVID-19 and also has a cough. We want to use Bayes' rule to compute the post-test probability of the patient actually having COVID-19, given the test result and the presence of a cough.
Bayes' rule can be expressed as:
P(A|B) = (P(B|A) * P(A)) / P(B)
where:
P(A|B) is the probability of A given B (post-test probability of having COVID-19 given the positive test and cough),
P(B|A) is the probability of B given A (sensitivity of the test),
P(A) is the probability of A (prevalence of COVID-19 in the population),
P(B) is the probability of B (probability of testing positive with the cough).
Let's assume the sensitivity of the COVID-19 test is 0.85 (or 85%) and the specificity is 0.95 (or 95%). Additionally, let's say the prevalence of COVID-19 in the population is 0.05 (or 5%).
To calculate the post-test probability, we can substitute the given values into Bayes' rule:
P(A|B) = (P(B|A) * P(A)) / P(B)
P(A|B) = (0.85 * 0.05) / P(B)
Now, we need to determine P(B), the probability of testing positive with a cough. This can be calculated by considering both true positives and false positives:
P(B) = (P(B|A) * P(A)) + (P(B|(*$\sim$*)A) * P((*$\sim$*)A))
P(B) = (0.85 * 0.05) + (1 - specificity) * (1 - P(A))
Given that specificity is 0.95 (or 95%), we can substitute these values:
P(B) = (0.85 * 0.05) + (1 - 0.95) * (1 - 0.05)
P(B) = 0.0425 + 0.05 * 0.05
P(B) = 0.0425 + 0.0025
P(B) = 0.045
Now, substitute the values back into the original equation to find the post-test probability:
P(A|B) = (0.85 * 0.05) / 0.045
P(A|B) = 0.0425 / 0.045
P(A|B) (*$\approx$*) 0.9444
Therefore, the post-test probability of the patient actually having COVID-19, given the positive test and the presence of a cough, is approximately 0.9444 or 94.44%.
\end{lstlisting}
It does not define sensitivity correctly; it says that the joint probability  $p(cough, test + | covid )$ is sensitivity, when it should be $p(test +|cough,covid).$ However this aligns with the fact that it has unconditional prevalence - it defines pre-test as $P(A)$, the probability of A (prevalence of COVID-19 in the population), which is Error \ref{error:uncondPrior}.

3
\url{https://chatgpt.com/share/4be6dd8f-e256-429d-b620-431f3e349766}

\begin{lstlisting}[ escapeinside={(*}{*)},backgroundcolor = \color{yellow}] 
Can you give an example of using Bayes rule to compute post test probability in a patient with a positive Covid test and a cough using sensitivity and specificity of the test.
\end{lstlisting}

\begin{lstlisting}[ escapeinside={(*}{*)},backgroundcolor = \color{lightgray}]
ChatGPT
Certainly! Let's assume we have the following information:
	Sensitivity of the Covid test: 90% (the probability of a positive test result given that the patient has Covid)
	Specificity of the Covid test: 95% (the probability of a negative test result given that the patient does not have Covid)
	Prevalence of Covid in the population: 5% (the proportion of people who have Covid in the population)
Now, let's calculate the post-test probability of having Covid given a positive test result and the presence of a cough.
Step 1: Calculate the probability of a positive test result in a person with Covid (True Positive):
P(Positive test | Covid) = Sensitivity = 0.90
Step 2: Calculate the probability of a negative test result in a person without Covid (True Negative):
P(Negative test | No Covid) = Specificity = 0.95
Step 3: Calculate the probability of having Covid (prior probability):
P(Covid) = Prevalence = 0.05
Step 4: Calculate the probability of not having Covid (prior probability):
P(No Covid) = 1 - P(Covid) = 1 - 0.05 = 0.95
Step 5: Calculate the probability of a positive test result regardless of having Covid or not (Total Positive):
P(Positive test) = P(Positive test | Covid) * P(Covid) + P(Positive test | No Covid) * P(No Covid)
= 0.90 * 0.05 + (1 - 0.95) * 0.95
= 0.045 + 0.0475
= 0.0925
Step 6: Calculate the post-test probability of having Covid given a positive test result and a cough using Bayes' rule:
P(Covid | Positive test, Cough) = (P(Positive test | Covid) * P(Covid | Cough)) / P(Positive test)
We need to calculate P(Covid | Cough) for the numerator:
P(Covid | Cough) = (P(Cough | Covid) * P(Covid)) / P(Cough)
Let's assume the probability of having a cough given that the patient has Covid is 70% (sensitivity of the cough):
P(Cough | Covid) = Sensitivity of the cough = 0.70
P(Covid | Cough) = (0.70 * 0.05) / P(Cough)
Since we don't have the information about the prevalence of cough in the population, we cannot calculate P(Cough). However, if we assume that the prevalence of cough is similar to the prevalence of Covid (5%), we can estimate it:
P(Covid | Cough) = (0.70 * 0.05) / 0.05
= 0.70
Now, we can calculate the final post-test probability:
P(Covid | Positive test, Cough) = (P(Positive test | Covid) * P(Covid | Cough)) / P(Positive test)
= (0.90 * 0.70) / 0.0925
(*$\approx$*) 0.90
Therefore, the post-test probability of having Covid given a positive test result and the presence of a cough is approximately 0.90 or 90%.
\end{lstlisting}
It states that 
$P(Covid | Positive test, Cough)$ equals \[(P(Positive test | Covid) * P(Covid | Cough)) / P(Positive test).\] It has assumed cough is independent of test, which is Error \ref{error:constantSens}. Note that under Error \ref{error:constantSens}, it factors the denominator, and so $P(cough)$ cancels from the numerator.

4
\url{https://chatgpt.com/share/1e7168db-b714-48ff-affb-a18aba62502e}
\begin{lstlisting}[ escapeinside={(*}{*)},backgroundcolor = \color{yellow}] 
Can you give an example of using Bayes rule to compute post test probability in a patient with a positive Covid test and a cough using sensitivity and specificity of the test.
\end{lstlisting}

\begin{lstlisting}[ escapeinside={(*}{*)},backgroundcolor = \color{lightgray}]
ChatGPT
Certainly! Let's say we have a patient who has just tested positive for COVID-19 and is also experiencing a cough. We want to use Bayes' rule to calculate the post-test probability of the patient having COVID-19 given the positive test result and the presence of a cough.
To apply Bayes' rule, we need to know the sensitivity and specificity of the test. Sensitivity refers to the probability of the test correctly identifying a positive case, while specificity is the probability of the test correctly identifying a negative case.
Let's assume that the sensitivity of the COVID-19 test is 0.90 (or 90%) and the specificity is 0.95 (or 95%).
To calculate the post-test probability using Bayes' rule, we need to consider the prior probability (prevalence) of the disease in the population. Let's assume that the prevalence of COVID-19 in the population is 0.05 (or 5%).
Now, let's calculate the post-test probability of the patient having COVID-19 given the positive test result and the presence of a cough.
Step 1: Calculate the probability of having a positive test result given COVID-19 and a cough (P(Positive Test | COVID-19, Cough)) using sensitivity.
P(Positive Test | COVID-19, Cough) = Sensitivity = 0.90
Step 2: Calculate the probability of having a positive test result given no COVID-19 and a cough (P(Positive Test | No COVID-19, Cough)) using specificity.
P(Positive Test | No COVID-19, Cough) = 1 - Specificity = 1 - 0.95 = 0.05
Step 3: Calculate the prior probability of having COVID-19 (P(COVID-19)).
P(COVID-19) = Prevalence = 0.05
Step 4: Calculate the prior probability of not having COVID-19 (P(No COVID-19)).
P(No COVID-19) = 1 - P(COVID-19) = 1 - 0.05 = 0.95
Step 5: Apply Bayes' rule to calculate the post-test probability of having COVID-19 given a positive test result and a cough.
P(COVID-19 | Positive Test, Cough) = (P(Positive Test | COVID-19, Cough) * P(COVID-19)) / [(P(Positive Test | COVID-19, Cough) * P(COVID-19)) + (P(Positive Test | No COVID-19, Cough) * P(No COVID-19))]
= (0.90 * 0.05) / [(0.90 * 0.05) + (0.05 * 0.95)]
= 0.045 / (0.045 + 0.0475)
= 0.045 / 0.0925
(*$\approx$*) 0.4865
Therefore, the post-test probability of the patient having COVID-19 given a positive test result and a cough is approximately 0.4865 or 48.65%.
Please note that this example assumes idealized values for sensitivity, specificity, and prevalence, and real-world values may vary.
\end{lstlisting}
It writes 
$P(COVID-19 | Positive Test, Cough) = (P(Positive Test | COVID-19, Cough) * P(COVID-19)) / d
$ where 
\begin{align*}
d&=[(P(Positive Test | COVID-19, Cough) * P(COVID-19)) \\
&+ (P(Positive Test | No COVID-19, Cough) * P(No COVID-19))]
\end{align*}
It uses $P(covid)$ instead of $P(covid|cough)$, which is Error \ref{error:uncondPrior}.
It should be $P(covid-19,cough)$ in the denominator, not $p(covid)$ Error \ref{error:miscProb} (probability).  Possibly, this is due to the prototypical prior, or pre-test, being defined as unconditional prevalence.  The expression above could be correct under the version of Error \ref{error:noConditionCough} in which cough is considered independent of Covid-19, but ChatGPT did not omit cough from the posterior expression $P(Covid-19|test,cough),$ so we do not count it as this error.

5
\url{https://chatgpt.com/share/9e1cd0a8-0db6-4f18-919f-4fac08fd9af2}
\begin{lstlisting}[ escapeinside={(*}{*)},backgroundcolor = \color{yellow}] 
Can you give an example of using Bayes rule to compute post test probability in a patient with a positive Covid test and a cough using sensitivity and specificity of the test.
\end{lstlisting}

\begin{lstlisting}[ escapeinside={(*}{*)},backgroundcolor = \color{lightgray}]
ChatGPT
Certainly! Let's assume we have the following information:
	Sensitivity of the Covid test: 0.95 (95%)
	Specificity of the Covid test: 0.90 (90%)
	Prevalence of Covid in the population: 0.05 (5%)
	Probability of having a cough given Covid: 0.80 (80%)
We want to compute the post-test probability of a patient having Covid given that they have tested positive and have a cough.
To do this, we can use Bayes' rule, which states:
P(Covid|Positive Test, Cough) = [P(Positive Test|Covid) * P(Cough|Covid) * P(Covid)] / P(Positive Test)
Now, let's break down the terms:
P(Positive Test|Covid) is the sensitivity of the test, which is 0.95.
P(Cough|Covid) is the probability of having a cough given Covid, which is 0.80.
P(Covid) is the prevalence of Covid in the population, which is 0.05.
To calculate P(Positive Test), we need to consider both the true positive rate and the false positive rate:
P(Positive Test) = [P(Positive Test|Covid) * P(Covid)] + [P(Positive Test|No Covid) * P(No Covid)]
Since the specificity of the test is 0.90, the false positive rate is 1 - specificity = 0.10. Given that the complement of Covid (No Covid) has a prevalence of 1 - P(Covid) = 0.95, we can calculate P(Positive Test) as follows:
P(Positive Test) = [0.95 * 0.05] + [0.10 * 0.95] = 0.0475 + 0.095 = 0.1425
Now, let's plug in these values into Bayes' rule:
P(Covid|Positive Test, Cough) = [0.95 * 0.80 * 0.05] / 0.1425
Simplifying further:
P(Covid|Positive Test, Cough) = 0.038 / 0.1425
P(Covid|Positive Test, Cough) (*$\approx$*) 0.267 (or 26.7%)
Therefore, the post-test probability of a patient having Covid given that they have tested positive and have a cough is approximately 26.7%.
\end{lstlisting}
It writes $P(Covid|Positive Test, Cough) = \frac{[P(Positive Test|Covid) * P(Cough|Covid) * P(Covid)]}{P(Positive Test)}.$ This has
Error \ref{error:constantSens} in $P(Positive Test|Covid)$ and 
also has Error \ref{error:miscProb}, since it should be $P(positive test, cough)$ in the denominator.

6
\url{https://chatgpt.com/share/bd1339dd-cad0-4e54-a783-1512e5300e53}
\begin{lstlisting}[ escapeinside={(*}{*)},backgroundcolor = \color{yellow}] 
Can you give an example of using Bayes rule to compute post test probability in a patient with a positive Covid test and a cough using sensitivity and specificity of the test.
\end{lstlisting}

\begin{lstlisting}[ escapeinside={(*}{*)},backgroundcolor = \color{lightgray}]
ChatGPT
Certainly! Bayes' theorem is a useful tool for updating the probability of an event based on new information. In this case, let's assume we have a patient who tests positive for COVID-19 and also has a cough. We want to compute the post-test probability of the patient actually having COVID-19 given this information, using the sensitivity and specificity of the test.
Let's assume the sensitivity of the COVID-19 test is 0.90, meaning that the probability of a positive test result given that the patient has COVID-19 is 0.90. Similarly, let's assume the specificity of the test is 0.95, meaning that the probability of a negative test result given that the patient does not have COVID-19 is 0.95.
To calculate the post-test probability, we can use Bayes' theorem:
Posterior Probability = (Sensitivity * Prior Probability) / ((Sensitivity * Prior Probability) + ((1 - Specificity) * (1 - Prior Probability)))
In this case, the prior probability represents the probability of a patient having COVID-19 before any test results are considered. We will assume a prior probability of 0.10, which means 10% of the population has COVID-19 before any testing.
Let's calculate the post-test probability:
Posterior Probability = (0.90 * 0.10) / ((0.90 * 0.10) + ((1 - 0.95) * (1 - 0.10)))
Posterior Probability = 0.09 / (0.09 + (0.05 * 0.90))
Posterior Probability = 0.09 / (0.09 + 0.045)
Posterior Probability = 0.09 / 0.135
Posterior Probability (*$\approx$*) 0.667 or 66.7%
Therefore, given a positive COVID-19 test result and the presence of a cough, the post-test probability of the patient actually having COVID-19 is approximately 66.7%.
It's important to note that this calculation is based on the given sensitivity and specificity values, as well as the assumed prior probability. The actual accuracy of the test and the prevalence of COVID-19 in the population can affect the accuracy of the post-test probability.
\end{lstlisting}
Subtly, it does say “given a positive COVID-19 test result and the presence of a cough, the post-test probability,” so it looks for the right posterior. It’s difficult to tell, because it did not write $P(covid|test,cough)$ - so it might just be writing this but computing $p(covid|test),$ but we will give it the benefit of the doubt.
However, the sensitivity does not condition on cough (Error \ref{error:constantSens}), since it says
“Let's assume the sensitivity of the COVID-19 test is 0.90, meaning that the probability of a positive test result given that the patient has COVID-19 is 0.90.”
Also, the prior is defined as a ``prior probability [that] represents the probability of a patient having COVID-19 before any test results are considered,'' which is  Error \ref{error:uncondPrior}.

7
\url{https://chatgpt.com/share/e17eafdc-9cc3-4b3a-864b-9fc6cde67c77}

\begin{lstlisting}[ escapeinside={(*}{*)},backgroundcolor = \color{yellow}] 
Can you give an example of using Bayes rule to compute post test probability in a patient with a positive Covid test and a cough using sensitivity and specificity of the test.
\end{lstlisting}

\begin{lstlisting}[ escapeinside={(*}{*)},backgroundcolor = \color{lightgray}]
ChatGPT
Certainly! Let's say we have a patient who has undergone a COVID-19 test, which has a sensitivity of 0.90 (or 90%) and a specificity of 0.95 (or 95%). The patient's test result came back positive, and they also have a cough. We can use Bayes' rule to compute the post-test probability of the patient having COVID-19 given these test characteristics.
To apply Bayes' rule, we need to consider the prior probability of the patient having COVID-19, denoted as P(COVID). Let's assume that, based on the patient's symptoms and medical history, the physician estimates the prior probability as 0.05 (or 5%), which means there is a 5% chance of the patient having COVID-19 before the test result.
We can calculate the post-test probability using the following steps:
Step 1: Calculate the complement of the prior probability (P((*$\neg$*)COVID)):
P((*$\neg$*)COVID) = 1 - P(COVID) = 1 - 0.05 = 0.95 (or 95%)
Step 2: Calculate the true positive rate (TPR) using sensitivity:
TPR = sensitivity = 0.90 (or 90%)
Step 3: Calculate the false positive rate (FPR) using specificity:
FPR = 1 - specificity = 1 - 0.95 = 0.05 (or 5%)
Step 4: Apply Bayes' rule to compute the post-test probability (P(COVID|Positive)):
P(COVID|Positive) = (TPR * P(COVID)) / [(TPR * P(COVID)) + (FPR * P((*$\neg$*)COVID))]
P(COVID|Positive) = (0.90 * 0.05) / [(0.90 * 0.05) + (0.05 * 0.95)]
P(COVID|Positive) = 0.045 / (0.045 + 0.0475)
P(COVID|Positive) = 0.045 / 0.0925
P(COVID|Positive) (*$\approx$*) 0.4865 (or 48.65%)
Therefore, the post-test probability of the patient having COVID-19, given a positive test result and a cough, is approximately 48.65%.
It's important to note that this calculation assumes that the sensitivity and specificity values provided are accurate and that the test is the only factor influencing the probability estimation. In a real clinical scenario, additional factors such as the prevalence of COVID-19 in the population, other symptoms, and the patient's medical history would also be taken into account for a more comprehensive analysis.
\end{lstlisting}
It does not condition on cough in the posterior, which is Error \ref{error:noConditionCough}. We prompt it to do so.

\begin{lstlisting}[ escapeinside={(*}{*)},backgroundcolor = \color{yellow}] 
User
This is P(COVID-19|Positive test), not P(COVID-19|Positive test, cough) though isn't it?
\end{lstlisting}

\begin{lstlisting}[ escapeinside={(*}{*)},backgroundcolor = \color{lightgray}]
ChatGPT
Apologies for the confusion. You are correct. The probability I calculated in the previous response is P(COVID-19|Positive test), not P(COVID-19|Positive test, cough). To compute the probability of having COVID-19 given both a positive test and a cough, we need additional information: the conditional probability of having a cough given COVID-19, denoted as P(Cough|COVID-19).
Let's assume that the physician estimates the conditional probability of having a cough given COVID-19 as 0.80 (or 80%). With this additional information, we can recalculate the post-test probability.
Using Bayes' rule, the new calculation would be as follows:
P(COVID-19|Positive test, cough) = [P(Positive test|COVID-19) * P(Cough|COVID-19) * P(COVID-19)] / P(Positive test, cough)
We already know P(Positive test|COVID-19) as the sensitivity, which is 0.90, and we have P(COVID-19) as the prior probability, which is 0.05. Now we need to calculate P(Positive test, cough) using the law of total probability:
P(Positive test, cough) = [P(Positive test|COVID-19) * P(Cough|COVID-19) * P(COVID-19)] + [P(Positive test|(*$\neg$*)COVID-19) * P(Cough|(*$\neg$*)COVID-19) * P((*$\neg$*)COVID-19)]
Assuming the physician estimates P(Cough|(*$\neg$*)COVID-19) as 0.10 (or 10%), we can proceed with the calculation:
P(Positive test, cough) = [0.90 * 0.80 * 0.05] + [P(Positive test|(*$\neg$*)COVID-19) * P(Cough|(*$\neg$*)COVID-19) * P((*$\neg$*)COVID-19)]
We need the specificity to calculate P(Positive test|(*$\neg$*)COVID-19). Assuming the specificity is 0.95, we can proceed:
P(Positive test, cough) = [0.90 * 0.80 * 0.05] + [0.05 * 0.10 * 0.95]
Now, we can calculate the final probability:
P(COVID-19|Positive test, cough) = [0.90 * 0.80 * 0.05] / [0.90 * 0.80 * 0.05 + 0.05 * 0.10 * 0.95]
I apologize for not having all the necessary values to provide a complete calculation in this response. Please provide the missing value (P(Positive test|(*$\neg$*)COVID-19)) for a more accurate calculation, and I'll be happy to assist you further.
\end{lstlisting}
 It writes $P(COVID-19|Positive test, cough) = \frac{[P(Positive test|COVID-19) * P(Cough|COVID-19) * P(COVID-19)]}{P(Positive test, cough)}$, which has Error \ref{error:constantSens}.

8
\url{https://chatgpt.com/share/23727ebe-c107-43bc-ab25-fabbe303e766}
\begin{lstlisting}[ escapeinside={(*}{*)},backgroundcolor = \color{yellow}] 
Can you give an example of using Bayes rule to compute post test probability in a patient with a positive Covid test and a cough using sensitivity and specificity of the test.
\end{lstlisting}

\begin{lstlisting}[ escapeinside={(*}{*)},backgroundcolor = \color{lightgray}]
ChatGPT
Certainly! Let's assume we have a patient who has taken a COVID-19 test and the result came back positive. The patient also presents with a cough, which is one of the symptoms associated with COVID-19. We can use Bayes' theorem to compute the post-test probability of the patient having COVID-19 given the positive test result and the presence of a cough.
Let's define the following terms:
P(COVID-19): Probability of having COVID-19 (prior probability).
P(Positive|COVID-19): Sensitivity of the test (probability of a positive test result given that the patient has COVID-19).
P(Cough|COVID-19): Probability of having a cough given that the patient has COVID-19.
P(Positive|No COVID-19): False positive rate (probability of a positive test result given that the patient does not have COVID-19).
P(Cough|No COVID-19): Probability of having a cough given that the patient does not have COVID-19.
Now, let's use Bayes' theorem to calculate the post-test probability:
P(COVID-19|Positive, Cough) = (P(Positive|COVID-19) * P(Cough|COVID-19) * P(COVID-19)) / (P(Positive|COVID-19) * P(Cough|COVID-19) * P(COVID-19) + P(Positive|No COVID-19) * P(Cough|No COVID-19) * P(No COVID-19))
You would need the actual values for sensitivity, specificity, and the prevalence of COVID-19 in the population to calculate the post-test probability. Let's assume the following hypothetical values:
P(Positive|COVID-19) = 0.95 (95% sensitivity)
P(Cough|COVID-19) = 0.70 (70% of COVID-19 patients have a cough)
P(Positive|No COVID-19) = 0.05 (5% false positive rate)
P(Cough|No COVID-19) = 0.10 (10% of non-COVID-19 patients have a cough)
P(COVID-19) = 0.02 (2% prevalence of COVID-19 in the population)
Substituting these values into the formula:
P(COVID-19|Positive, Cough) = (0.95 * 0.70 * 0.02) / (0.95 * 0.70 * 0.02 + 0.05 * 0.10 * (1 - 0.02))
P(COVID-19|Positive, Cough) (*$\approx$*) 0.658 or 65.8%
Therefore, the post-test probability of the patient having COVID-19, given a positive test result and the presence of a cough, is approximately 65.8%.
It's important to note that these values are hypothetical and are used for illustrative purposes. In reality, the sensitivity, specificity, and prevalence rates may vary, and it's crucial to use accurate and up-to-date values for precise 
calculations.
\end{lstlisting}
It writes  that $P(COVID-19|Positive, Cough)$ equals
\[
\frac{(P(Positive|COVID-19) * P(Cough|COVID-19) * P(COVID-19))}{den}
\]
where 
\begin{align*}
den&=(P(Positive|COVID-19) * P(Cough|COVID-19) * P(COVID-19)\\ 
&+ P(Positive|No COVID-19) * P(Cough|No COVID-19) * P(No COVID-19))
\end{align*}
It makes Error \ref{error:constantSens} (sensitivity does not depend on cough) and Error \ref{error:uncondPrior}, since it writes $P(covid)$ instead of $p(covid|cough).$

9
\url{https://chatgpt.com/share/f69ac9ff-a4df-48ca-b25c-d3329da05a08}
\begin{lstlisting}[ escapeinside={(*}{*)},backgroundcolor = \color{yellow}] 
Can you give an example of using Bayes rule to compute post test probability in a patient with a positive Covid test and a cough using sensitivity and specificity of the test.
\end{lstlisting}

\begin{lstlisting}[ escapeinside={(*}{*)},backgroundcolor = \color{lightgray}]
ChatGPT
Certainly! Let's assume we have the following information:
Sensitivity of the Covid test: 0.95 (or 95%)
Specificity of the Covid test: 0.90 (or 90%)
We also have the prevalence of Covid in the population, which we'll assume is 0.05 (or 5%).
Now, let's use Bayes' rule to calculate the post-test probability of having Covid given a positive test result and a cough.
First, we need to define the relevant probabilities:
P(Covid) - The probability of having Covid (prevalence in the population) = 0.05
P(Positive|Covid) - The probability of testing positive given that you have Covid (sensitivity) = 0.95
P(Negative|No Covid) - The probability of testing negative given that you don't have Covid (specificity) = 0.90
Now, we can use Bayes' rule:
P(Covid|Positive) = (P(Positive|Covid) * P(Covid)) / P(Positive)
To calculate P(Positive), we need to consider both the true positive and false positive rates:
P(Positive) = P(Positive|Covid) * P(Covid) + P(Positive|No Covid) * P(No Covid)
P(Positive|No Covid) is equal to 1 - specificity:
P(Positive|No Covid) = 1 - 0.90 = 0.10
P(No Covid) is equal to 1 - P(Covid):
P(No Covid) = 1 - 0.05 = 0.95
Now, let's calculate P(Positive):
P(Positive) = (0.95 * 0.05) + (0.10 * 0.95) = 0.0475 + 0.095 = 0.1425
Finally, we can substitute the values into the Bayes' rule equation:
P(Covid|Positive) = (0.95 * 0.05) / 0.1425 (*$\approx$*) 0.3333
Therefore, the post-test probability of having Covid, given a positive test result and a cough, is approximately 0.3333 or 33.33%.
It's important to note that this calculation is based on the assumption of the given sensitivity, specificity, and prevalence values. In practice, these values may vary, and additional clinical information should be considered for a more accurate assessment.
\end{lstlisting}

\begin{lstlisting}[ escapeinside={(*}{*)},backgroundcolor = \color{yellow}] 
User
This is P(COVID-19|Positive test), not P(COVID-19|Positive test, cough) though isn't it?
\end{lstlisting}
It makes 
Error \ref{error:noConditionCough} (first response), since it writes 
\[P(Covid|Positive) = (P(Positive|Covid) * P(Covid)) / P(Positive).\]

\begin{lstlisting}[ escapeinside={(*}{*)},backgroundcolor = \color{lightgray}]
ChatGPT
You're right, my apologies for the confusion. Let's calculate the post-test probability of having Covid given a positive test result and a cough, taking into account both the test result and the presence of a cough.
Let's define the following probabilities:
P(Covid) - The probability of having Covid (prevalence in the population) = 0.05
P(Positive|Covid) - The probability of testing positive given that you have Covid (sensitivity) = 0.95
P(Cough|Covid) - The probability of having a cough given that you have Covid = let's assume it's 0.80 (or 80%)
P(Cough|No Covid) - The probability of having a cough given that you don't have Covid = let's assume it's 0.10 (or 10%)
Now, we can use Bayes' rule to calculate the post-test probability:
P(Covid|Positive, Cough) = (P(Positive|Covid) * P(Covid) * P(Cough|Covid)) / P(Positive, Cough)
To calculate P(Positive, Cough), we consider both true positive and false positive rates:
P(Positive, Cough) = P(Positive|Covid) * P(Covid) * P(Cough|Covid) + P(Positive|No Covid) * P(No Covid) * P(Cough|No Covid)
P(Positive|No Covid) is equal to 1 - specificity:
P(Positive|No Covid) = 1 - 0.90 = 0.10
P(No Covid) is equal to 1 - P(Covid):
P(No Covid) = 1 - 0.05 = 0.95
Now, let's calculate P(Positive, Cough):
P(Positive, Cough) = (0.95 * 0.05 * 0.80) + (0.10 * 0.95 * 0.10) = 0.038 + 0.0095 = 0.0475
Finally, we can substitute the values into the Bayes' rule equation:
P(Covid|Positive, Cough) = (0.95 * 0.05 * 0.80) / 0.0475 (*$\approx$*) 0.84
Therefore, the post-test probability of having Covid, given a positive test result and a cough, is approximately 0.84 or 84%.
\end{lstlisting}
 It  makes
Error \ref{error:constantSens} and Error \ref{error:uncondPrior} in its follow up response, since it writes 
$P(Covid|Positive, Cough) = \frac{(P(Positive|Covid) * P(Covid) * P(Cough|Covid))}{P(Positive, Cough)}.$

10
\url{https://chatgpt.com/share/2c298ebc-21ef-43bf-a361-e2afa592f199}

\begin{lstlisting}[ escapeinside={(*}{*)},backgroundcolor = \color{yellow}] 
Can you give an example of using Bayes rule to compute post test probability in a patient with a positive Covid test and a cough using sensitivity and specificity of the test.
\end{lstlisting}

\begin{lstlisting}[ escapeinside={(*}{*)},backgroundcolor = \color{lightgray}]
ChatGPT
Certainly! Let's assume we have the following information:
	Sensitivity of the Covid test: 95% (the probability of a positive test given that the patient has Covid)
	Specificity of the Covid test: 90% (the probability of a negative test given that the patient does not have Covid)
	Prevalence of Covid in the population: 5% (the proportion of people who actually have Covid)
We want to compute the post-test probability of a patient having Covid given that they have tested positive and have a cough.
To solve this problem using Bayes' rule, we'll need to calculate the prior probability and likelihood.
	1	Prior Probability:
	2	The prior probability is the probability of having Covid before the test result and cough information is known. In this case, the prevalence of Covid in the population is 5%, so the prior probability of a random individual having Covid is 0.05.
	3	Likelihood:
	4	The likelihood is the probability of observing a positive test and a cough given that the patient has Covid. Since we don't have the specific data on the likelihood of a cough for Covid-positive patients, let's assume it to be 80%. This means that 80% of Covid-positive patients have a cough.
	5	Computation:
	6	Let P(Covid) represent the probability of having Covid, P(Positive) represent the probability of testing positive, and P(Cough) represent the probability of having a cough. We want to calculate P(Covid|Positive and Cough), the post-test probability.
According to Bayes' rule, the formula is as follows:
P(Covid|Positive and Cough) = (P(Positive and Cough|Covid) * P(Covid)) / P(Positive and Cough)
We can break down P(Positive and Cough) using the law of total probability:
P(Positive and Cough) = P(Positive and Cough|Covid) * P(Covid) + P(Positive and Cough|Not Covid) * P(Not Covid)
Since we know the sensitivity and specificity of the test, we can calculate P(Positive and Cough|Covid) and P(Positive and Cough|Not Covid):
P(Positive and Cough|Covid) = Sensitivity * Likelihood = 0.95 * 0.80 = 0.76
P(Positive and Cough|Not Covid) = (1 - Specificity) * (1 - Likelihood) = 0.10 * 0.20 = 0.02
Now we can calculate P(Positive and Cough):
P(Positive and Cough) = P(Positive and Cough|Covid) * P(Covid) + P(Positive and Cough|Not Covid) * P(Not Covid)
= 0.76 * 0.05 + 0.02 * (1 - 0.05)
= 0.038 + 0.019
= 0.057
Finally, we can compute P(Covid|Positive and Cough):
P(Covid|Positive and Cough) = (P(Positive and Cough|Covid) * P(Covid)) / P(Positive and Cough)
= (0.76 * 0.05) / 0.057
(*$\approx$*) 0.6667 (rounded to four decimal places)
Therefore, the post-test probability of a patient having Covid given a positive test result and a cough is approximately 0.6667 or 66.67%.
\end{lstlisting}
It writes $P(Covid|Positive and Cough) = \frac{(P(Positive and Cough|Covid) * P(Covid))}{P(Positive and Cough)},$ which has Error \ref{error:uncondPrior}.


11
\url{https://chatgpt.com/share/e58ccdb3-d52c-4785-8345-2521a6d8d074}

\begin{lstlisting}[ escapeinside={(*}{*)},backgroundcolor = \color{yellow}] 
Can you give an example of using Bayes rule to compute post test probability in a patient with a positive Covid test and a cough using sensitivity and specificity of the test.
\end{lstlisting}

\begin{lstlisting}[ escapeinside={(*}{*)},backgroundcolor = \color{lightgray}]
ChatGPT
Certainly! Bayes' rule is a powerful tool for updating probabilities based on new information. Let's assume we have the following information:
	Sensitivity: The probability of a positive test result given that the patient has COVID-19 is 0.95 (or 95%).
	Specificity: The probability of a negative test result given that the patient does not have COVID-19 is 0.90 (or 90%).
	Prevalence: The probability of a randomly selected person having COVID-19 is 0.05 (or 5%).
Now, let's use Bayes' rule to calculate the post-test probability of a patient having COVID-19 given a positive test result and a cough.
Step 1: Define the events:
A: The patient has COVID-19.
B: The patient has a positive test result and a cough.
Step 2: Calculate the prior probability:
P(A) = Prevalence = 0.05
Step 3: Calculate the likelihoods:
P(B|A) = Sensitivity = 0.95
P(B|not A) = 1 - Specificity = 1 - 0.90 = 0.10
Step 4: Calculate the marginal probability of B:
P(B) = P(A) * P(B|A) + P(not A) * P(B|not A)
= 0.05 * 0.95 + (1 - 0.05) * 0.10
= 0.0475 + 0.095
= 0.1425
Step 5: Apply Bayes' rule to calculate the post-test probability:
P(A|B) = P(A) * P(B|A) / P(B)
= 0.05 * 0.95 / 0.1425
(*$\approx$*) 0.333
Therefore, the post-test probability of a patient having COVID-19, given a positive test result and a cough, is approximately 0.333 or 33.3%.
Please note that this is a simplified example and does not take into account other factors or potential confounding variables. It's important to consult with healthcare professionals for accurate and comprehensive medical advice.
\end{lstlisting}
It states that ``the post-test probability of a patient having COVID-19, given a positive test result and a cough,''
but sensitivity is defined as just ``[t]he probability of a positive test result given that the patient has COVID-19.''  It then refers to $P(B|A)$ as sensitivity, i.e. it states that $P(B|A) = Sensitivity,$ so it has implied that $P(B|A)=P(test|covid,cough)=p(test|covid),$ which is Error \ref{error:constantSens}. It also makes Error \ref{error:uncondPrior}.

12
\url{https://chatgpt.com/share/a0124217-1d9e-4594-8ff4-03b59892166a}
\begin{lstlisting}[ escapeinside={(*}{*)},backgroundcolor = \color{yellow}] 
Can you give an example of using Bayes rule to compute post test probability in a patient with a positive Covid test and a cough using sensitivity and specificity of the test.
\end{lstlisting}

\begin{lstlisting}[ escapeinside={(*}{*)},backgroundcolor = \color{lightgray}]
ChatGPT
Certainly! Bayes' theorem is commonly used in medical diagnostics to calculate the post-test probability of a condition given the result of a diagnostic test. Let's consider an example of using Bayes' rule to compute the post-test probability of a patient having COVID-19, given a positive test result and the presence of a cough.
Assumptions:
		Sensitivity: The sensitivity of the COVID-19 test refers to its ability to correctly identify positive cases. Let's assume the sensitivity of the test is 90%.
		Specificity: The specificity of the COVID-19 test refers to its ability to correctly identify negative cases. Let's assume the specificity of the test is 95%.
		Prevalence: The prevalence of COVID-19 in the population being tested is 2%. This means that 2% of the population has COVID-19 before the test is performed.
Now, let's calculate the post-test probability using Bayes' theorem:
Step 1: Define the probabilities:
		P(COVID): Probability of having COVID-19 before the test. We assumed a prevalence of 2%, so P(COVID) = 0.02.
		P((*$\neg$*)COVID): Probability of not having COVID-19 before the test. P((*$\neg$*)COVID) = 1 - P(COVID) = 1 - 0.02 = 0.98.
		P(Pos|COVID): Sensitivity of the test. Given that a person has COVID-19, the probability of testing positive. P(Pos|COVID) = 0.90.
		P(Neg|(*$\neg$*)COVID): Specificity of the test. Given that a person does not have COVID-19, the probability of testing negative. P(Neg|(*$\neg$*)COVID) = 0.95.
Step 2: Calculate the post-test probabilities using Bayes' theorem:
		P(Pos): Probability of testing positive. This can be calculated using the sensitivity and specificity of the test:
		P(Pos) = [P(Pos|COVID) * P(COVID)] + [P(Pos|(*$\neg$*)COVID) * P((*$\neg$*)COVID)]
		= (0.90 * 0.02) + (0.05 * 0.98)
		= 0.018 + 0.049
		= 0.067
		P(COVID|Pos): Probability of having COVID-19 given a positive test result. This is what we want to calculate using Bayes' theorem:
		P(COVID|Pos) = [P(Pos|COVID) * P(COVID)] / P(Pos)
		= (0.90 * 0.02) / 0.067
		= 0.018 / 0.067
		(*$\approx$*) 0.2687 or 26.87% (approximately)
Therefore, in this example, if a patient tests positive for COVID-19 and has a cough, the post-test probability of the patient actually having COVID-19 is approximately 26.87%.
\end{lstlisting}

\begin{lstlisting}[ escapeinside={(*}{*)},backgroundcolor = \color{yellow}] 
This is P(COVID-19|Positive test), not P(COVID-19|Positive test, cough) though isn't it?
\end{lstlisting}
It only conditions on test result, giving $P(COVID|Pos)$ (Error \ref{error:noConditionCough}). 

\begin{lstlisting}[ escapeinside={(*}{*)},backgroundcolor = \color{lightgray}]
ChatGPT
Apologies for the oversight. You are correct. In the previous calculation, we only considered the post-test probability of having COVID-19 given a positive test result, without taking into account the presence of a cough. Let's revise the calculation to include the cough as well.
Assumptions:
		Sensitivity: The sensitivity of the COVID-19 test refers to its ability to correctly identify positive cases. Let's assume the sensitivity of the test is 90%.
		Specificity: The specificity of the COVID-19 test refers to its ability to correctly identify negative cases. Let's assume the specificity of the test is 95%.
		Prevalence: The prevalence of COVID-19 in the population being tested is 2%. This means that 2% of the population has COVID-19 before the test is performed.
		Cough: The presence of a cough increases the probability of having COVID-19. Let's assume that the conditional probability of having a cough given COVID-19 is 80%.
Now, let's calculate the post-test probability of having COVID-19 given a positive test result and the presence of a cough using Bayes' theorem:
Step 1: Define the probabilities:
		P(COVID): Probability of having COVID-19 before the test. We assumed a prevalence of 2%, so P(COVID) = 0.02.
		P((*$\neg$*)COVID): Probability of not having COVID-19 before the test. P((*$\neg$*)COVID) = 1 - P(COVID) = 1 - 0.02 = 0.98.
		P(Cough|COVID): Conditional probability of having a cough given COVID-19. We assumed P(Cough|COVID) = 0.80.
		P(Cough|(*$\neg$*)COVID): Conditional probability of having a cough given not having COVID-19. Let's assume P(Cough|(*$\neg$*)COVID) = 0.10.
		P(Pos|COVID): Sensitivity of the test. Given that a person has COVID-19, the probability of testing positive. P(Pos|COVID) = 0.90.
		P(Neg|(*$\neg$*)COVID): Specificity of the test. Given that a person does not have COVID-19, the probability of testing negative. P(Neg|(*$\neg$*)COVID) = 0.95.
Step 2: Calculate the post-test probability using Bayes' theorem:
		P(Pos and Cough): Probability of testing positive and having a cough. This can be calculated using the sensitivity and specificity of the test, as well as the conditional probabilities of having a cough given COVID-19 and not having COVID-19:
		P(Pos and Cough) = [P(Pos|COVID) * P(COVID) * P(Cough|COVID)] + [P(Pos|(*$\neg$*)COVID) * P((*$\neg$*)COVID) * P(Cough|(*$\neg$*)COVID)]
		= (0.90 * 0.02 * 0.80) + (0.05 * 0.98 * 0.10)
		= 0.0288 + 0.049
		= 0.0778
		P(COVID|Pos and Cough): Probability of having COVID-19 given a positive test result and the presence of a cough. This is what we want to calculate using Bayes' theorem:
		P(COVID|Pos and Cough) = [P(Pos|COVID) * P(COVID) * P(Cough|COVID)] / P(Pos and Cough)
		= (0.90 * 0.02 * 0.80) / 0.0778
		= 0.0288 / 0.0778
		(*$\approx$*) 0
\end{lstlisting}
In this follow up, it makes Error \ref{error:constantSens} by writing $P(Pos|COVID),$ and then it also makes Error \ref{error:uncondPrior} by writing the pre-test probability as $P(covid).$

13
\url{https://chatgpt.com/share/3bec1811-509f-449f-ad18-363905b3d13f}
\begin{lstlisting}[ escapeinside={(*}{*)},backgroundcolor = \color{yellow}] 
Can you give an example of using Bayes rule to compute post test probability in a patient with a positive Covid test and a cough using sensitivity and specificity of the test.
\end{lstlisting}

\begin{lstlisting}[ escapeinside={(*}{*)},backgroundcolor = \color{lightgray}]
ChatGPT
Certainly! Let's assume we have the following information:
Sensitivity of the Covid test: 0.95 (or 95%)
Specificity of the Covid test: 0.90 (or 90%)
Prevalence of Covid in the population: 0.05 (or 5%)
To compute the post-test probability of having Covid given a positive test result and a cough, we can use Bayes' rule. Bayes' rule states:
P(A|B) = (P(B|A) * P(A)) / P(B)
In this case:
A: Having Covid
B: Positive Covid test and a cough
Now let's calculate the post-test probability.
Step 1: Calculate P(B|A) - the probability of having a positive Covid test and a cough given that the patient has Covid.
This is the sensitivity of the test, which is given as 0.95.
P(B|A) = 0.95
Step 2: Calculate P(A) - the probability of having Covid, also known as the prevalence.
The prevalence of Covid in the population is given as 0.05.
P(A) = 0.05
Step 3: Calculate P(B) - the probability of having a positive Covid test and a cough.
To calculate this, we need to consider both the true positives (patients who have Covid and test positive) and the false positives (patients who do not have Covid but test positive). We can calculate it as follows:
P(B) = (P(B|A) * P(A)) + (P(B|not A) * P(not A))
P(B|not A) can be calculated using the specificity of the test, which is given as 0.90.
P(B|not A) = 1 - specificity = 1 - 0.90 = 0.10
P(not A) is the complement of P(A), i.e., 1 - P(A).
P(not A) = 1 - 0.05 = 0.95
Now we can calculate P(B):
P(B) = (0.95 * 0.05) + (0.10 * 0.95) = 0.0475 + 0.095 = 0.1425
Step 4: Calculate P(A|B) - the post-test probability of having Covid given a positive test result and a cough.
P(A|B) = (P(B|A) * P(A)) / P(B)
P(A|B) = (0.95 * 0.05) / 0.1425
P(A|B) (*$\approx$*) 0.3333 or approximately 33.33%
Therefore, given a positive Covid test result and a cough, the post-test probability of having Covid is approximately 33.33%.
\end{lstlisting}
It correctly conditions on test and cough. It defines sensitivity as $P(test,cough|dz),$ which is incorrect, but not a probability error. It does make Error \ref{error:uncondPrior}. 

14
\url{https://chatgpt.com/share/4d3988f8-5379-4fc7-9a5d-4bc53705a4b2}

\begin{lstlisting}[ escapeinside={(*}{*)},backgroundcolor = \color{yellow}] 
Can you give an example of using Bayes rule to compute post test probability in a patient with a positive Covid test and a cough using sensitivity and specificity of the test.
\end{lstlisting}

\begin{lstlisting}[ escapeinside={(*}{*)},backgroundcolor = \color{lightgray}]
ChatGPT
Certainly! Let's assume we have the following information:
Sensitivity of the Covid test: 90% (probability of a positive test given that the patient has Covid)
Specificity of the Covid test: 95% (probability of a negative test given that the patient does not have Covid)
Prevalence of Covid in the population: 5% (probability of a randomly selected person having Covid)
Now let's calculate the post-test probability of having Covid given a positive test and a cough using Bayes' rule.
Step 1: Assign the probabilities:
P(Covid) = 0.05 (prevalence of Covid in the population)
P(No Covid) = 1 - P(Covid) = 0.95 (complement of the prevalence of Covid)
Step 2: Calculate the joint probabilities using the sensitivity and specificity:
P(Positive Test | Covid) = 0.90 (sensitivity)
P(Negative Test | No Covid) = 0.95 (specificity)
Step 3: Apply Bayes' rule:
P(Covid | Positive Test and Cough) = [P(Positive Test | Covid) * P(Covid)] / [P(Positive Test | Covid) * P(Covid) + P(Positive Test | No Covid) * P(No Covid)]
P(Covid | Positive Test and Cough) = [0.90 * 0.05] / [0.90 * 0.05 + (1 - 0.95) * 0.95]
P(Covid | Positive Test and Cough) = 0.0434 / (0.0434 + 0.00225)
P(Covid | Positive Test and Cough) = 0.0434 / 0.04565
P(Covid | Positive Test and Cough) (*$\approx$*) 0.951
Therefore, the post-test probability of having Covid given a positive test and a cough is approximately 0.951, or 95.1%.
Please note that this example assumes independence between the Covid test result and the presence of a cough, which may not always be the case in reality. Additionally, the prevalence of Covid, as well as the sensitivity and specificity of the test, may vary depending on the specific circumstances and testing protocols.
\end{lstlisting}
It correctly conditions on cough and test, but it makes Error \ref{error:constantSens}, which it acknowledges: ``Please note that this example assumes independence between the Covid test result and the presence of a cough, which may not always be the case in reality''. It also uses $p(covid)$ as pre-test (Error \ref{error:uncondPrior}).

\begin{lstlisting}[ escapeinside={(*}{*)},backgroundcolor = \color{yellow}] 
Can you give an example of using Bayes rule to compute post test probability in a patient with a positive Covid test and a cough using sensitivity and specificity of the test.
\end{lstlisting}

15
\url{https://chatgpt.com/share/69fa0d82-0e94-4504-ae22-8c285de052e6}

\begin{lstlisting}[ escapeinside={(*}{*)},backgroundcolor = \color{lightgray}]
ChatGPT
Certainly! Let's assume we have a patient who has taken a COVID-19 test and received a positive result. The patient also has a cough, which is one of the symptoms associated with COVID-19. We can use Bayes' theorem to calculate the post-test probability of the patient having COVID-19 given the positive test result and the presence of a cough.
Bayes' theorem states:
P(A|B) = (P(B|A) * P(A)) / P(B)
Where:
P(A|B) is the probability of event A given event B (posterior probability).
P(B|A) is the probability of event B given event A (likelihood).
P(A) is the probability of event A occurring (prior probability).
P(B) is the probability of event B occurring.
In this case, let's define the events as follows:
A: Patient has COVID-19
B: Patient has a positive test result and a cough
We need to know the sensitivity and specificity of the test to calculate the likelihood (P(B|A)):
Sensitivity: The probability of a positive test result given that the patient has COVID-19.
Specificity: The probability of a negative test result given that the patient does not have COVID-19.
Let's assume the sensitivity of the test is 0.95 (95%) and the specificity is 0.90 (90%).
Now, we can calculate the post-test probability using Bayes' theorem:
P(A|B) = (P(B|A) * P(A)) / P(B)
P(A|B) = (Sensitivity * P(A)) / (Sensitivity * P(A) + (1 - Specificity) * P(not A))
To calculate the prior probability P(A), we need to know the prevalence of COVID-19 in the population. Let's assume the prevalence is 0.02 (2%).
P(A) = Prevalence = 0.02
Using the provided values, we can substitute them into the formula:
P(A|B) = (0.95 * 0.02) / (0.95 * 0.02 + (1 - 0.90) * (1 - 0.02))
P(A|B) = 0.019 / (0.019 + 0.098)
P(A|B) = 0.019 / 0.117
P(A|B) (*$\approx$*) 0.1624
Therefore, the post-test probability of the patient having COVID-19, given a positive test result and a cough, is approximately 0.1624 or 16.24%.
Please note that this example assumes the sensitivity and specificity values provided and a specific prevalence rate. These values can vary depending on the test and the population being considered. Additionally, this calculation is a simplified example for illustrative purposes and should not replace professional medical advice or diagnosis.
\end{lstlisting}
It makes Error \ref{error:uncondPrior}, since it defines $A$ as Covid. It does correctly condition on cough and test result in the posterior, since it defines $B$ as both of those events; however, it defines sensitivity as the ``probability of a positive test result given that the patient has COVID-19,'' and then it writes  $P(A|B) = (P(B|A) * P(A)) / P(B)$ and then plugs in sensitivity
$P(A|B) =\frac{(Sensitivity * P(A))}{den},$ where $den=(Sensitivity * P(A) + (1 - Specificity) * P(not A))$ which means that $P(B|A)=p(test|covid,cough)$ is considered to be $p(test|covid),$ which is Error \ref{error:constantSens}.

\begin{lstlisting}[ escapeinside={(*}{*)},backgroundcolor = \color{yellow}] 
Can you give an example of using Bayes rule to compute post test probability in a patient with a positive Covid test and a cough using sensitivity and specificity of the test.
\end{lstlisting}

16
\url{https://chatgpt.com/share/909e6d41-65fa-4a80-b2fc-b28c32ed0edb}

\begin{lstlisting}[ escapeinside={(*}{*)},backgroundcolor = \color{lightgray}]
ChatGPT
Certainly! Let's say we have a patient who recently took a COVID-19 test and the result came back positive. We also know that the patient has a cough. We can use Bayes' rule to compute the post-test probability of the patient actually having COVID-19, given the positive test result and the presence of a cough.
Let's assume we have the following information:
		Sensitivity: The sensitivity of the COVID-19 test is the probability that the test correctly identifies individuals who have COVID-19. Let's say the sensitivity of the test is 0.90, or 90%. This means that if a person has COVID-19, there is a 90% chance that the test will correctly identify them as positive.
		Specificity: The specificity of the COVID-19 test is the probability that the test correctly identifies individuals who do not have COVID-19. Let's say the specificity of the test is 0.95, or 95%. This means that if a person does not have COVID-19, there is a 95% chance that the test will correctly identify them as negative.
Now, let's calculate the post-test probability using Bayes' rule.
Let's define the following probabilities:
		P(COVID-19): The prior probability of a randomly selected person having COVID-19. This can be estimated based on the prevalence of COVID-19 in the population.
		P(Positive|COVID-19): The probability of testing positive given that the person has COVID-19 (sensitivity).
		P(Negative|No COVID-19): The probability of testing negative given that the person does not have COVID-19 (specificity).
		P(Cough|COVID-19): The probability of a person having a cough given that they have COVID-19. This can be estimated based on available data.
		P(Cough|No COVID-19): The probability of a person having a cough given that they do not have COVID-19. This can also be estimated based on available data.
Now, we can apply Bayes' rule:
P(COVID-19|Positive, Cough) = (P(Positive|COVID-19) * P(Cough|COVID-19) * P(COVID-19)) / (P(Positive|COVID-19) * P(Cough|COVID-19) * P(COVID-19) + P(Positive|No COVID-19) * P(Cough|No COVID-19) * P(No COVID-19))
Let's assume the following values:
		P(COVID-19) = 0.05 (5% of the population has COVID-19)
		P(Cough|COVID-19) = 0.80 (80% of COVID-19 patients have a cough)
		P(Cough|No COVID-19) = 0.10 (10% of non-COVID-19 patients have a cough)
Plugging these values into the equation, we get:
P(COVID-19|Positive, Cough) = (0.90 * 0.80 * 0.05) / (0.90 * 0.80 * 0.05 + 0.05 * 0.10 * 0.95)
Simplifying the equation further:
P(COVID-19|Positive, Cough) = 0.036 / (0.036 + 0.00475)
Calculating the final result:
P(COVID-19|Positive, Cough) (*$\approx$*) 0.88
Therefore, given a positive test result and the presence of a cough, the post-test probability of the patient actually having COVID-19 is approximately 88%.
\end{lstlisting}
It writes $P(COVID-19|Positive, Cough) = (P(Positive|COVID-19) * P(Cough|COVID-19) * P(COVID-19)) / den,$ where $den=(P(Positive|COVID-19) * P(Cough|COVID-19) * P(COVID-19) + P(Positive|No COVID-19) * P(Cough|No COVID-19) * P(No COVID-19)),$ which has Error \ref{error:constantSens} and Error \ref{error:uncondPrior}.

\begin{lstlisting}[ escapeinside={(*}{*)},backgroundcolor = \color{yellow}] 
Can you give an example of using Bayes rule to compute post test probability in a patient with a positive Covid test and a cough using sensitivity and specificity of the test.
\end{lstlisting}

17
\url{https://chatgpt.com/share/ee35bf98-73c9-4317-be4d-e85a5f1e3aea}

\begin{lstlisting}[ escapeinside={(*}{*)},backgroundcolor = \color{lightgray}]
ChatGPT
Certainly! Let's say we have a patient who recently took a COVID-19 test and tested positive. The patient also has a cough. We want to compute the post-test probability of the patient having COVID-19 using Bayes' rule, given the sensitivity and specificity of the test.
Let's assume the sensitivity of the COVID-19 test is 0.90, meaning that the test correctly identifies 90% of the people who actually have COVID-19. The specificity of the test is 0.95, indicating that the test correctly identifies 95% of the people who do not have COVID-19.
To use Bayes' rule, we need to know the prevalence of COVID-19 in the population. Let's assume that the prevalence is 0.02, meaning that 2% of the population has COVID-19.
Bayes' rule states:
Posterior Probability = (Prior Probability (*$\times$*) Sensitivity) / [(Prior Probability (*$\times$*) Sensitivity) + [(1 - Prior Probability) (*$\times$*) (1 - Specificity)]]
In this case, the prior probability is the prevalence of COVID-19, which is 0.02.
Let's calculate the post-test probability using the provided values:
Posterior Probability = (0.02 (*$\times$*) 0.90) / [(0.02 (*$\times$*) 0.90) + [(1 - 0.02) (*$\times$*) (1 - 0.95)]]
Posterior Probability = 0.018 / [0.018 + (0.98 (*$\times$*) 0.05)]
Posterior Probability = 0.018 / [0.018 + 0.049]
Posterior Probability = 0.018 / 0.067
Posterior Probability (*$\approx$*) 0.269 (approximately)
Therefore, the post-test probability of the patient having COVID-19, given a positive test result and a cough, is approximately 26.9%.
\end{lstlisting}
It makes Error \ref{error:constantSens} when stating ``sensitivity of the COVID-19 test is 0.90, meaning that the test correctly identifies 90\% of the people who actually have COVID-19.'' It makes Error \ref{error:uncondPrior} by defining prevalence as ``the prevalence is 0.02, meaning that 2\% of the population has COVID-19.'' It does state that it conditions on both cough and test, ``the post-test probability of the patient having COVID-19, given a positive test result and a cough.''

18
\url{https://chatgpt.com/share/3a38ca3b-a107-4859-878a-49b6c2f8cf6d}

\begin{lstlisting}[ escapeinside={(*}{*)},backgroundcolor = \color{yellow}] 
Can you give an example of using Bayes rule to compute post test probability in a patient with a positive Covid test and a cough using sensitivity and specificity of the test.
\end{lstlisting}

\begin{lstlisting}[ escapeinside={(*}{*)},backgroundcolor = \color{lightgray}]
ChatGPT
Certainly! Let's suppose we have the following information:
Sensitivity of the Covid test: 90% (0.9)
Specificity of the Covid test: 95% (0.95)
Prevalence of Covid in the population: 5% (0.05)
Now, let's use Bayes' rule to calculate the post-test probability of having Covid given a positive test result and the presence of a cough.
Step 1: Define the events:
A: Having Covid
B: Testing positive for Covid
C: Having a cough
Step 2: Calculate the prior probability (prevalence) and the complement:
P(A) = 0.05 (prevalence of Covid in the population)
P((*$\sim$*)A) = 1 - P(A) = 1 - 0.05 = 0.95 (complement of Covid prevalence)
Step 3: Calculate the conditional probabilities:
P(B|A) = Sensitivity = 0.9 (probability of testing positive given having Covid)
P(B|(*$\sim$*)A) = 1 - Specificity = 1 - 0.95 = 0.05 (probability of testing positive given not having Covid)
P(C|A) = Assuming no information about the cough's relationship with Covid, let's say it's 50% (0.5) (probability of having a cough given having Covid)
P(C|(*$\sim$*)A) = Assuming no information about the cough's relationship with Covid, let's say it's 10% (0.1) (probability of having a cough given not having Covid)
Step 4: Calculate the joint probabilities using Bayes' rule:
P(A|B (*$\cap$*) C) = (P(B|A) * P(C|A) * P(A)) / [(P(B|A) * P(C|A) * P(A)) + (P(B|(*$\sim$*)A) * P(C|(*$\sim$*)A) * P((*$\sim$*)A))]
Substituting the given values:
P(A|B (*$\cap$*) C) = (0.9 * 0.5 * 0.05) / [(0.9 * 0.5 * 0.05) + (0.05 * 0.1 * 0.95)]
Calculating further:
P(A|B (*$\cap$*) C) = 0.0225 / (0.0225 + 0.0475)
P(A|B (*$\cap$*) C) = 0.0225 / 0.07
P(A|B (*$\cap$*) C) (*$\approx$*) 0.3214
Therefore, the post-test probability of having Covid, given a positive test result and the presence of a cough, is approximately 32.14%.
\end{lstlisting}
It correctly conditions on cough and test.  It writes $\frac{P(A|B,C) = (P(B|A) * P(C|A) * P(A))}{den},$ where $den=(P(B|A) * P(C|A) * P(A)) + (P(B|\neg A) * P(C|\neg A) * P(\neg A)),$ which has Error \ref{error:constantSens} ($P(B|A)$) and Error \ref{error:uncondPrior} ($P(A)$).

19
\url{https://chatgpt.com/share/4714e019-f69c-4a81-80a1-59e12ae52866}
\begin{lstlisting}[ escapeinside={(*}{*)},backgroundcolor = \color{yellow}] 
Can you give an example of using Bayes rule to compute post test probability in a patient with a positive Covid test and a cough using sensitivity and specificity of the test.
\end{lstlisting}

\begin{lstlisting}[ escapeinside={(*}{*)},backgroundcolor = \color{lightgray}]
ChatGPT
Certainly! Let's suppose we have the following information:
		Sensitivity of the Covid test: 95% (probability of a positive test given that the patient has Covid)
		Specificity of the Covid test: 90% (probability of a negative test given that the patient does not have Covid)
		Prevalence of Covid in the population: 5% (probability that a randomly chosen person has Covid)
Now, let's calculate the post-test probability of a patient having Covid given that they have a positive test and a cough.
According to Bayes' theorem, the post-test probability can be calculated using the following formula:
P(Covid | Positive Test, Cough) = (P(Positive Test | Covid) * P(Covid)) / P(Positive Test)
Here,
P(Positive Test | Covid) = Sensitivity = 0.95
P(Covid) = Prevalence = 0.05
To calculate P(Positive Test), we need to consider both the cases where the patient has Covid and does not have Covid:
P(Positive Test) = P(Positive Test | Covid) * P(Covid) + P(Positive Test | No Covid) * P(No Covid)
P(Positive Test | No Covid) is the probability of a false positive, which is equal to 1 - Specificity.
P(No Covid) is the complement of P(Covid), which is 1 - P(Covid).
Plugging in the values, we have:
P(Positive Test) = (0.95 * 0.05) + ((1 - 0.90) * (1 - 0.05))
= 0.0475 + 0.045
= 0.0925
Finally, substituting the values into the formula:
P(Covid | Positive Test, Cough) = (0.95 * 0.05) / 0.0925
= 0.0475 / 0.0925
(*$\approx$*) 0.5135
Therefore, the post-test probability of a patient having Covid given that they have a positive test and a cough is approximately 51.35%.
\end{lstlisting}
It claims to condition on both test and cough in $P(Covid | Positive Test, Cough).$
It defines ``[s]ensitivity of the Covid test: 95\% (probability of a positive test given that the patient has Covid),'' which is Error \ref{error:constantSens}, and it uses an unconditional pre-test, $P(covid),$ which is Error \ref{error:uncondPrior}. It makes Error \ref{error:miscProb} by writing $p(positive test)$ within the denominator of $P(Covid | Positive Test, Cough) = (P(Positive Test | Covid) * P(Covid)) / P(Positive Test)$.  It is possible one could arrive at this expression by assuming independence between cough and covid, and only writing this independence on the right hand side of the expression, which would be Error \ref{error:noConditionCough} and Error \ref{error:constantSens}, but because ChatGPT still defines the posterior to depend on cough, we instead consider it to be Error \ref{error:uncondPrior} and Error \ref{error:miscProb}.

20
\url{https://chatgpt.com/share/2593e6f7-5710-4dc6-9887-edb787930c45}

\begin{lstlisting}[ escapeinside={(*}{*)},backgroundcolor = \color{yellow}] 
Can you give an example of using Bayes rule to compute post test probability in a patient with a positive Covid test and a cough using sensitivity and specificity of the test.
\end{lstlisting}

\begin{lstlisting}[ escapeinside={(*}{*)},backgroundcolor = \color{lightgray}]
ChatGPT
Certainly! Let's say we have a patient who undergoes a COVID-19 test and the result comes back positive. The patient also has a cough. We want to use Bayes' rule to compute the post-test probability of the patient having COVID-19, given the positive test result and the presence of a cough.
To apply Bayes' rule, we need to consider the sensitivity and specificity of the test. Sensitivity refers to the ability of the test to correctly identify true positive cases, while specificity refers to the ability of the test to correctly identify true negative cases. Let's assume the sensitivity of the test is 0.90 (or 90%) and the specificity is 0.95 (or 95%).
Let's denote the following:
		A: The patient has COVID-19
		B: The patient's test result is positive
		C: The patient has a cough
We want to calculate P(A|B(*$\cap$*)C), the probability of the patient having COVID-19 given a positive test result and the presence of a cough.
Using Bayes' rule, we have:
P(A|B(*$\cap$*)C) = (P(B(*$\cap$*)C|A) * P(A)) / P(B(*$\cap$*)C)
To calculate P(B(*$\cap$*)C|A), we multiply the sensitivity of the test (P(B|A)) with the probability of having a cough given COVID-19 (P(C|A)). Let's assume that P(C|A) is 0.80 (or 80%).
P(B(*$\cap$*)C|A) = P(B|A) * P(C|A)
= 0.90 * 0.80
= 0.72 (or 72%)
Now, we need to calculate P(B(*$\cap$*)C), which represents the probability of having a positive test result and a cough, irrespective of whether the patient has COVID-19.
P(B(*$\cap$*)C) = [P(B|A) * P(A)] + [P(B|(*$\neg$*)A) * P((*$\neg$*)A)]
To calculate P(B|(*$\neg$*)A), we subtract the specificity of the test (P(B|(*$\neg$*)A)) from 1. Since specificity represents the probability of a true negative result, 1 - specificity gives us the probability of a false positive result. Let's assume P(B|(*$\neg$*)A) is 0.05 (or 5%).
P(B(*$\cap$*)C) = [P(B|A) * P(A)] + [P(B|(*$\neg$*)A) * P((*$\neg$*)A)]
= [0.90 * P(A)] + [0.05 * P((*$\neg$*)A)]
To calculate P((*$\neg$*)A), the probability of not having COVID-19, we can subtract P(A) from 1. Assuming P(A) is 0.10 (or 10%).
P((*$\neg$*)A) = 1 - P(A)
= 1 - 0.10
= 0.90 (or 90%)
Now, we can substitute the values into the equation:
P(B(*$\cap$*)C) = [0.90 * 0.10] + [0.05 * 0.90]
= 0.09 + 0.045
= 0.135 (or 13.5%)
Finally, we can substitute all the values into the main equation to calculate the post-test probability:
P(A|B(*$\cap$*)C) = (P(B(*$\cap$*)C|A) * P(A)) / P(B(*$\cap$*)C)
= (0.72 * 0.10) / 0.135
(*$\approx$*) 0.533 (or 53.3%)
Therefore, the post-test
\end{lstlisting}
It correctly conditions on cough and test. In $P(A|B,C) = (P(B,C|A) * P(A)) / P(B,C).$ It uses an unconditional prior, which is Error \ref{error:uncondPrior}. However it writes $P(B,C|A) = P(B|A) * P(C|A),$ which assumes that $P(B|A,C)=P(B|A),$ which is Error \ref{error:constantSens}.
Note that $P(B,C) = [P(B|A) * P(A)] + [P(B|\neg A) * P(\neg A)]
= [0.90 * P(A)] + [0.05 * P(\neg A)]$ is valid under Error \ref{error:constantSens}. 

\subsection{Prompt engineering}
\subsubsection{Example conversation}
\url{https://chatgpt.com/share/bbacdb98-d476-4fe6-8548-2453392e9b74}
\label{app:fullconvos:promptEngineerExample}
\begin{lstlisting}[ escapeinside={(*}{*)},backgroundcolor = \color{yellow}]
User
In the following conversation, I will give you prompts. Please use the following steps to respond.
Step 1.  Determine what is the disease, what is the test result, and what is the covariate. 
Step 2. Write P(disease present | test result positive, covariates) using Bayes rule. Condition on covariate throughout. Expand the denominator using the law of total probability.  Do not make any independence assumptions.
Step 3. Write the last formula in terms of the disease, test result, and covariate given in the prompt.
Step 4. State which parts of the last formula correspond to the sensitivity and pre-test probability. 
If you understand these steps, please say only "understood".
\end{lstlisting}

\begin{lstlisting}[ escapeinside={(*}{*)},backgroundcolor = \color{lightgray}]
ChatGPT
Understood.
\end{lstlisting}

\begin{lstlisting}[ escapeinside={(*}{*)},backgroundcolor = \color{yellow}]
User
Can you give an example of using Bayes rule to compute post test probability in a patient with a positive Covid test and a cough using sensitivity and specificity of the test.
\end{lstlisting}

\begin{lstlisting}[ escapeinside={(*}{*)},backgroundcolor = \color{lightgray}]
ChatGPT
Certainly! Let's go through the steps using your example:

Step 1: Disease, Test Result, and Covariate

Disease: COVID-19
Test Result: Positive
Covariate: Cough
Step 2: Writing P(disease | test result, covariates) using Bayes rule
According to Bayes rule, we can express the posterior probability as follows:
P(disease | test result, covariates) = (P(test result | disease, covariates) * P(disease | covariates)) / P(test result | covariates)

Step 3: Writing the formula using the provided information
In this case, we want to calculate the probability of having COVID-19 given a positive test result and the presence of a cough:
P(disease = COVID-19 | test result = Positive, covariate = Cough)

Step 4: Identifying sensitivity and pre-test probability

Sensitivity corresponds to the term P(test result = Positive | disease = COVID-19, covariate = Cough)
Pre-test probability corresponds to the term P(disease = COVID-19 | covariate = Cough)
By substituting these terms into the formula from Step 3, we can calculate the post-test probability.

Please note that the specific values for sensitivity, specificity, and pre-test probability would be needed to obtain a numerical result in this example.
\end{lstlisting}
\subsubsection{Replicated conversations}
\label{app:fullconvos:promptEngineerReplicates}

1
\url{https://chatgpt.com/share/54aca7f2-aad2-426f-a9d2-e1e1c980f688}
\begin{lstlisting}[ escapeinside={(*}{*)},backgroundcolor = \color{yellow}]
In the following conversation, I will give you prompts. Please use the following steps to respond.
Step 1.  Determine what is the disease, what is the test result, and what is the covariate. 
Step 2. Write P(disease | test result, covariates) using Bayes rule. Condition on covariates throughout. Do not make any independence assumptions. Write the denominator using the law of total probability, conditional on covariates.
Step 3. Write the last formula in terms of the disease, test result, and covariate given in the prompt.
Step 4. State which parts of the last formula correspond to the sensitivity and pre-test probability. 
If you understand these steps, please say only "understood".
\end{lstlisting}

\begin{lstlisting}[ escapeinside={(*}{*)},backgroundcolor = \color{lightgray}]
ChatGPT
Understood.
\end{lstlisting}

\begin{lstlisting}[ escapeinside={(*}{*)},backgroundcolor = \color{yellow}]
Can you give an example of using Bayes rule to compute post test probability in a patient with a positive Covid test and a cough using sensitivity and specificity of the test.
\end{lstlisting}

\begin{lstlisting}[ escapeinside={(*}{*)},backgroundcolor = \color{lightgray}]
ChatGPT
Sure! Let's go through the steps using the example you provided:
Step 1:
		Disease: COVID-19
		Test result: Positive COVID test
		Covariate: Cough
Step 2:
Using Bayes' rule, we can write the formula as:
P(disease | test result, cough) = (P(test result | disease, cough) * P(disease | cough)) / P(test result | cough)
The denominator can be expanded using the law of total probability:
P(test result | cough) = P(test result | disease, cough) * P(disease | cough) + P(test result | no disease, cough) * P(no disease | cough)
Step 3:
The formula, specific to the given example, becomes:
P(COVID-19 | positive test, cough) = (P(positive test | COVID-19, cough) * P(COVID-19 | cough)) / (P(positive test | COVID-19, cough) * P(COVID-19 | cough) + P(positive test | no COVID-19, cough) * P(no COVID-19 | cough))
Step 4:
		Sensitivity: P(positive test | COVID-19, cough) corresponds to the sensitivity of the test. It represents the probability of obtaining a positive test result given that the patient has COVID-19 and a cough.
		Pre-test probability: P(COVID-19 | cough) corresponds to the pre-test probability. It represents the probability of having COVID-19 given the presence of a cough.
These are the components involved in computing the post-test probability using Bayes' rule.
Please note that to provide a specific numerical calculation, we would need the actual values of sensitivity, specificity, and pre-test probability for the given test and condition.
\end{lstlisting} 
The following is correct 
\begin{lstlisting}
P(COVID-19 | positive test, cough) = (P(positive test | COVID-19, cough) * P(COVID-19 | cough)) / (P(positive test | COVID-19, cough) * P(COVID-19 | cough) + P(positive test | no COVID-19, cough) * P(no COVID-19 | cough)).
\end{lstlisting}

2
\url{https://chatgpt.com/share/17dee359-0205-4f09-a119-1fb48402b4a4}
\begin{lstlisting}[ escapeinside={(*}{*)},backgroundcolor = \color{yellow}]
In the following conversation, I will give you prompts. Please use the following steps to respond.
Step 1.  Determine what is the disease, what is the test result, and what is the covariate. 
Step 2. Write P(disease | test result, covariates) using Bayes rule. Condition on covariates throughout. Do not make any independence assumptions. Write the denominator using the law of total probability, conditional on covariates.
Step 3. Write the last formula in terms of the disease, test result, and covariate given in the prompt.
Step 4. State which parts of the last formula correspond to the sensitivity and pre-test probability. 
If you understand these steps, please say only "understood".
\end{lstlisting}

\begin{lstlisting}[ escapeinside={(*}{*)},backgroundcolor = \color{lightgray}]
ChatGPT
Understood.
\end{lstlisting}

\begin{lstlisting}[ escapeinside={(*}{*)},backgroundcolor = \color{yellow}]
Can you give an example of using Bayes rule to compute post test probability in a patient with a positive Covid test and a cough using sensitivity and specificity of the test.
\end{lstlisting}

\begin{lstlisting}[ escapeinside={(*}{*)},backgroundcolor = \color{lightgray}]
ChatGPT
Certainly! Let's walk through the steps using the example you provided:
Step 1:
		Disease: Covid
		Test result: Positive
		Covariate: Cough
Step 2:
The formula using Bayes' rule to compute the post-test probability is:
P(disease | test result, covariate) = (P(test result | disease, covariate) * P(disease | covariate)) / P(test result | covariate)
To compute the denominator, we can use the law of total probability:
P(test result | covariate) = P(test result | disease, covariate) * P(disease | covariate) + P(test result | no disease, covariate) * P(no disease | covariate)
Step 3:
Plugging in the disease, test result, and covariate from the prompt, the formula becomes:
P(Covid | positive test, cough) = (P(positive test | Covid, cough) * P(Covid | cough)) / P(positive test | cough)
Step 4:
		The sensitivity of the test corresponds to P(positive test | Covid, cough), as it measures the probability of obtaining a positive test result given that the patient has Covid and cough.
		The pre-test probability, which is the probability of having the disease before considering the test result, corresponds to P(Covid | cough).
By substituting the specific values and using the sensitivity and pre-test probability, we can calculate the post-test probability.
\end{lstlisting} 
The following is correct, but the denominator is not expanded.
\begin{lstlisting}
P(Covid | positive test, cough) = (P(positive test | Covid, cough) * P(Covid | cough)) / P(positive test | cough).
\end{lstlisting}

3
\url{https://chatgpt.com/share/3c7aa36d-3386-42fd-9aa0-65b400a2d49a}

\begin{lstlisting}[ escapeinside={(*}{*)},backgroundcolor = \color{yellow}]
In the following conversation, I will give you prompts. Please use the following steps to respond.
Step 1.  Determine what is the disease, what is the test result, and what is the covariate. 
Step 2. Write P(disease | test result, covariates) using Bayes rule. Condition on covariates throughout. Do not make any independence assumptions. Write the denominator using the law of total probability, conditional on covariates.
Step 3. Write the last formula in terms of the disease, test result, and covariate given in the prompt.
Step 4. State which parts of the last formula correspond to the sensitivity and pre-test probability. 
If you understand these steps, please say only "understood".
\end{lstlisting}

\begin{lstlisting}[ escapeinside={(*}{*)},backgroundcolor = \color{lightgray}]
ChatGPT
Understood.
\end{lstlisting}

\begin{lstlisting}[ escapeinside={(*}{*)},backgroundcolor = \color{yellow}]
Can you give an example of using Bayes rule to compute post test probability in a patient with a positive Covid test and a cough using sensitivity and specificity of the test.
\end{lstlisting}

\begin{lstlisting}[ escapeinside={(*}{*)},backgroundcolor = \color{lightgray}]
ChatGPT
Certainly! Let's go through the steps using the example you provided:
Step 1:
		Disease: Covid-19
		Test result: Positive Covid test
		Covariate: Cough
Step 2:
We want to compute P(Covid | Positive test, Cough) using Bayes rule. Using the law of total probability, we can express the denominator as follows:
P(Positive test, Cough) = P(Positive test | Covid, Cough) * P(Covid | Cough) + P(Positive test | No Covid, Cough) * P(No Covid | Cough)
Step 3:
Applying Bayes rule and plugging in the values from the example:
P(Covid | Positive test, Cough) = P(Positive test | Covid, Cough) * P(Covid | Cough) / P(Positive test, Cough)
Step 4:
		Sensitivity of the test corresponds to P(Positive test | Covid, Cough), as it represents the probability of getting a positive test result given that the patient has Covid and cough.
		Pre-test probability refers to P(Covid | Cough), which is the probability of having Covid given that the patient has a cough.
By substituting these values into the formula, we can compute the post-test probability of having Covid given a positive test result and a cough.
Please note that the actual values for sensitivity, specificity, and pre-test probability would need to be provided for a specific calculation.
\end{lstlisting} 
In the following expression,
\begin{lstlisting}
    P(Covid | Positive test, Cough) = P(Positive test | Covid, Cough) * P(Covid | Cough) / P(Positive test, Cough),
\end{lstlisting}
 Error \ref{error:miscProb} is made.
 The term $P(Positive test, Cough)$ in the denominator should be P(positive test|cough). However, ChatGPT does correctly write the right hand side of the expanded expression,
 \begin{lstlisting}
     P(Positive test, Cough) = P(Positive test | Covid, Cough) * P(Covid | Cough) + P(Positive test | No Covid, Cough) * P(No Covid | Cough). 
     \end{lstlisting}
     
4
\url{https://chatgpt.com/share/a3084ac0-c248-4a89-9a17-47f21b24e8de}
\begin{lstlisting}[ escapeinside={(*}{*)},backgroundcolor = \color{yellow}]
In the following conversation, I will give you prompts. Please use the following steps to respond.
Step 1.  Determine what is the disease, what is the test result, and what is the covariate. 
Step 2. Write P(disease | test result, covariates) using Bayes rule. Condition on covariates throughout. Do not make any independence assumptions. Write the denominator using the law of total probability, conditional on covariates.
Step 3. Write the last formula in terms of the disease, test result, and covariate given in the prompt.
Step 4. State which parts of the last formula correspond to the sensitivity and pre-test probability. 
If you understand these steps, please say only "understood".
\end{lstlisting}

\begin{lstlisting}[ escapeinside={(*}{*)},backgroundcolor = \color{lightgray}]
ChatGPT
Understood.
\end{lstlisting}

\begin{lstlisting}[ escapeinside={(*}{*)},backgroundcolor = \color{yellow}]
Can you give an example of using Bayes rule to compute post test probability in a patient with a positive Covid test and a cough using sensitivity and specificity of the test.
\end{lstlisting}

\begin{lstlisting}[ escapeinside={(*}{*)},backgroundcolor = \color{lightgray}]
ChatGPT
Certainly! Let's go through the steps to compute the post-test probability using Bayes' rule:
Step 1:
		Disease: COVID-19
		Test result: Positive
		Covariate: Cough
Step 2:
Bayes' rule states:
P(disease | test result, covariates) = [P(test result | disease, covariates) * P(disease | covariates)] / P(test result | covariates)
The denominator, P(test result | covariates), can be expanded using the law of total probability as:
P(test result | covariates) = P(test result | disease, covariates) * P(disease | covariates) + P(test result | no disease, covariates) * P(no disease | covariates)
Step 3:
Substituting the disease, test result, and covariate given in the prompt:
P(disease | positive test, cough) = [P(positive test | disease, cough) * P(disease | cough)] / [P(positive test | disease, cough) * P(disease | cough) + P(positive test | no disease, cough) * P(no disease | cough)]
Step 4:
		The sensitivity of the test, P(positive test | disease, cough), corresponds to the probability of obtaining a positive test result when the disease is present and the patient has a cough. It captures the true positive rate.
		The pre-test probability, P(disease | cough), corresponds to the probability of having the disease given the covariate (in this case, cough) before conducting the test.
I hope this example helps illustrate the application of Bayes' rule in computing post-test probabilities. Let me know if you have any further questions!
\end{lstlisting} 
The following is correct. 
\begin{lstlisting}
    P(disease | positive test, cough) = [P(positive test | disease, cough) * P(disease | cough)] / [P(positive test | disease, cough) * P(disease | cough) + P(positive test | no disease, cough) * P(no disease | cough)]
\end{lstlisting}

5
\url{https://chatgpt.com/share/4f1dbe09-26b4-4537-a300-03a33040489b}
\begin{lstlisting}[ escapeinside={(*}{*)},backgroundcolor = \color{yellow}]
In the following conversation, I will give you prompts. Please use the following steps to respond.
Step 1.  Determine what is the disease, what is the test result, and what is the covariate. 
Step 2. Write P(disease | test result, covariates) using Bayes rule. Condition on covariates throughout. Do not make any independence assumptions. Write the denominator using the law of total probability, conditional on covariates.
Step 3. Write the last formula in terms of the disease, test result, and covariate given in the prompt.
Step 4. State which parts of the last formula correspond to the sensitivity and pre-test probability. 
If you understand these steps, please say only "understood".
\end{lstlisting}

\begin{lstlisting}[ escapeinside={(*}{*)},backgroundcolor = \color{lightgray}]
ChatGPT
Understood. Please provide the prompts, and I will follow the steps you provided to respond.
\end{lstlisting}

\begin{lstlisting}[ escapeinside={(*}{*)},backgroundcolor = \color{yellow}]
Can you give an example of using Bayes rule to compute post test probability in a patient with a positive Covid test and a cough using sensitivity and specificity of the test.
\end{lstlisting}

\begin{lstlisting}[ escapeinside={(*}{*)},backgroundcolor = \color{lightgray}]
ChatGPT
Certainly! Let's apply Bayes' rule to compute the post-test probability of having Covid in a patient with a positive test and a cough, using the sensitivity and specificity of the test.
Step 1:
Disease: Covid
Test result: Positive
Covariate: Cough
Step 2:
We can write Bayes' rule as follows:
P(Covid | Positive Test, Cough) = (P(Positive Test | Covid, Cough) * P(Covid | Cough)) / P(Positive Test | Cough)
To compute the denominator, we can use the law of total probability:
P(Positive Test | Cough) = P(Positive Test | Covid, Cough) * P(Covid | Cough) + P(Positive Test | No Covid, Cough) * P(No Covid | Cough)
Step 3:
Rewriting the formula using the specific values given in the prompt:
P(Covid | Positive Test, Cough) = (P(Positive Test | Covid, Cough) * P(Covid | Cough)) / (P(Positive Test | Covid, Cough) * P(Covid | Cough) + P(Positive Test | No Covid, Cough) * P(No Covid | Cough))
Step 4:
In the last formula:
		Sensitivity (True Positive Rate) corresponds to P(Positive Test | Covid, Cough).
		Pre-test probability corresponds to P(Covid | Cough).
Please note that to calculate the post-test probability, we would need to know the values of sensitivity, specificity, and pre-test probability.
\end{lstlisting} 
The following is correct.
\begin{lstlisting}
    P(Covid | Positive Test, Cough) = (P(Positive Test | Covid, Cough) * P(Covid | Cough)) / (P(Positive Test | Covid, Cough) * P(Covid | Cough) + P(Positive Test | No Covid, Cough) * P(No Covid | Cough))
\end{lstlisting}

6
\url{https://chatgpt.com/share/65212075-b4a1-4056-9b8d-9c51058ce85e}
\begin{lstlisting}[ escapeinside={(*}{*)},backgroundcolor = \color{yellow}]
In the following conversation, I will give you prompts. Please use the following steps to respond.
Step 1.  Determine what is the disease, what is the test result, and what is the covariate. 
Step 2. Write P(disease | test result, covariates) using Bayes rule. Condition on covariates throughout. Do not make any independence assumptions. Write the denominator using the law of total probability, conditional on covariates.
Step 3. Write the last formula in terms of the disease, test result, and covariate given in the prompt.
Step 4. State which parts of the last formula correspond to the sensitivity and pre-test probability. 
If you understand these steps, please say only "understood".
\end{lstlisting}

\begin{lstlisting}[ escapeinside={(*}{*)},backgroundcolor = \color{lightgray}]
ChatGPT
Understood.
\end{lstlisting}

\begin{lstlisting}[ escapeinside={(*}{*)},backgroundcolor = \color{yellow}]
Can you give an example of using Bayes rule to compute post test probability in a patient with a positive Covid test and a cough using sensitivity and specificity of the test.
\end{lstlisting}

\begin{lstlisting}[ escapeinside={(*}{*)},backgroundcolor = \color{lightgray}]
ChatGPT
Sure! Let's go through the steps using the given example:
Step 1:
		Disease: COVID-19
		Test result: Positive
		Covariate: Cough
Step 2:
We want to compute P(disease | test result, cough) using Bayes' rule. According to Bayes' rule, we have:
P(disease | test result, cough) = (P(test result | disease, cough) * P(disease | cough)) / P(test result | cough)
To calculate the denominator, we can use the law of total probability, conditioning on the covariate (cough):
P(test result | cough) = P(test result | disease, cough) * P(disease | cough) + P(test result | no disease, cough) * P(no disease | cough)
Step 3:
Now, let's write the formula in terms of the given example:
P(COVID-19 | Positive test result, cough) = (P(Positive test result | COVID-19, cough) * P(COVID-19 | cough)) / P(Positive test result | cough)
Step 4:
In the last formula:
		Sensitivity corresponds to P(Positive test result | COVID-19, cough)
		Pre-test probability corresponds to P(COVID-19 | cough)
These values are needed to compute the post-test probability of COVID-19 given a positive test result and the presence of a cough.
Please note that to provide a more accurate computation, we would need specific values for sensitivity, specificity, and pre-test probability.
\end{lstlisting} 
The following is correct.
\begin{lstlisting}
P(COVID-19 | Positive test result, cough) = (P(Positive test result | COVID-19, cough) * P(COVID-19 | cough)) / P(Positive test result | cough). It does expand the denominator P(test result | cough) = P(test result | disease, cough) * P(disease | cough) + P(test result | no disease, cough) * P(no disease | cough).
\end{lstlisting}

7
\url{https://chatgpt.com/share/2483b54e-7867-4ee5-84f9-cf3fcf56e1e7}
\begin{lstlisting}[ escapeinside={(*}{*)},backgroundcolor = \color{yellow}]
In the following conversation, I will give you prompts. Please use the following steps to respond.
Step 1.  Determine what is the disease, what is the test result, and what is the covariate. 
Step 2. Write P(disease | test result, covariates) using Bayes rule. Condition on covariates throughout. Do not make any independence assumptions. Write the denominator using the law of total probability, conditional on covariates.
Step 3. Write the last formula in terms of the disease, test result, and covariate given in the prompt.
Step 4. State which parts of the last formula correspond to the sensitivity and pre-test probability. 
If you understand these steps, please say only "understood".
\end{lstlisting}

\begin{lstlisting}[ escapeinside={(*}{*)},backgroundcolor = \color{lightgray}]
ChatGPT
Understood.
\end{lstlisting}

\begin{lstlisting}[ escapeinside={(*}{*)},backgroundcolor = \color{yellow}]
Can you give an example of using Bayes rule to compute post test probability in a patient with a positive Covid test and a cough using sensitivity and specificity of the test.
\end{lstlisting}

\begin{lstlisting}[ escapeinside={(*}{*)},backgroundcolor = \color{lightgray}]
ChatGPT
Certainly! Let's go through the steps using the example you provided.
Step 1: Determine the disease, test result, and covariate.
Disease: COVID-19
Test result: Positive COVID-19 test
Covariate: Presence of cough
Step 2: Write P(disease | test result, covariates) using Bayes rule.
P(disease | test result, covariates) = [P(test result | disease, covariates) * P(disease | covariates)] / P(test result | covariates)
Step 3: Write the formula in terms of the given variables.
P(COVID-19 | positive test, cough) = [P(positive test | COVID-19, cough) * P(COVID-19 | cough)] / P(positive test | cough)
Step 4: Identify sensitivity and pre-test probability.
In this case:
		Sensitivity: P(positive test | COVID-19, cough)
		Pre-test probability: P(COVID-19 | cough)
To compute the post-test probability, we would need the values of sensitivity (the probability of a positive test given the disease and cough) and pre-test probability (the probability of having COVID-19 given the presence of a cough) based on available data or prior knowledge.
Please note that the values of sensitivity and pre-test probability may vary based on the specific characteristics of the test and the population being considered.
\end{lstlisting} 
The following is correct, but the denominator is not expanded.
\begin{lstlisting}
P(COVID-19 | positive test, cough) = [P(positive test | COVID-19, cough) * P(COVID-19 | cough)] / P(positive test | cough).
\end{lstlisting}

8
\url{https://chatgpt.com/share/e309b433-28f2-4015-9210-47e789d1cbec}
\begin{lstlisting}[ escapeinside={(*}{*)},backgroundcolor = \color{yellow}]
In the following conversation, I will give you prompts. Please use the following steps to respond.
Step 1.  Determine what is the disease, what is the test result, and what is the covariate. 
Step 2. Write P(disease | test result, covariates) using Bayes rule. Condition on covariates throughout. Do not make any independence assumptions. Write the denominator using the law of total probability, conditional on covariates.
Step 3. Write the last formula in terms of the disease, test result, and covariate given in the prompt.
Step 4. State which parts of the last formula correspond to the sensitivity and pre-test probability. 
If you understand these steps, please say only "understood".
\end{lstlisting}

\begin{lstlisting}[ escapeinside={(*}{*)},backgroundcolor = \color{lightgray}]
ChatGPT
Understood.
\end{lstlisting}

\begin{lstlisting}[ escapeinside={(*}{*)},backgroundcolor = \color{yellow}]
Can you give an example of using Bayes rule to compute post test probability in a patient with a positive Covid test and a cough using sensitivity and specificity of the test.
\end{lstlisting}

\begin{lstlisting}[ escapeinside={(*}{*)},backgroundcolor = \color{lightgray}]
ChatGPT
Certainly! Let's go through the steps using the example you provided:
Step 1: Disease, Test Result, and Covariate
		Disease: COVID-19
		Test Result: Positive COVID-19 test
		Covariate: Cough
Step 2: Applying Bayes Rule
We want to compute P(disease | test result, cough) using Bayes rule. The formula for Bayes rule is:
P(disease | test result, cough) = (P(test result | disease, cough) * P(disease | cough)) / P(test result | cough)
We need to calculate the three probabilities involved in this formula.
Step 3: Formula in Terms of the Prompt
P(disease | test result, cough) = (P(test result | disease, cough) * P(disease | cough)) / P(test result | cough)
Step 4: Sensitivity and Pre-Test Probability
In this context, the sensitivity of the test corresponds to P(test result | disease, cough), which represents the probability of testing positive given that the patient has COVID-19 and a cough. The pre-test probability is P(disease | cough), which is the probability of having COVID-19 given the presence of a cough.
So, in summary:
		Sensitivity: P(test result | disease, cough)
		Pre-Test Probability: P(disease | cough)
I hope this clarifies the steps and how the sensitivity and pre-test probability fit into the formula. Let me know if you have any further questions!
\end{lstlisting} 
The following is correct, but the denominator is not expanded and disease is not named as Covid-19.
\begin{lstlisting}
P(disease | test result, cough) = (P(test result | disease, cough) * P(disease | cough)) / P(test result | cough)
\end{lstlisting}

9
\url{https://chatgpt.com/share/ea85750d-4267-4c6f-8976-6adc3ba80007}
\begin{lstlisting}[ escapeinside={(*}{*)},backgroundcolor = \color{yellow}]
In the following conversation, I will give you prompts. Please use the following steps to respond.
Step 1.  Determine what is the disease, what is the test result, and what is the covariate. 
Step 2. Write P(disease | test result, covariates) using Bayes rule. Condition on covariates throughout. Do not make any independence assumptions. Write the denominator using the law of total probability, conditional on covariates.
Step 3. Write the last formula in terms of the disease, test result, and covariate given in the prompt.
Step 4. State which parts of the last formula correspond to the sensitivity and pre-test probability. 
If you understand these steps, please say only "understood".
\end{lstlisting}

\begin{lstlisting}[ escapeinside={(*}{*)},backgroundcolor = \color{lightgray}]
ChatGPT
Understood.
\end{lstlisting}

\begin{lstlisting}[ escapeinside={(*}{*)},backgroundcolor = \color{yellow}]
Can you give an example of using Bayes rule to compute post test probability in a patient with a positive Covid test and a cough using sensitivity and specificity of the test.
\end{lstlisting}

\begin{lstlisting}[ escapeinside={(*}{*)},backgroundcolor = \color{lightgray}]
ChatGPT
Sure! Let's go through the steps using Bayes' rule to compute the post-test probability in a patient with a positive COVID test and a cough, given the sensitivity and specificity of the test.
Step 1:
		Disease: COVID-19
		Test result: Positive
		Covariate: Cough
Step 2:
We can write the formula using Bayes' rule as follows:
P(Disease | Test result, Cough) = [P(Test result | Disease, Cough) * P(Disease | Cough)] / P(Test result | Cough)
Using the law of total probability, we can expand the denominator as:
P(Test result | Cough) = P(Test result | Disease, Cough) * P(Disease | Cough) + P(Test result | No disease, Cough) * P(No disease | Cough)
Step 3:
Substituting the values from the prompt, the formula becomes:
P(Disease | Positive test, Cough) = [P(Positive test | Disease, Cough) * P(Disease | Cough)] / [P(Positive test | Disease, Cough) * P(Disease | Cough) + P(Positive test | No disease, Cough) * P(No disease | Cough)]
Step 4:
		Sensitivity corresponds to P(Positive test | Disease, Cough), as it represents the probability of obtaining a positive test result given that the patient has the disease and a cough.
		Pre-test probability corresponds to P(Disease | Cough), which is the probability of the patient having the disease given that they have a cough before considering the test result.
These are the key components used to calculate the post-test probability using Bayes' rule.
I hope this example clarifies the steps involved. Let me know if you have any further questions!
\end{lstlisting} 
The following is correct.
\begin{lstlisting}
P(Disease | Positive test, Cough) = [P(Positive test | Disease, Cough) * P(Disease | Cough)] / [P(Positive test | Disease, Cough) * P(Disease | Cough) + P(Positive test | No disease, Cough) * P(No disease | Cough)]
\end{lstlisting}

10
\url{https://chatgpt.com/share/e986d64a-fac2-4e98-b8dd-2a2abf56912b}
\begin{lstlisting}[ escapeinside={(*}{*)},backgroundcolor = \color{yellow}]
In the following conversation, I will give you prompts. Please use the following steps to respond.
Step 1.  Determine what is the disease, what is the test result, and what is the covariate. 
Step 2. Write P(disease | test result, covariates) using Bayes rule. Condition on covariates throughout. Do not make any independence assumptions. Write the denominator using the law of total probability, conditional on covariates.
Step 3. Write the last formula in terms of the disease, test result, and covariate given in the prompt.
Step 4. State which parts of the last formula correspond to the sensitivity and pre-test probability. 
If you understand these steps, please say only "understood".
\end{lstlisting}

\begin{lstlisting}[ escapeinside={(*}{*)},backgroundcolor = \color{lightgray}]
ChatGPT
Understood.
\end{lstlisting}

\begin{lstlisting}[ escapeinside={(*}{*)},backgroundcolor = \color{yellow}]
Can you give an example of using Bayes rule to compute post test probability in a patient with a positive Covid test and a cough using sensitivity and specificity of the test.
\end{lstlisting}

\begin{lstlisting}[ escapeinside={(*}{*)},backgroundcolor = \color{lightgray}]
ChatGPT
Certainly! Let's go through the steps using the example you provided:

Step 1:

Disease: COVID-19
Test Result: Positive
Covariate: Cough
Step 2:
We can write the formula using Bayes' rule as follows:

P(Disease | Test Result, Covariate) = (P(Test Result | Disease, Covariate) * P(Disease | Covariate)) / P(Test Result | Covariate)

Using the law of total probability, we can expand the denominator as:

P(Test Result | Covariate) = P(Test Result | Disease, Covariate) * P(Disease | Covariate) + P(Test Result | No Disease, Covariate) * P(No Disease | Covariate)

Step 3:
Let's rewrite the formula in terms of the given example:

P(COVID-19 | Positive Test, Cough) = (P(Positive Test | COVID-19, Cough) * P(COVID-19 | Cough)) / (P(Positive Test | COVID-19, Cough) * P(COVID-19 | Cough) + P(Positive Test | No COVID-19, Cough) * P(No COVID-19 | Cough))

Step 4:
In the last formula:

The sensitivity of the test (P(Positive Test | COVID-19, Cough)) corresponds to the probability of getting a positive test result given that the patient has COVID-19 and a cough.
The pre-test probability (P(COVID-19 | Cough)) corresponds to the probability of having COVID-19 given the presence of a cough.
If you provide the specific values for sensitivity, specificity, and pre-test probability, I can help you calculate the post-test probability.
\end{lstlisting} 
The following is correct.
\begin{lstlisting}
P(COVID-19 | Positive Test, Cough) = (P(Positive Test | COVID-19, Cough) * P(COVID-19 | Cough)) / (P(Positive Test | COVID-19, Cough) * P(COVID-19 | Cough) + P(Positive Test | No COVID-19, Cough) * P(No COVID-19 | Cough)).
\end{lstlisting}

11
\url{https://chatgpt.com/share/a419cef6-24bd-4777-913e-8b7ac7f1cc69}
\begin{lstlisting}[ escapeinside={(*}{*)},backgroundcolor = \color{yellow}]
In the following conversation, I will give you prompts. Please use the following steps to respond.
Step 1.  Determine what is the disease, what is the test result, and what is the covariate. 
Step 2. Write P(disease | test result, covariates) using Bayes rule. Condition on covariates throughout. Do not make any independence assumptions. Write the denominator using the law of total probability, conditional on covariates.
Step 3. Write the last formula in terms of the disease, test result, and covariate given in the prompt.
Step 4. State which parts of the last formula correspond to the sensitivity and pre-test probability. 
If you understand these steps, please say only "understood".
\end{lstlisting}

\begin{lstlisting}[ escapeinside={(*}{*)},backgroundcolor = \color{lightgray}]
ChatGPT
Understood. I will follow the steps you provided to respond to the prompts.
\end{lstlisting}

\begin{lstlisting}[ escapeinside={(*}{*)},backgroundcolor = \color{yellow}]
Can you give an example of using Bayes rule to compute post test probability in a patient with a positive Covid test and a cough using sensitivity and specificity of the test.
\end{lstlisting}

\begin{lstlisting}[ escapeinside={(*}{*)},backgroundcolor = \color{lightgray}]
ChatGPT
Sure! Let's go through the steps using Bayes' rule to compute the post-test probability in a patient with a positive Covid test and a cough, given the sensitivity and specificity of the test.
Step 1:
		Disease: Covid
		Test Result: Positive
		Covariate: Cough
Step 2:
We can write the formula using Bayes' rule as follows:
P(Covid | Positive test, Cough) = (P(Positive test | Covid, Cough) * P(Covid | Cough)) / P(Positive test | Cough)
The denominator can be expanded using the law of total probability:
P(Positive test | Cough) = P(Positive test | Covid, Cough) * P(Covid | Cough) + P(Positive test | No Covid, Cough) * P(No Covid | Cough)
Step 3:
P(Covid | Positive test, Cough) = (P(Positive test | Covid, Cough) * P(Covid | Cough)) / (P(Positive test | Covid, Cough) * P(Covid | Cough) + P(Positive test | No Covid, Cough) * P(No Covid | Cough))
Step 4:
		The term P(Covid | Cough) represents the pre-test probability or the probability of having Covid given the presence of a cough.
		The term P(Positive test | Covid, Cough) represents the sensitivity of the test.
		The entire numerator (P(Positive test | Covid, Cough) * P(Covid | Cough)) represents the joint probability of having a positive test and Covid.
		The denominator (P(Positive test | Covid, Cough) * P(Covid | Cough) + P(Positive test | No Covid, Cough) * P(No Covid | Cough)) represents the sum of the joint probabilities of having a positive test and Covid, and having a positive test and not having Covid.
\end{lstlisting}
The following is correct.
\begin{lstlisting}
P(Covid | Positive test, Cough) = (P(Positive test | Covid, Cough) * P(Covid | Cough)) / (P(Positive test | Covid, Cough) * P(Covid | Cough) + P(Positive test | No Covid, Cough) * P(No Covid | Cough))
\end{lstlisting}

12
\url{https://chatgpt.com/share/05ec44e4-02df-43f8-a2fc-4ff4a46058e1}

\begin{lstlisting}[ escapeinside={(*}{*)},backgroundcolor = \color{yellow}]
In the following conversation, I will give you prompts. Please use the following steps to respond.
Step 1.  Determine what is the disease, what is the test result, and what is the covariate. 
Step 2. Write P(disease | test result, covariates) using Bayes rule. Condition on covariates throughout. Do not make any independence assumptions. Write the denominator using the law of total probability, conditional on covariates.
Step 3. Write the last formula in terms of the disease, test result, and covariate given in the prompt.
Step 4. State which parts of the last formula correspond to the sensitivity and pre-test probability. 
If you understand these steps, please say only "understood".
\end{lstlisting}

\begin{lstlisting}[ escapeinside={(*}{*)},backgroundcolor = \color{lightgray}]
ChatGPT
Understood.
\end{lstlisting}

\begin{lstlisting}[ escapeinside={(*}{*)},backgroundcolor = \color{yellow}]
Can you give an example of using Bayes rule to compute post test probability in a patient with a positive Covid test and a cough using sensitivity and specificity of the test.
\end{lstlisting}

\begin{lstlisting}[ escapeinside={(*}{*)},backgroundcolor = \color{lightgray}]
ChatGPT
Certainly! Let's go through the steps using the example you provided:
Step 1:
		Disease: Covid
		Test result: Positive
		Covariate: Cough
Step 2:
The Bayes rule formula is:
P(disease | test result, covariates) = (P(test result | disease, covariates) * P(disease | covariates)) / P(test result | covariates)
Let's break down the terms:
		P(disease | test result, covariates): Posterior probability of having the disease given the test result and covariates.
		P(test result | disease, covariates): Sensitivity of the test, which is the probability of getting a positive test result given that the patient has the disease and the covariates.
		P(disease | covariates): Pre-test probability of having the disease given the covariates, also known as the baseline probability.
		P(test result | covariates): The probability of getting a positive test result given the covariates, which can be calculated using the sensitivity and specificity of the test.
Step 3:
In our example, the formula would be:
P(Covid | Positive test, Cough) = (P(Positive test | Covid, Cough) * P(Covid | Cough)) / P(Positive test | Cough)
Step 4:
		Sensitivity corresponds to P(Positive test | Covid, Cough), as it represents the probability of getting a positive test result given that the patient has Covid and cough.
		Pre-test probability corresponds to P(Covid | Cough), which is the probability of having Covid given that the patient has a cough.
I hope this clarifies the steps involved in using Bayes rule to compute post-test probability. Let me know if you have any further questions!
\end{lstlisting}
The following is correct, but the denominator is not expanded.
\begin{lstlisting}
P(Covid | Positive test, Cough) = (P(Positive test | Covid, Cough) * P(Covid | Cough)) / P(Positive test | Cough)
\end{lstlisting}

13
\url{https://chatgpt.com/share/4eca4b5a-773e-47cc-81d4-c1e2f7ce1071}
\begin{lstlisting}[ escapeinside={(*}{*)},backgroundcolor = \color{yellow}]
In the following conversation, I will give you prompts. Please use the following steps to respond.
Step 1.  Determine what is the disease, what is the test result, and what is the covariate. 
Step 2. Write P(disease | test result, covariates) using Bayes rule. Condition on covariates throughout. Do not make any independence assumptions. Write the denominator using the law of total probability, conditional on covariates.
Step 3. Write the last formula in terms of the disease, test result, and covariate given in the prompt.
Step 4. State which parts of the last formula correspond to the sensitivity and pre-test probability. 
If you understand these steps, please say only "understood".
\end{lstlisting}

\begin{lstlisting}[ escapeinside={(*}{*)},backgroundcolor = \color{lightgray}]
ChatGPT
Understood.
\end{lstlisting}

\begin{lstlisting}[ escapeinside={(*}{*)},backgroundcolor = \color{yellow}]
Can you give an example of using Bayes rule to compute post test probability in a patient with a positive Covid test and a cough using sensitivity and specificity of the test.
\end{lstlisting}

\begin{lstlisting}[ escapeinside={(*}{*)},backgroundcolor = \color{lightgray}]
ChatGPT
Certainly! Let's go through the steps using your example:
Step 1:
		Disease: Covid
		Test result: Positive
		Covariate: Cough
Step 2:
We want to compute P(Covid | Positive test, Cough) using Bayes' rule. Rewriting the formula, we have:
P(Covid | Positive test, Cough) = (P(Positive test | Covid, Cough) * P(Covid | Cough)) / P(Positive test | Cough)
Using the law of total probability, we can expand the denominator as:
P(Positive test | Cough) = P(Positive test | Covid, Cough) * P(Covid | Cough) + P(Positive test | No Covid, Cough) * P(No Covid | Cough)
Step 3:
Now, let's rewrite the formula in terms of the given information:
P(Covid | Positive test, Cough) = (P(Positive test | Covid, Cough) * P(Covid | Cough)) / (P(Positive test | Covid, Cough) * P(Covid | Cough) + P(Positive test | No Covid, Cough) * P(No Covid | Cough))
Step 4:
In the last formula:
		The sensitivity (probability of a positive test given the presence of the disease) corresponds to P(Positive test | Covid, Cough).
		The pre-test probability (probability of having the disease before the test) corresponds to P(Covid | Cough).
If you have specific values for sensitivity, specificity, and pre-test probability, I can help you calculate the post-test probability using the formula.
\end{lstlisting}
The following is correct.
\begin{lstlisting}
P(Covid | Positive test, Cough) = (P(Positive test | Covid, Cough) * P(Covid | Cough)) / (P(Positive test | Covid, Cough) * P(Covid | Cough) + P(Positive test | No Covid, Cough) * P(No Covid | Cough))
\end{lstlisting}

14
\url{https://chatgpt.com/share/e2a208e5-28e7-40a9-a724-788bf830b0bb}
\begin{lstlisting}[ escapeinside={(*}{*)},backgroundcolor = \color{yellow}]
In the following conversation, I will give you prompts. Please use the following steps to respond.
Step 1.  Determine what is the disease, what is the test result, and what is the covariate. 
Step 2. Write P(disease | test result, covariates) using Bayes rule. Condition on covariates throughout. Do not make any independence assumptions. Write the denominator using the law of total probability, conditional on covariates.
Step 3. Write the last formula in terms of the disease, test result, and covariate given in the prompt.
Step 4. State which parts of the last formula correspond to the sensitivity and pre-test probability. 
If you understand these steps, please say only "understood".
\end{lstlisting}

\begin{lstlisting}[ escapeinside={(*}{*)},backgroundcolor = \color{lightgray}]
ChatGPT
Understood.
\end{lstlisting}

\begin{lstlisting}[ escapeinside={(*}{*)},backgroundcolor = \color{yellow}]
Can you give an example of using Bayes rule to compute post test probability in a patient with a positive Covid test and a cough using sensitivity and specificity of the test.
\end{lstlisting}

\begin{lstlisting}[ escapeinside={(*}{*)},backgroundcolor = \color{lightgray}]
ChatGPT
Certainly! Let's go through the steps using the example you provided:
Step 1:
		Disease: Covid
		Test result: Positive
		Covariate: Cough
Step 2:
We want to calculate P(Covid | Positive test, Cough) using Bayes' rule.
According to Bayes' rule, we have:
P(Covid | Positive test, Cough) = [P(Positive test | Covid, Cough) * P(Covid | Cough)] / P(Positive test | Cough)
We'll break down each term:
		P(Positive test | Covid, Cough): Sensitivity of the test
		P(Covid | Cough): Pre-test probability of having Covid given the presence of a cough
		P(Positive test | Cough): Total probability of testing positive given the presence of a cough (includes true positives and false positives)
Step 3:
Plugging in the terms from Step 2 into the formula, we get:
P(Covid | Positive test, Cough) = [Sensitivity * P(Covid | Cough)] / P(Positive test | Cough)
Step 4:
		Sensitivity: Corresponds to P(Positive test | Covid, Cough), as it represents the probability of testing positive given that the person has Covid and a cough.
		Pre-test probability: Corresponds to P(Covid | Cough), as it represents the probability of having Covid given the presence of a cough.
I hope this clarifies the steps involved in using Bayes' rule to compute the post-test probability. Let me know if you have any further questions!
\end{lstlisting}
The following is correct, but the denominator is not expanded.
\begin{lstlisting}
P(Covid | Positive test, Cough) = [P(Positive test | Covid, Cough) * P(Covid | Cough)] / P(Positive test | Cough)
\end{lstlisting}

15
\url{https://chatgpt.com/share/d8697a12-015d-4cfb-b550-e80d1fb5c766}
\begin{lstlisting}[ escapeinside={(*}{*)},backgroundcolor = \color{yellow}]
In the following conversation, I will give you prompts. Please use the following steps to respond.
Step 1.  Determine what is the disease, what is the test result, and what is the covariate. 
Step 2. Write P(disease | test result, covariates) using Bayes rule. Condition on covariates throughout. Do not make any independence assumptions. Write the denominator using the law of total probability, conditional on covariates.
Step 3. Write the last formula in terms of the disease, test result, and covariate given in the prompt.
Step 4. State which parts of the last formula correspond to the sensitivity and pre-test probability. 
If you understand these steps, please say only "understood".
\end{lstlisting}

\begin{lstlisting}[ escapeinside={(*}{*)},backgroundcolor = \color{lightgray}]
ChatGPT
Understood.
\end{lstlisting}

\begin{lstlisting}[ escapeinside={(*}{*)},backgroundcolor = \color{yellow}]
Can you give an example of using Bayes rule to compute post test probability in a patient with a positive Covid test and a cough using sensitivity and specificity of the test.
\end{lstlisting}

\begin{lstlisting}[ escapeinside={(*}{*)},backgroundcolor = \color{lightgray}]
ChatGPT
Sure! Let's work through the example using the steps you provided:
Step 1:
Disease: COVID-19
Test Result: Positive
Covariate: Cough
Step 2:
P(disease | test result, covariates) = P(test result | disease, covariates) * P(disease | covariates) / P(test result | covariates)
Step 3:
P(disease | test result, covariates) = P(test result | disease, covariates) * P(disease | covariates) / P(test result | covariates)
Step 4:
In this formula, the sensitivity of the test corresponds to P(test result | disease, covariates), which is the probability of obtaining a positive test result given that the patient has the disease and exhibits the covariate (cough).
The pre-test probability, which represents the initial likelihood of having the disease before considering the test result, is denoted by P(disease | covariates). It reflects the probability of having the disease based on the covariate (cough) alone, without considering the test result.
I hope this example helps clarify the steps involved in using Bayes' rule to compute the post-test probability. Let me know if you have any further questions!
\end{lstlisting}
The following is correct, but the disease and covariates are not named as covid-19 and cough.
\begin{lstlisting}
P(disease | test result, covariates) = P(test result | disease, covariates) * P(disease | covariates) / P(test result | covariates). Did not rename disease, covariates to covid-19, cough.
\end{lstlisting}

16
\url{https://chatgpt.com/share/0cfe226c-c710-44c2-9727-1b073826efca}
\begin{lstlisting}[ escapeinside={(*}{*)},backgroundcolor = \color{yellow}]
In the following conversation, I will give you prompts. Please use the following steps to respond.
Step 1.  Determine what is the disease, what is the test result, and what is the covariate. 
Step 2. Write P(disease | test result, covariates) using Bayes rule. Condition on covariates throughout. Do not make any independence assumptions. Write the denominator using the law of total probability, conditional on covariates.
Step 3. Write the last formula in terms of the disease, test result, and covariate given in the prompt.
Step 4. State which parts of the last formula correspond to the sensitivity and pre-test probability. 
If you understand these steps, please say only "understood".
\end{lstlisting}

\begin{lstlisting}[ escapeinside={(*}{*)},backgroundcolor = \color{lightgray}]
ChatGPT
Understood.
\end{lstlisting}

\begin{lstlisting}[ escapeinside={(*}{*)},backgroundcolor = \color{yellow}]
Can you give an example of using Bayes rule to compute post test probability in a patient with a positive Covid test and a cough using sensitivity and specificity of the test.
\end{lstlisting}

\begin{lstlisting}[ escapeinside={(*}{*)},backgroundcolor = \color{lightgray}]
ChatGPT
Sure! Let's go through the steps using your example:
Step 1:
		Disease: Covid
		Test result: Positive
		Covariate: Cough
Step 2:
We want to calculate P(Covid | Positive test, Cough) using Bayes rule. According to Bayes rule:
P(Covid | Positive test, Cough) = [P(Positive test | Covid, Cough) * P(Covid | Cough)] / P(Positive test | Cough)
Using the law of total probability, we can express the denominator as:
P(Positive test | Cough) = [P(Positive test | Covid, Cough) * P(Covid | Cough)] + [P(Positive test | Not Covid, Cough) * P(Not Covid | Cough)]
Step 3:
Substituting the values from the prompt into the formula, we have:
P(Covid | Positive test, Cough) = [P(Positive test | Covid, Cough) * P(Covid | Cough)] / [P(Positive test | Covid, Cough) * P(Covid | Cough) + P(Positive test | Not Covid, Cough) * P(Not Covid | Cough)]
Step 4:
In the last formula:
		The numerator [P(Positive test | Covid, Cough) * P(Covid | Cough)] corresponds to the sensitivity and pre-test probability.
		The denominator [P(Positive test | Covid, Cough) * P(Covid | Cough) + P(Positive test | Not Covid, Cough) * P(Not Covid | Cough)] corresponds to the total probability of a positive test given the disease and covariate.
By plugging in the appropriate values for sensitivity, pre-test probability, and specificity, you can compute the post-test probability of Covid given a positive test and the presence of a cough.
\end{lstlisting}
The following is correct.
\begin{lstlisting}
P(Covid | Positive test, Cough) = [P(Positive test | Covid, Cough) * P(Covid | Cough)] / [P(Positive test | Covid, Cough) * P(Covid | Cough) + P(Positive test | Not Covid, Cough) * P(Not Covid | Cough)]
\end{lstlisting}

17
\url{https://chatgpt.com/share/16e27531-cf77-4175-9f86-6ac825fb1908}
\begin{lstlisting}[ escapeinside={(*}{*)},backgroundcolor = \color{yellow}]
In the following conversation, I will give you prompts. Please use the following steps to respond.
Step 1.  Determine what is the disease, what is the test result, and what is the covariate. 
Step 2. Write P(disease | test result, covariates) using Bayes rule. Condition on covariates throughout. Do not make any independence assumptions. Write the denominator using the law of total probability, conditional on covariates.
Step 3. Write the last formula in terms of the disease, test result, and covariate given in the prompt.
Step 4. State which parts of the last formula correspond to the sensitivity and pre-test probability. 
If you understand these steps, please say only "understood".
\end{lstlisting}

\begin{lstlisting}[ escapeinside={(*}{*)},backgroundcolor = \color{lightgray}]
ChatGPT
Understood.
\end{lstlisting}

\begin{lstlisting}[ escapeinside={(*}{*)},backgroundcolor = \color{yellow}]
Can you give an example of using Bayes rule to compute post test probability in a patient with a positive Covid test and a cough using sensitivity and specificity of the test.
\end{lstlisting}

\begin{lstlisting}[ escapeinside={(*}{*)},backgroundcolor = \color{lightgray}]
ChatGPT
Certainly! Let's go through the steps using your example:
Step 1.
		Disease: Covid
		Test result: Positive
		Covariate: Cough
Step 2.
Using Bayes' rule, we can write the formula as:
P(Covid | Positive test, Cough) = (P(Positive test | Covid, Cough) * P(Covid | Cough)) / P(Positive test | Cough)
The denominator, P(Positive test | Cough), can be expanded using the law of total probability as:
P(Positive test | Cough) = P(Positive test | Covid, Cough) * P(Covid | Cough) + P(Positive test | No Covid, Cough) * P(No Covid | Cough)
Step 3.
Plugging in the values from the prompt, the formula becomes:
P(Covid | Positive test, Cough) = (P(Positive test | Covid, Cough) * P(Covid | Cough)) / (P(Positive test | Covid, Cough) * P(Covid | Cough) + P(Positive test | No Covid, Cough) * P(No Covid | Cough))
Step 4.
In this formula:
		P(Covid | Positive test, Cough) corresponds to the post-test probability of having Covid given a positive test and cough.
		P(Positive test | Covid, Cough) corresponds to the sensitivity of the test.
		P(Covid | Cough) corresponds to the pre-test probability or the prior probability of having Covid given the presence of a cough.
Please note that the specificity of the test (probability of a negative test result given the absence of the disease) is not explicitly used in this formula. If you'd like, I can provide a more detailed explanation or perform a numerical calculation using specific values
\end{lstlisting}
The following is correct.
\begin{lstlisting}
P(Covid | Positive test, Cough) = (P(Positive test | Covid, Cough) * P(Covid | Cough)) / (P(Positive test | Covid, Cough) * P(Covid | Cough) + P(Positive test | No Covid, Cough) * P(No Covid | Cough))
\end{lstlisting}

18
\url{https://chatgpt.com/share/8586010a-4c2c-4587-afb9-a779f17df408}
\begin{lstlisting}[ escapeinside={(*}{*)},backgroundcolor = \color{yellow}]
In the following conversation, I will give you prompts. Please use the following steps to respond.
Step 1.  Determine what is the disease, what is the test result, and what is the covariate. 
Step 2. Write P(disease | test result, covariates) using Bayes rule. Condition on covariates throughout. Do not make any independence assumptions. Write the denominator using the law of total probability, conditional on covariates.
Step 3. Write the last formula in terms of the disease, test result, and covariate given in the prompt.
Step 4. State which parts of the last formula correspond to the sensitivity and pre-test probability. 
If you understand these steps, please say only "understood".
\end{lstlisting}

\begin{lstlisting}[ escapeinside={(*}{*)},backgroundcolor = \color{lightgray}]
ChatGPT
Understood.
\end{lstlisting}

\begin{lstlisting}[ escapeinside={(*}{*)},backgroundcolor = \color{yellow}]
Can you give an example of using Bayes rule to compute post test probability in a patient with a positive Covid test and a cough using sensitivity and specificity of the test.
\end{lstlisting}

\begin{lstlisting}[ escapeinside={(*}{*)},backgroundcolor = \color{lightgray}]
ChatGPT
Certainly! Let's go through the steps using the example you provided.
Step 1: Determine the disease, test result, and covariate.
Disease: COVID-19
Test result: Positive
Covariate: Cough
Step 2: Write P(disease | test result, covariates) using Bayes rule.
P(disease | test result, covariates) = (P(test result | disease, covariates) * P(disease | covariates)) / P(test result | covariates)
Step 3: Write the formula in terms of the given information.
P(COVID-19 | Positive test, Cough) = (P(Positive test | COVID-19, Cough) * P(COVID-19 | Cough)) / P(Positive test | Cough)
Step 4: Identify sensitivity and pre-test probability.
In this case, sensitivity is the probability of a positive test given that the patient has COVID-19 and a cough, and pre-test probability is the probability of having COVID-19 given the presence of a cough. These values are not explicitly given in the prompt and would need to be provided to calculate the post-test probability accurately.
Please note that without specific values for sensitivity, specificity, and pre-test probability, we cannot calculate the exact post-test probability in this example. These values are essential for an accurate calculation.
If you have the specific values, please provide them, and I can help you calculate the post-test probability.
\end{lstlisting}

The following is correct, but the denominator is not expanded.
\begin{lstlisting}
P(COVID-19 | Positive test, Cough) = (P(Positive test | COVID-19, Cough) * P(COVID-19 | Cough)) / P(Positive test | Cough)
\end{lstlisting}

19
\url{https://chatgpt.com/share/5d0e0059-d22e-4020-8967-d8ec4bdc97ab}
\begin{lstlisting}[ escapeinside={(*}{*)},backgroundcolor = \color{yellow}]
In the following conversation, I will give you prompts. Please use the following steps to respond.
Step 1.  Determine what is the disease, what is the test result, and what is the covariate. 
Step 2. Write P(disease | test result, covariates) using Bayes rule. Condition on covariates throughout. Do not make any independence assumptions. Write the denominator using the law of total probability, conditional on covariates.
Step 3. Write the last formula in terms of the disease, test result, and covariate given in the prompt.
Step 4. State which parts of the last formula correspond to the sensitivity and pre-test probability. 
If you understand these steps, please say only "understood".
\end{lstlisting}

\begin{lstlisting}[ escapeinside={(*}{*)},backgroundcolor = \color{lightgray}]
ChatGPT
Understood.
\end{lstlisting}

\begin{lstlisting}[ escapeinside={(*}{*)},backgroundcolor = \color{yellow}]
Can you give an example of using Bayes rule to compute post test probability in a patient with a positive Covid test and a cough using sensitivity and specificity of the test.
\end{lstlisting}

\begin{lstlisting}[ escapeinside={(*}{*)},backgroundcolor = \color{lightgray}]
ChatGPT
Certainly! Let's go through the steps using the provided example:
Step 1:
		Disease: Covid
		Test result: Positive
		Covariate: Cough
Step 2:
We can write the formula using Bayes' rule as follows:
P(Covid | Positive test, Cough) = (P(Positive test | Covid, Cough) * P(Covid | Cough)) / P(Positive test | Cough)
The denominator can be expanded using the law of total probability:
P(Positive test | Cough) = P(Positive test | Covid, Cough) * P(Covid | Cough) + P(Positive test | No Covid, Cough) * P(No Covid | Cough)
Step 3:
Now, let's rewrite the formula using the given information:
P(Covid | Positive test, Cough) = (Sensitivity * P(Covid | Cough)) / (Sensitivity * P(Covid | Cough) + (1 - Specificity) * P(No Covid | Cough))
Step 4:
In the formula:
		Sensitivity corresponds to the probability of a positive test given the presence of the disease (Covid).
		Pre-test probability corresponds to P(Covid | Cough), which is the probability of having Covid given the presence of a cough.
Please note that we need additional information about the pre-test probability and specific values for sensitivity and specificity in order to calculate the post-test probability accurately.
Is there anything else I can assist you with?
\end{lstlisting}
The following is correct.
\begin{lstlisting}
P(Covid | Positive test, Cough) = (P(Positive test | Covid, Cough) * P(Covid | Cough)) / P(Positive test | Cough) with denominator expanded P(Positive test | Cough) = P(Positive test | Covid, Cough) * P(Covid | Cough) + P(Positive test | No Covid, Cough) * P(No Covid | Cough)
\end{lstlisting}

20
\url{https://chatgpt.com/share/d2038337-ebb6-4718-9fae-3ee05912e7ed}
\begin{lstlisting}[ escapeinside={(*}{*)},backgroundcolor = \color{yellow}]
In the following conversation, I will give you prompts. Please use the following steps to respond.
Step 1.  Determine what is the disease, what is the test result, and what is the covariate. 
Step 2. Write P(disease | test result, covariates) using Bayes rule. Condition on covariates throughout. Do not make any independence assumptions. Write the denominator using the law of total probability, conditional on covariates.
Step 3. Write the last formula in terms of the disease, test result, and covariate given in the prompt.
Step 4. State which parts of the last formula correspond to the sensitivity and pre-test probability. 
If you understand these steps, please say only "understood".
\end{lstlisting}

\begin{lstlisting}[ escapeinside={(*}{*)},backgroundcolor = \color{lightgray}]
ChatGPT
Understood. Please proceed with the first prompt.
\end{lstlisting}

\begin{lstlisting}[ escapeinside={(*}{*)},backgroundcolor = \color{yellow}]
Can you give an example of using Bayes rule to compute post test probability in a patient with a positive Covid test and a cough using sensitivity and specificity of the test.
\end{lstlisting}

\begin{lstlisting}[ escapeinside={(*}{*)},backgroundcolor = \color{lightgray}]
ChatGPT
Understood. Let's go through the steps using the provided example:
Step 1: Determine the disease, test result, and covariate.
		Disease: Covid
		Test result: Positive
		Covariate: Cough
Step 2: Write P(disease | test result, covariates) using Bayes rule.
P(disease | test result, covariates) = [P(test result | disease, covariates) * P(disease | covariates)] / P(test result | covariates)
Step 3: Write the formula in terms of the given disease, test result, and covariate.
P(Covid | positive test, cough) = [P(positive test | Covid, cough) * P(Covid | cough)] / P(positive test | cough)
Step 4: Identify sensitivity and pre-test probability.
		Sensitivity corresponds to P(positive test | Covid, cough)
		Pre-test probability corresponds to P(Covid | cough)
Please note that we haven't used specificity in this example as it was not mentioned in the prompt. If you'd like to include specificity, please provide the corresponding value.
\end{lstlisting}
The following is correct, but the denominator is not expanded.
\begin{lstlisting}
P(Covid | positive test, cough) = [P(positive test | Covid, cough) * P(Covid | cough)] / P(positive test | cough)
\end{lstlisting}

\normalsize







\end{document}